%% file: main.tex
\pdfoutput=1

\documentclass[11pt]{article}

\usepackage{acl}

\usepackage{times}
\usepackage{latexsym}

\usepackage[T1]{fontenc}

\usepackage[utf8]{inputenc}

\usepackage{microtype}

\usepackage{amsfonts,amsmath}
\usepackage{IEEEtrantools}
\usepackage{booktabs}
\usepackage{tabularx}
\usepackage{graphicx}
\usepackage{caption}
\usepackage{subcaption}
\usepackage{bbm}
\usepackage{enumitem}

\newcommand*{\Scale}[2][4]{\scalebox{#1}{$#2$}}%

\title{Measuring the Instability of Fine-Tuning}

\author{
  Yupei Du and Dong Nguyen\\
  Utrecht University\\
  Utrecht, the Netherlands \\
  \texttt{\{y.du,d.p.nguyen\}@uu.nl} \\
  }

\begin{document}
\maketitle
\begin{abstract}
  Fine-tuning pre-trained language models 
  on downstream tasks 
  with varying random seeds has been shown to be unstable,
  especially on small datasets. 
  Many previous studies have investigated this instability and proposed methods to mitigate it.
  However, most studies only 
  used the standard deviation of performance scores (SD) as their measure, 
  which is a narrow characterization of instability.
  In this paper, we analyze SD and six other measures 
  quantifying instability at different levels of granularity.
  Moreover, we propose a systematic framework to evaluate the validity of these measures. 
  Finally, 
  we analyze the consistency and difference between different measures 
  by reassessing existing instability mitigation methods.
  We hope our results will inform the development of better measurements of fine-tuning instability.\footnote{
    Our implementation is available at \url{https://github.com/nlpsoc/instability_measurement}. 
  }

\end{abstract}

\input{intro}

\input{related_work}

\input{instability_measures}
\input{experimental_setup}
\input{measure_analyses}

\input{measure_comparison}

\section{Conclusion}\label{sec:conclusion}
In this paper, we study measures that quantify the 
instability of fine-tuning PLMs. 
In addition to the most commonly used measure, SD, 
we study six other measures at different granularity levels
and propose a framework to evaluate their validity.
Using this framework, 
we show that all these measures except SVCCA have good validity.
Moreover, by reassessing existing IMMs,
we show that different instability measures are not always consistent,
and that they are more consistent when the models are less stable.
Finally, 
based on our observations, 
we offer two suggestions for selecting instability measures 
in future studies.

\section*{Limitations}\label{sec:limitation}
Our study leaves room for future work.
First, we would like to highlight the difficulty of 
applying the validity assessment framework from measurement theory to instability measures. 
For example, in \S\ref{subsec:convergent-validity}, 
our low convergent validity scores may have different interpretations  
because there are no well-established instability measures.
Further, in \S\ref{subsec:concurrent-validity}, 
because no previous studies have built theoretical foundations of 
factors that impact the prediction and representation instability, 
both our tests do not rigorously follow the concurrent validity definition: 
our first test of successful and failed runs 
is based on an assumption derived from observations of \citet{mosbach2021on} rather than theory, 
and our second test of differences among test datasets 
examines the consistency between theoretically indistinguishable groups 
instead of the differences between theoretically distinguishable groups. 

Second, we only experimented with
a limited number of tasks, instability measures, PLMs, and validity types.
Future work can use our framework
to further validate the generalizability of our observations. 
For example, to apply our validity testing framework to larger datasets, 
to include other measures 
(e.g. functional similarity measures,~\citealp{csiszarik2021similarity} 
and jitter,~\citealp{model-stability-with-continuous-data}), 
to study generative PLMs (e.g.~T5,~\citealp{t5} and OPT,~\citealp{zhang2022opt}), 
and to test other types and validity (e.g.~discriminative and predictive validity). 

Third, we focused on general text classification tasks in this paper.
One promising direction is to investigate
which measures to use for specific settings. 
For example, to extend our framework to more recent generative models 
(e.g.~BART,~\citealp{bart} and GPT-3,~\citep{gpt3}). 
However, in this case, 
because our prediction measures in \S\ref{sec:instability_measures} 
are only useful for classification, 
new prediction measures should be developed, 
and our tests should be adjusted accordingly.

\section*{Acknowledgements}
This work is part of the research programme Veni
with project number VI.Veni.192.130, which is
(partly) financed by the Dutch Research Council
(NWO).

\bibliography{custom}
\bibliographystyle{acl_natbib}

\clearpage
\appendix

\input{app.tex}

\end{document}

%% file: intro.tex
\section{Introduction}\label{sec:intro}

Since the introduction of BERT~\citep{devlin-etal-2019-bert},
the pre-train-then-fine-tune paradigm has achieved state-of-the-art performance
across many NLP benchmarks \citep{sun2021ernie,fedus2021switch,chi2021xlm}.
However, despite its wide success, 
the fine-tuning process,
especially when fine-tuning large models on \emph{small datasets}, 
is shown to be unstable:
fine-tuning a given model with varying random seeds can lead to different performance results
\citep{Lee2020Mixout,dodge2020fine,mosbach2021on,hua-etal-2021-noise}.
This instability makes the investigation of better architectures 
and instability mitigation methods (IMMs) challenging
~\citep{zhang2021revisiting}.

Many previous studies have investigated fine-tuning instability
\citep{dodge2020fine,Lee2020Mixout,mosbach2021on,zhang2021revisiting}.
In these studies, the most prevalent instability measure 
is the \emph{standard deviation of performance}
(\textbf{SD}, e.g. the standard deviation of F1-scores). 
However, as we discuss in \S\ref{sec:instability_measures} and \S\ref{sec:relationship},
SD can only offer very limited assessments.
For example, classifiers can obtain the same accuracy score (i.e.~zero SD) even when
they neither make the same predictions on each example (\emph{prediction instability})
nor have the same hidden representations (\emph{representation instability}).
Therefore, it is important to also use other measures that
can address the weaknesses of SD. 

However, it is difficult to decide which measures to use:
because instability is an abstract concept, 
it is hard to examine 
\emph{to which extent a measure indeed quantifies what it intends to measure}.
This property is called \textbf{validity} in measurement theory \citep{trochim2001research}.
For example, 
using the average accuracy of models as an instability measure would have low validity, 
because how accurate these models make predictions
does not reflect their stability.

To better assess the instability of fine-tuning pre-trained language models (PLMs), 
we study more measures concerning instability at different granularity levels
\citep{pmlr-v139-summers21a,khurana-etal-2021-emotionally,svcca,pmlr-v97-kornblith19a,ding2021grounding} 
and develop a framework to assess their validity. 
We focus on BERT and RoBERTa for their popularity, 
but our framework can also be applied to other PLMs. 
Concretely,
\setlist{nolistsep,leftmargin=*}
\begin{itemize}[noitemsep]
    \item First, we discuss six other instability measures 
        at different granularity levels in addition to SD, and
        categorize them into \emph{prediction measures} and \emph{representation measures}
        based on the type of instability they focus on (\S\ref{sec:instability_measures}).
    \item Second, we propose a framework to systematically assess  
        two types of validity of these measures,
        without relying on labelled data 
        (\S\ref{sec:validity-of-representation-measures}).
    \item Third, we investigate the consistency and differences 
        between different measures 
        by reassessing the effectiveness of existing IMMs,
        analyzing their correlations (\S\ref{subsec:reassess}),
        and performing bootstrap analyses (\S\ref{subsec:bootstrap}). 
        We find that measures at different granularity levels
        do not always produce consistent instability scores with each other 
        and tend to differ more when the models are more stable.
        Moreover, based on our observations, 
        we offer two suggestions for future studies:
        (1) use multiple instability measures, 
        especially when models are more stable;
        (2) use only one prediction and one representation measure 
        when limited computational resources are available (\S\ref{subsec:implications}). 
        
\end{itemize}

%% file: related_work.tex
\section{Background}\label{sec:background}

\subsection{Instability of  Fine-tuning}\label{subsec:instability-of-bert-fine-tuning}

The seminal work of BERT by \citet{devlin-etal-2019-bert} 
has already shown that fine-tuning PLMs is unstable 
regarding the choices of random seeds. 
This observation was further confirmed by other studies on more PLMs, 
including RoBERTa 
\citep{liu2019roberta,Lan2020ALBERT,phang2018sentence,Lee2020Mixout,dodge2020fine,mosbach2021on,zhang2021revisiting,sellam2021multiberts}.
Most of these studies used SD to measure the instability.

Different explanations have been proposed to account for 
the instability of fine-tuning PLMs on small datasets, including
catastrophic forgetting \citep{Lee2020Mixout}\footnote{
Although several studies assumed that catastrophic forgetting causes instability 
\citep{Kirkpatrick3521,pmlr-v80-schwarz18a,Lee2020Mixout}, 
\citet{mosbach2021on} argued against it.}, 
the lack of Adam bias correction \citep{mosbach2021on,zhang2021revisiting}, 
too few training steps \citep{mosbach2021on}, 
and task-specific top layers \citep{zhang2021revisiting}. 

\subsection{Instability Mitigation Methods (IMMs)}\label{subsec:instability-mitigation-methods}

Various IMMs have been used to mitigate the instability of fine-tuning PLMs.
Following \citet{zhang2021revisiting}, 
we focus on four methods for their popularity. 
Nevertheless,  we acknowledge the existence of other methods, 
including entropy regularization and co-distillation \citep{churn}, 
and component-wise gradient norm clipping \citep{yang-ma-2022-improving}. 

\paragraph{Mixout} \citep{Lee2020Mixout} 
is a generalized version of Dropout \citep{JMLR:v15:srivastava14a}.
It randomly replaces the outputs of neurons with 
the ones produced by the pre-trained weights by a probability $p$. 
In this way, it can mitigate the catastrophic forgetting of pre-trained knowledge
which potentially stabilizes fine-tuning.

\paragraph{$\mathbf{{WD}_{pre}}$} \citep{pmlr-v80-li18a} is a variant of weight decay:
after each optimization step,
each model weight $w$ will move a step size of $\lambda w$ 
towards the pre-trained weights,
where $\lambda$ is a hyper-parameter. 
$\mathsf{{WD}_{pre}}$ also aims to improve the fine-tuning instability 
by mitigating catastrophic forgetting. 

\paragraph{Layer-wise Learning Rate Decay} 
(\citealp{howard-ruder-2018-universal}, LLRD) 
assigns decreasing learning rates
from the topmost layer to the bottom layer by a constant hyper-parameter discounting factor $\eta$.
\citet{howard-ruder-2018-universal} empirically show that 
models trained using LLRD are more stable, 
by retaining more generalizable pre-trained knowledge in bottom layers,
while forgetting specialized pre-train knowledge in top layers. 

\paragraph{Re-init} \citep{zhang2021revisiting} stabilizes fine-tuning 
by re-initializing the top $k$ layers of PLMs.
The underlying intuition is similar to LLRD:
top layers of PLMs contain more pre-train task specific knowledge, 
and transferring it may hurt stability.

%% file: instability_measures.tex
\section{Instability Measures}\label{sec:instability_measures}

Despite its wide usage,
\emph{SD only provides a narrow view of the instability of models}.
For example, consider fine-tuning two pre-trained models on the same classification task.
If one of them makes correct predictions only on half of the test data,
while the other model makes correct predictions only on the other half,
these two models will both have a 0.5 accuracy score
and therefore no instability would be measured using SD.
However, they actually make different \emph{predictions} on each data point
(i.e.~\textbf{prediction instability}).
Moreover, even if these two models 
achieve the same accuracy by making identical predictions,
due to the over-parameterization of PLMs \citep{pmlr-v139-roeder21a},
they can have different sets of hidden \emph{representations}
(i.e.~\textbf{representation instability}).

To better assess these two types of instability,
we study six other instability measures at different granularity levels 
in addition to SD.
Furthermore, according to the instability types 
that these measures intend to quantify,
we categorize these measures into two types:
\emph{prediction measures} (\S\ref{subsec:prediction_measures})
and \emph{representation measures} (\S\ref{subsec:representation_measures}). 
All these instability measures 
have a continuous output range 0--1,
with higher values indicating lower stability. 

It is worth noting that 
similar categorizations have been used before. 
For example, \citet{csiszarik2021similarity} categorized measures 
as \emph{functional} and \emph{representational}. 
However, they used functional similarity 
to refer to the function compositions that different components of the models realize. 
Also, \citet{pmlr-v139-summers21a} categorized measures as 
\emph{performance variability} and \emph{representation diversity}. 
However, they used \emph{performance variability} to specifically refer to SD 
and used \emph{representation diversity} to refer to 
all other measures at different granularity levels that we study here.

\paragraph{Notation}
Formally, suppose we have a dataset consisting of $n$ data points.
We fine-tune $m$ BERT models $\{M_1, M_2, \dots, M_m\}$,
with the same settings except for $m$ different random seeds.
We use $p_i^k$ and $\hat{y}_i^k$ to denote the class probability and the prediction
of $M_i$ on the $k$-th test sample.

Assume the $l$-th layer of $M_i$ consists of $e$ neurons,
we use $M_i^l \in \mathbb{R}^{n \times e}$ to denote this layer's \emph{centered} representation,
w.r.t. all $n$ data points 
(all representation measures discussed below require us to center the representations).
Representation measures involve computing the distances
between the representations derived from the same layer of two different models.
We use $d_{i, j}^l$ to represent the distance between $M_i^l$ and $M_j^l$.

\subsection{Prediction Measures}\label{subsec:prediction_measures}
We refer to measures that assess the prediction instability of models as prediction measures.
In other words, prediction measures 
only assess the output of the models (i.e.~logits and predictions).
In this paper, we study three prediction measures besides SD:
\emph{pairwise disagreement}, \emph{Fleiss' Kappa},
and \emph{pairwise Jensen-Shannon divergence (pairwise JSD)}.
Among these three measures, 
both pairwise disagreement and Fleiss' Kappa quantify 
the instability of the discrete predictions of models, 
and therefore are at the same granularity level. 
Pairwise JSD looks at continuous class probabilities 
and is thus more fine-grained. 
Nevertheless, they are all more fine-grained than SD, 
which only considers the overall performance. 

\paragraph{Pairwise Disagreement}
Following \citet{pmlr-v139-summers21a}, we measure the models' instability by
averaging the \emph{pairwise disagreement} among models' predictions.
Formally,
\begin{IEEEeqnarray}{rCl}
    \mathcal{I}_{\mathrm{pwd}} =
    \frac{2}{nm(m-1)}\sum\limits_{i=1}^{m}\sum\limits_{j=i+1}^{m}\sum\limits_{k=1}^{n}
    \mathbbm{1}(\hat{y}^{k}_{i} \neq \hat{y}^{k}_{j}), \notag
\end{IEEEeqnarray}
where
$\mathbbm{1}$ is the indicator function.
We note that 
our definition of pairwise disagreement relates closely to \emph{churn} and \emph{jitter} 
proposed and used by \citet{churn_origin} and \citet{churn,model-stability-with-continuous-data}. 

\paragraph{Fleiss' Kappa}
Similar to \citet{khurana-etal-2021-emotionally},
we adopt Fleiss' Kappa, which is
a popular measure for inter-rater consistency \citep{fleiss1971measuring},
to measure the consistency among different models' predictions.
Because Fleiss' Kappa is negatively correlated with models' instability
and ranges from 0 to 1, 
we use its difference with one as the output, to stay consistent with other measures.
Formally,
\begin{IEEEeqnarray}{rCl}
    \mathcal{I}_{\mathrm{\kappa}}=1 - \frac{p_{a}-p_{\epsilon}}{1-p_{\epsilon}}, \notag
\end{IEEEeqnarray}
where $p_{a}$ is a term evaluating the consistency of models' predictions on each test sample,
and $p_{\epsilon}$ is an error correction term
(Details in Appendix \ref{app:feliss_kappa}).

\paragraph{Pairwise JSD}
The previous two measures only look at discrete labels,
while continuous class probabilities contain richer information 
about a model's predictions.
Therefore, we average the pairwise JSD of models' class probabilities
to obtain a finer-grained evaluation of instability.
Formally,
\begin{IEEEeqnarray}{rCl}
    \mathcal{I}_{\mathrm{JSD}} =
    \frac{2}{nm(m-1)}\sum\limits_{i=1}^{m}\sum\limits_{j=i+1}^{m}\sum\limits_{k=1}^{n}
    JSD(p^{k}_{i} \| p^{k}_{j}), \notag
\end{IEEEeqnarray}
where
$JSD(\cdot\|\star)$ is the JSD between $\cdot$ and $\star$.

\subsection{Representation Measures}\label{subsec:representation_measures}

We refer to measures that assess the instability of models
based on their hidden representations as representation measures. 
Here, we study three representation measures:
\emph{singular vector canonical correlation analysis} (SVCCA,~\citealp{svcca}),
\emph{orthogonal Procrustes distance} (OP,~\citealp{schonemann1966generalized}),
and \emph{linear centered kernel alignment} (Linear-CKA,~\citealp{pmlr-v97-kornblith19a}).
Because all representation measures 
look at the hidden representations of models, 
they are at the same granularity level, 
which is more fine-grained than prediction measures. 

All these three measures are originally developed
to compute the distance between a pair of representations 
(although Linear-CKA is also used by \citet{pmlr-v139-summers21a} 
to study model instability).
With these measures, we are able to analyze the behavior of neural networks, 
going beyond the model predictions alone \citep{pmlr-v97-kornblith19a}.
To evaluate the instability of all $m$ models regarding a specific layer $l$, $\mathcal{I}^l$,
we average the distance $d$ of each possible pair of models.
Formally,
\begin{IEEEeqnarray}{rCl}
    \textstyle
    \mathcal{I}^l = \frac{2}{m(m-1)}\sum_{i=1}^{m}\sum_{j=i+1}^{m} d_{i, j}^l.\notag
\end{IEEEeqnarray}
We next describe how to calculate $d$ for each representation measure.
We respectively denote the instability score of each measure after aggregating $d$ as 
$\mathcal{I}_{\mathrm{SVCCA}}$, $\mathcal{I}_{\mathrm{CKA}}$, and $\mathcal{I}_{\mathrm{OP}}$.

\paragraph{SVCCA} \citep{svcca} 
is developed based on canonical correlation analysis
(CCA, \citealp{hardoon-etal-2004-canonical}).
For two representations $M_i^l$ and $M_j^l$,
CCA finds $e$ orthogonal bases so that their correlations after projection are maximized.

Formally, for $\textstyle{1 \leq k \leq e}$,
\begin{IEEEeqnarray}{rCl}
    &\rho_{k}=\max _{\mathbf{w}_{i}^{k}, \mathbf{w}_{j}^{k}}
    \operatorname{corr}\left(M_i^l \mathbf{w}_{i}^{k}, M_j^l \mathbf{w}_{i}^{k}\right), \notag \\
    &
    \Scale[0.85]{
    \text {s.t.}  \;\forall k_1<k_2, \;
    M_i^l \mathbf{w}_{i}^{k_1} \perp  M_i^l \mathbf{w}_{i}^{k_2}
    \text{ and } M_j^l \mathbf{w}_{j}^{k_1} \perp  M_j^l \mathbf{w}_{j}^{k_2}, \notag}
\end{IEEEeqnarray}
where $\mathbf{w}_{i}^{k}, \mathbf{w}_{j}^{k} \in \mathbb{R}^{p_1}$.
After obtaining $\rho$, we use the \emph{mean correlation coefficient} to transform $\rho$
into a scalar dissimilarity measure. Formally,
\begin{IEEEeqnarray}{cc}
    \textstyle
    d_{\mathrm{CCA}} =1-\frac{1}{e} \sum_{k=1}^e \rho_{k}. \notag
\end{IEEEeqnarray}

\citet{svcca} find that meaningful information usually
distributes in a lower-dimensional subspace of the neural representations.
To avoid overfitting on noise, 
SVCCA first uses singular-value decomposition
to find the most important subspace directions of the representations.\footnote{
    Following \citet{svcca}, we keep directions that
    explain 99\% of the representations.}
The representations are then projected onto these directions,
followed by CCA.
We again calculate the mean $\rho$ as the $d_{SVCCA}$.

\paragraph{OP} \citep{ding2021grounding} consists of computing the minimum Frobenius norm
of the difference between $M_i^l$ and $M_j^l$,
after $M_i^l$ being transformed by an orthogonal transformation.
Formally,
\begin{IEEEeqnarray}{cc}
    \min _{R}\|M_j^l- M_i^l R\|_{\mathrm{F}}^{2}, \;\; s.t. \;  R^{\top} R=I. \notag
\end{IEEEeqnarray}
\citet{schonemann1966generalized} provides a closed-form solution of this problem.
To constrain the output range to be between zero and one,
we normalize the representations with their Frobenius norms.
Formally,
\begin{IEEEeqnarray}{cc}
    \textstyle
    d_{\text{OP}}(M_i^l, M_j^l)= 1-\frac{\left\|M_i^{l \top} M_j^l \right\|_{*}}
    {\left\|M_i^{l \top} M_i^l\right\|_{F}\left\|M_j^{l \top} M_j^l \right\|_{F}},\notag
\end{IEEEeqnarray}
where $\|\cdot\|_{*}$ is the nuclear norm.

\paragraph{Linear-CKA} measures the representation distance
by the similarity between
representations' inter-sample similarity $\langle M_i^{l \top} M_i^l, M_j^{l \top} M_j^l\rangle$
\citep{pmlr-v97-kornblith19a}.
After normalizing the representations with Frobenius norms,
we obtain a similarity score between zero and one.
We then use its difference with one as the distance measure. Formally,
\begin{IEEEeqnarray}{cc}
    \textstyle
    d_{\text{CKA}}(M_i^l, M_j^l)= 1-\frac{\left\|M_i^{l \top} M_j^l \right\|_{F}^{2}}
    {\left\|M_i^{l \top} M_i^l\right\|_{F}\left\|M_j^{l \top} M_j^l \right\|_{F}}.\notag
\end{IEEEeqnarray}

%% file: experimental_setup.tex
\section{Experimental Setup}\label{sec:experimental-setup}

We study the instability
of fine-tuning BERT \citep{devlin-etal-2019-bert} and RoBERTa \citep{liu2019roberta} empirically. 
Following \citet{Lee2020Mixout} and \citet{zhang2021revisiting},
we perform our experiments on three small datasets of
the GLUE benchmark \citep{wang-etal-2018-glue}: 
RTE, MRPC~\citep{dolan-brockett-2005-automatically}, and CoLA \citep{warstadt-etal-2019-neural},
because models trained on small datasets are observed to be less stable
\citep{zhang2021revisiting}.\footnote{
    To study whether our findings also generalize to larger datasets,
    we include a pilot study on SST-2 ($8\times$ larger than CoLA) in Appendix~\ref{app:sst2}. 
    As expected, we observe higher stability.
    Furthermore, the behaviors of measures are consistent with those observed on smaller datasets. 
}

Unless specified,
we fine-tune BERT-large and RoBERTa-large models 
from HuggingFace Transformers \citep{wolf-etal-2020-transformers},
with a 16 batch size, a 0.1 Dropout rate, and a $2 \times 10^{-5}$ learning rate,
using de-biased Adam, 
as well as a linear learning rate warm-up during the first 10\% steps followed by a linear decay, 
following \citet{zhang2021revisiting}.
Consistent with \citet{mosbach2021on},
we train the models for five epochs with 20 random seeds.

Consistent with \citet{zhang2021revisiting},
we divide the validation data into two equally sized parts, 
respectively as new validation and test data, 
because we have no access to the GLUE test datasets.
Moreover, we keep the checkpoint with the highest validation performance
and obtain all our results on the test set.
More details are provided in Appendix~\ref{app:experimental_setup}.

%% file: measure_analyses.tex
\section{Assessing the Validity of Instability Measures}\label{sec:validity-of-representation-measures} 
\begin{figure*}
    \centering
    \begin{subfigure}[b]{0.31\textwidth}
        \centering
        \includegraphics[width=\textwidth]{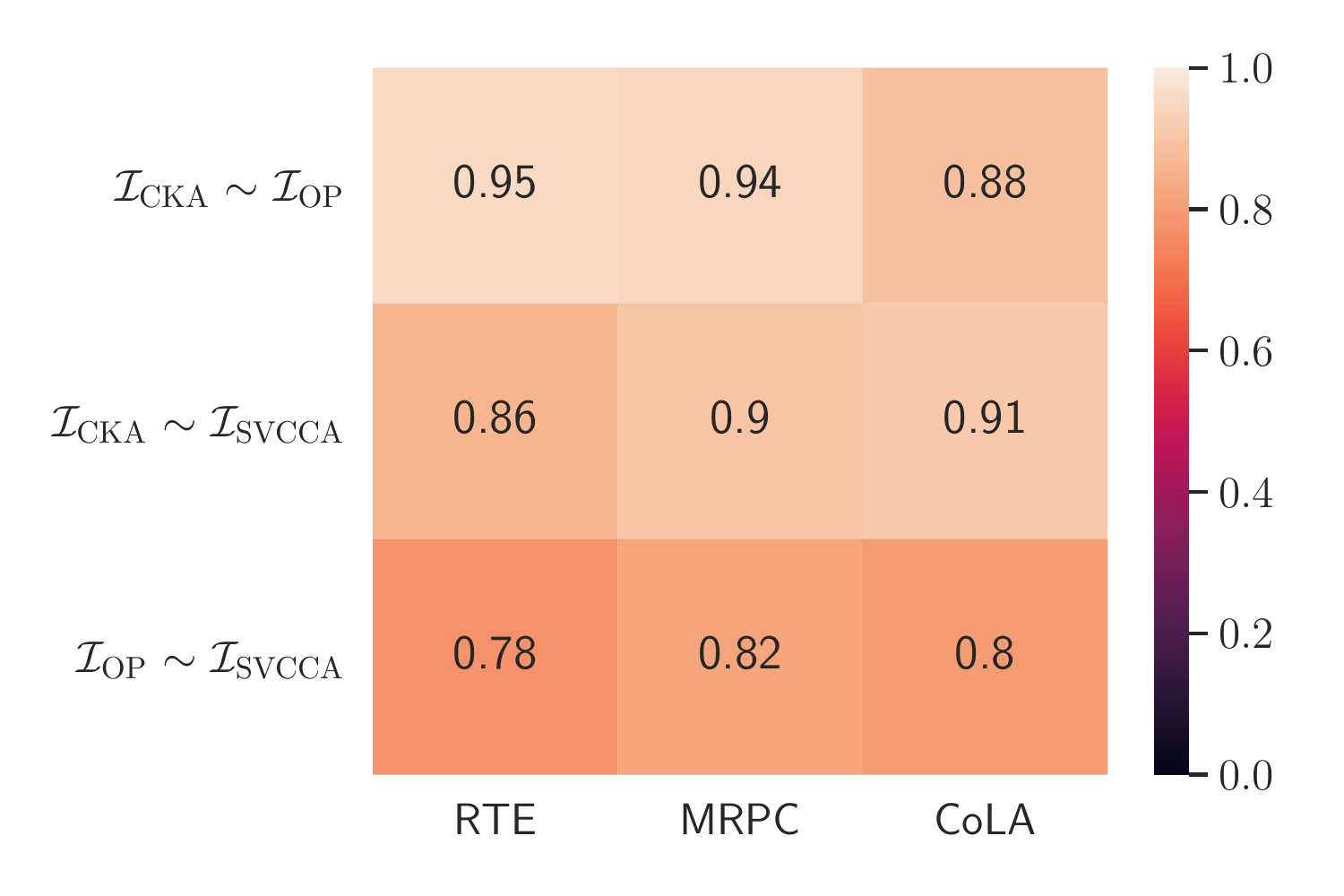}
        \caption{Convergent validity}
        \label{subfig:convergent}
    \end{subfigure}
    \hfill
    \begin{subfigure}[b]{0.31\textwidth}
        \centering
        \includegraphics[width=\textwidth]{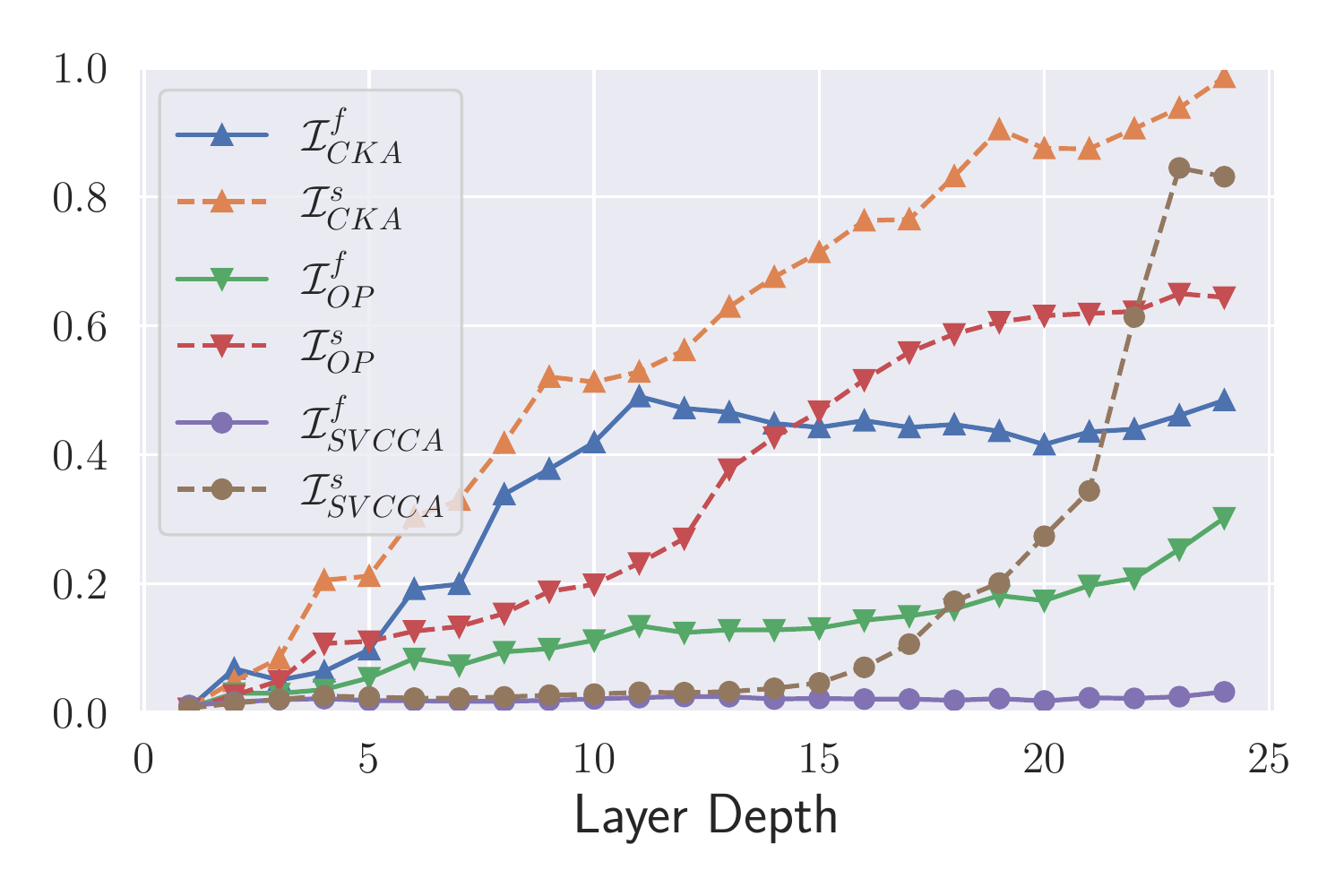}
        \caption{Successful vs. failed runs}
        \label{subfig:sf_runs}
    \end{subfigure}
    \hfill
    \begin{subfigure}[b]{0.31\textwidth}
        \centering
        \includegraphics[width=\textwidth]{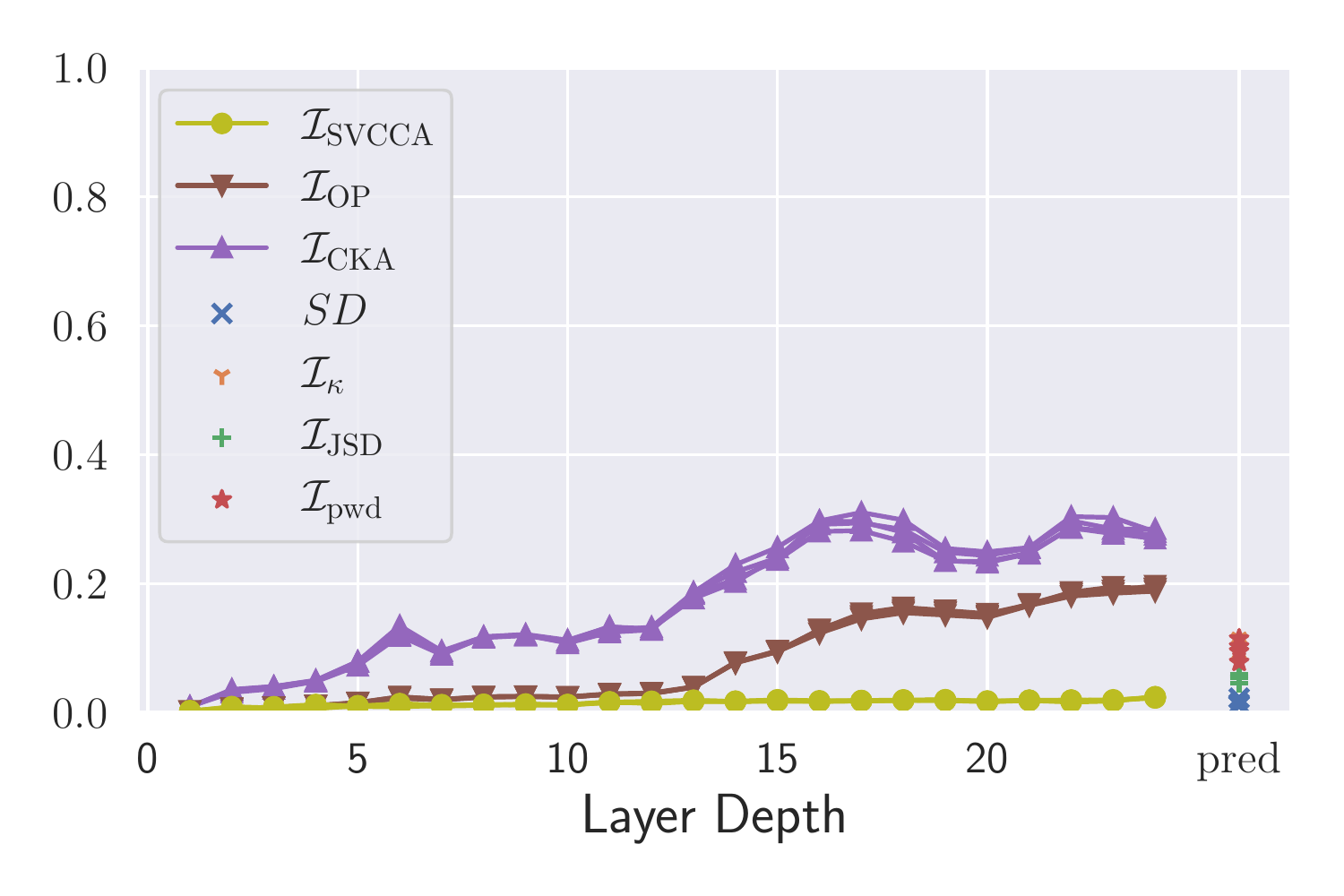}
        \caption{Consistency among sub-samples}
        \label{subfig:subsample}
    \end{subfigure}
    \caption{
        Validity assessment. 
        Figure~\ref{subfig:convergent} show Pearson's $r$ 
        between different representation measures for BERT.
        For Figure~\ref{subfig:sf_runs} and \ref{subfig:subsample}, 
        X-axis and Y-axis are the layer depth and instability scores respectively. 
        Figure~\ref{subfig:sf_runs} shows the differences of representation measures
        between successful ($*^s$) and failed runs ($*^f$) for BERT on RTE.
        Figure~\ref{subfig:subsample} shows the consistency of different measures for BERT on MRPC
        across different sub-samples (different lines).
        }
    \label{fig:three graphs}
\end{figure*}

It is not trivial to assess the validity of instability measures, 
because there is no clear ground truth. 
Nevertheless, 
we can still perform validity assessments 
by building on approaches from measurement theory.
Here, we propose a framework 
to assess two important types of validity \citep{trochim2001research}, 
by computing their correlations with each other 
(convergent validity, \S\ref{subsec:convergent-validity}) 
and observing their responses to different inputs 
(concurrent validity, \S\ref{subsec:concurrent-validity}). 
Except for SVCCA, 
all other measures show good validity in our tests, 
and hence they are suitable for examining fine-tuning instability.\footnote{
We note that passing our tests does not necessarily imply 
that a measure is perfectly valid: 
it is also possible that 
our validity tests/datasets/PLMs are not comprehensive
(i.e. mono-method and mono-operation biases, \citealp{trochim2001research}).
Moreover, as aforementioned in \S\ref{sec:instability_measures}, 
different measures may concern different aspects of instability 
and should usually be used together.
We offer a more extensive discussion in \S\ref{sec:relationship}. 
}
Although there are other types of validity
(e.g. face, content, discriminative, and predictive validity), we select these two types because of 
their relevance to our study and our lack of labelled test data. 
\citet{ding2021grounding} also provided a framework 
to evaluate the sensitivity (i.e.~responding to important changes) 
and specificity (i.e.~ignoring changes that do not matter) of the representation similarity metrics
(i.e. $d_{\mathrm{CKA}}, d_{\mathrm{OP}}, d_{\mathrm{SVCCA}}$).
However, their framework was not build on validity theory
and they did not consider prediction measures.

\subsection{Convergent Validity}\label{subsec:convergent-validity}

In measurement theory, convergent validity refers to
validity established by correlations with measures that are theoretically related 
\citep{gravetter2018research}.
In other words, \emph{if two measures aim to quantify the same underlying concept,
their measurements should have a high correlation}.
It is worth noting that 
convergent validity usually should be evaluated against established and validated measures 
\citep{measurement_book}.
However, in our case, none of the measures have been validated before. 
Therefore, low convergent validity may have different causes: 
for example, it can be that only one of the measures is invalid, 
or that these measures quantify different aspects of the concept.

For representation measures, 
we have an instability score for each hidden layer. 
We therefore assess their convergent validity
by computing Pearson's $r$ between instability scores 
that different measures assign to different layers of the same group of models (e.g. BERT fine-tuned on RTE with different random seeds).
We show the results on BERT in Figure~\ref{subfig:convergent}.
All three representation measures correlate highly 
($> 0.77$) with each other, 
which suggests a good convergent validity.
For prediction measures, 
we only have a single scalar output on each dataset/PLM combination.
It is thus not practical to estimate their convergent validity directly 
because the sample size (i.e.~the number of dataset/PLM combinations) is too small. 
In \S\ref{sec:relationship}, 
we offer a detailed discussion 
and observe that they actually show good convergent validity.

\subsection{Concurrent Validity}\label{subsec:concurrent-validity}

In measurement theory, concurrent validity refers to the
``ability to distinguish between groups that 
it should theoretically be able to distinguish between''~\citep{trochim2001research}. 
We therefore test concurrent validity based on the following assumption:
a valid instability measure
should not only be able to distinguish groups of models with substantial instability differences,
but also be unable to distinguish groups of models with trivial instability differences. 
Concretely, we treat substantial instability differences as
the differences \emph{between successful/failed fine-tuning runs},
and define trivial instability differences as \emph{different i.i.d. test datasets}.\footnote{
    Our tests are inspired by the concurrent validity definition,
    rather than strictly following it. 
    See the limitations section. 
} 
We accordingly present two analyses.

\paragraph{Differences between successful and failed runs}

Previous studies have identified failed fine-tuning runs
where the training fails to converge~\citep{dodge2020fine,mosbach2021on}.\footnote{
    Following \citet{dodge2020fine} and \citet{mosbach2021on},
    we define failed runs as runs whose accuracies at the end of the training
    $\leq$
    a majority classifier
    (i.e. a classifier that consistently predicts the majority label regardless of the inputs).
}
In particular, \citet{mosbach2021on} observe that
failed runs suffer from vanishing gradients. 
Because all runs start from the pre-trained weights,
and the vanishing gradient makes the models update less intensively, 
this observation leads to the following assumption:
\emph{compared with successful runs,
failed runs bear lower representation instability}.
In this analysis,
we use this assumption to evaluate the concurrent validity of representation measures,
by testing whether they are able to distinguish failed from successful runs.
Because of this former observation only applies to hidden representations, 
in this analysis we exclude prediction measures. 

Specifically, we train our models using the same 20 random seeds
and keep the last checkpoint for each seed.
We adopt larger learning rates: 
$5 \times 10^{-5}$ for BERT and $3 \times 10^{-5}$ for RoBERTa,
because failed runs occur 
more frequently with larger learning rates \citep{mosbach2021on}.
For each group of models, we obtain 9--13 failed runs out of 20 runs.

We show our results for BERT on RTE in Figure~\ref{subfig:sf_runs} 
and observe similar patterns on other PLMs/datasets
(see Appendix~\ref{app:additional_results}). 
Linear-CKA and OP indeed indicate a lower instability in failed runs.
This observation is consistent with our expectation,
suggesting the concurrent validity of these two measures.
However, 
SVCCA fails to distinguish successful and failed runs 
based on the representations in the bottom layers, 
and therefore fails this test. 
One plausible explanation is that 
because lower layers of models tend to update less intensively 
during fine-tuning (the nature of back-propagation), 
they are likely to be more stable, 
and SVCCA may ignore these smaller differences.

\paragraph{Differences among test datasets}

Because we aim to quantify the instability of models themselves, 
one desideratum of a valid measure is to be independent of 
the specific data samples used to obtain the predictions and representations of models, 
as these data samples are not inherent components of these models. 
Concretely, we expect 
\emph{a valid measure to produce similar outputs for the same group of models 
when the instability scores are computed using different i.i.d. datasets}.

To evaluate the input invariance of the measures,
we create four sub-samples with half the test dataset size for each task, 
by uniformly sampling without replacement.
We then compute the instability scores using 
both prediction and representation measures on all samples, 
and show the results for BERT on MRPC in Figure~\ref{subfig:subsample} 
(we include results for RoBERTa and on MRPC/CoLA in Appendix~\ref{app:additional_results}).
We observe that the variance among different samples is very small, 
suggesting that all these measures show good concurrent validity in this test.\footnote{
    To investigate the impact of sample sizes,
    we also include the results of sample 10\% of the test datasets in Appendix~\ref{app:sample_size}.
    Despite that the differences between different sub-sampled datasets are larger compared to 50\%,
    we still observe good concurrent validity,
    especially in lower layers.
}

%% file: measure_comparison.tex
\begin{table*}[ht]
    \centering
    {\fontsize{8}{11}\selectfont
    \begin{tabular}{@{}ccccccccccccc@{}}
    \toprule
                                                & \multicolumn{4}{c}{RTE}                                                                                                                      & \multicolumn{4}{c}{MRPC}                                                                                                                     & \multicolumn{4}{c}{CoLA}                                                                                                \\
    \multicolumn{1}{c|}{}                    & Acc $\pm$ SD            & $\mathcal{I}_{\mathrm{JSD}}$ & $\mathcal{I}_{\mathrm{\kappa}}$ & \multicolumn{1}{c|}{$\mathcal{I}_{\mathrm{pwd}}$} & F1 $\pm$ SD             & $\mathcal{I}_{\mathrm{JSD}}$ & $\mathcal{I}_{\mathrm{\kappa}}$ & \multicolumn{1}{c|}{$\mathcal{I}_{\mathrm{pwd}}$} & MCC $\pm$ SD            & $\mathcal{I}_{\mathrm{JSD}}$ & $\mathcal{I}_{\mathrm{\kappa}}$ & $\mathcal{I}_{\mathrm{pwd}}$ \\ \midrule
    \multicolumn{1}{c|}{Standard}            & \textbf{71.3} $\pm$ 1.8 & 6.8                          & 13.9                            & \multicolumn{1}{c|}{13.8}                         & 89.3 $\pm$ 1.2          & 5.1                          & 9.1                             & \multicolumn{1}{c|}{9.1}                          & 64.5 $\pm$ 5.5          & 4.0                          & 7.9                             & 7.9                          \\
    \multicolumn{1}{c|}{Mixout}              & 71.2 $\pm$ 3.2          & 7.9                          & 15.5                            & \multicolumn{1}{c|}{15.4}                         & 89.6 $\pm$ \textbf{0.7} & 4.8                          & 8.9                             & \multicolumn{1}{c|}{8.8}                          & \textbf{67.1} $\pm$ 1.9 & 3.6                          & 7.1                             & 7.1                          \\
    \multicolumn{1}{c|}{LLRD}                & 69.2 $\pm$ 2.8          & 5.4                          & 13.8                            & \multicolumn{1}{c|}{13.7}                         & 89.5 $\pm$ 1.3          & 4.0                          & 8.2                             & \multicolumn{1}{c|}{8.2}                          & 63.9 $\pm$ 2.3          & \textbf{2.8}                 & \textbf{5.3}                    & \textbf{5.3}                 \\
    \multicolumn{1}{c|}{Re-init}             & 70.4 $\pm$ \textbf{1.4} & \textbf{4.7}                 & \textbf{10.1}                   & \multicolumn{1}{c|}{\textbf{10.0}}                & 89.9 $\pm$ 0.8          & \textbf{3.9}                 & \textbf{7.1}                    & \multicolumn{1}{c|}{\textbf{7.1}}                 & 64.2 $\pm$ 2.9          & 3.7                          & 6.8                             & 6.8                          \\
    \multicolumn{1}{c|}{$\mathsf{WD_{pre}}$} & 70.5 $\pm$ 5.6          & 7.4                          & 18.0                              & \multicolumn{1}{c|}{17.9}                         & \textbf{90.2} $\pm$ 1.2 & 4.7                          & 8.8                             & \multicolumn{1}{c|}{8.7}                          & 65.6 $\pm$ \textbf{1.8} & 3.6                          & 6.7                             & 6.7                          \\ \bottomrule
    \end{tabular}
    }
    \caption{Prediction instability scores of BERT 
    after using different IMMs.
    Higher values indicate higher instability for the instability measures (SD, $\mathcal{I}_{\mathrm{JSD}}$, $\mathcal{I}_{\mathrm{\kappa}}$, $\mathcal{I}_{\mathrm{pwd}}$).  For better readability, all values are multiplied by 100. }
    \label{tab:reassess_pred}
\end{table*}

\begin{figure*}
    \centering
    \begin{subfigure}[b]{0.32\textwidth}
        \centering
        \includegraphics[width=\textwidth]{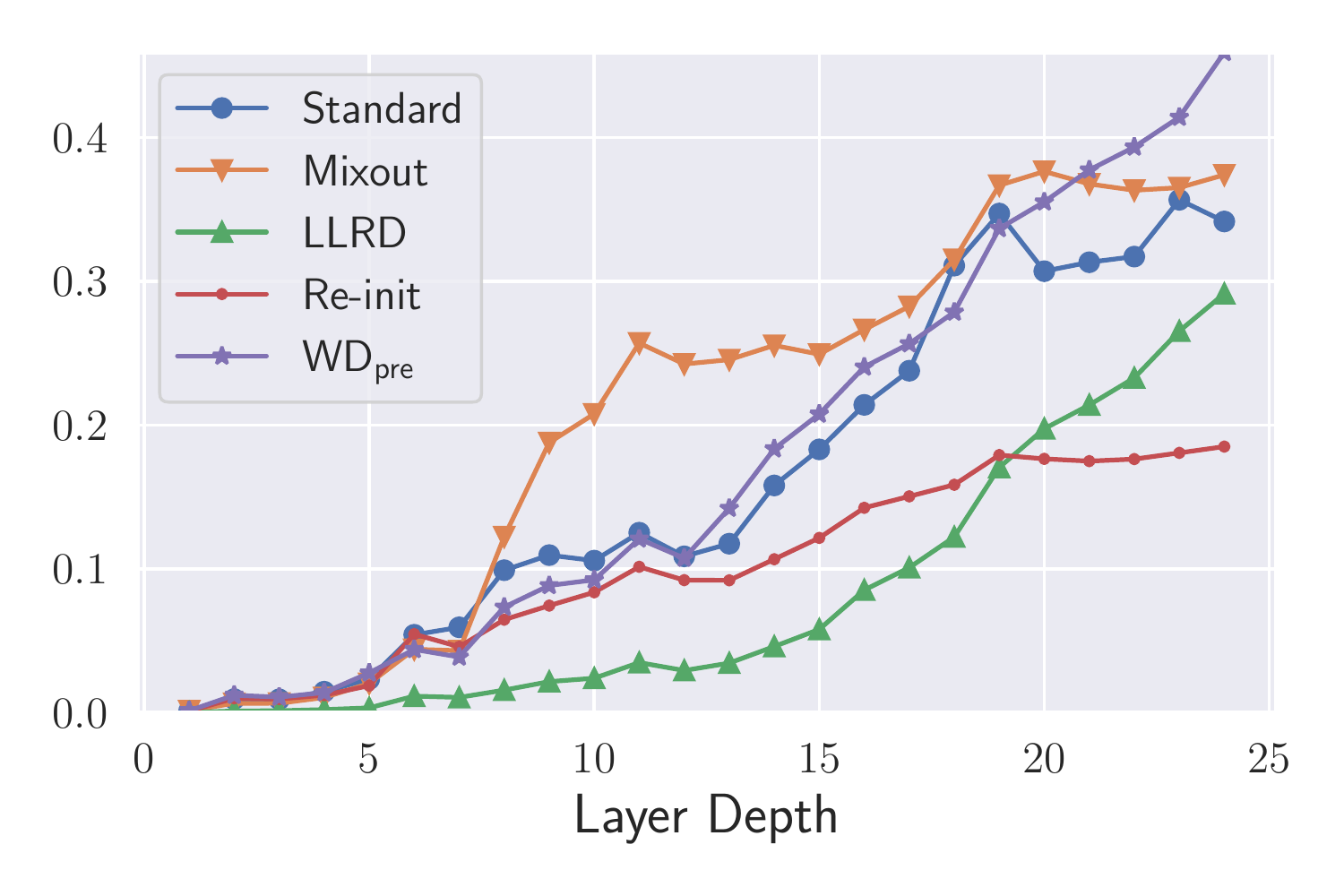}
        \caption{RTE}
        \label{subfig:reassess_rte}
    \end{subfigure}
    \hfill
    \begin{subfigure}[b]{0.32\textwidth}
        \centering
        \includegraphics[width=\textwidth]{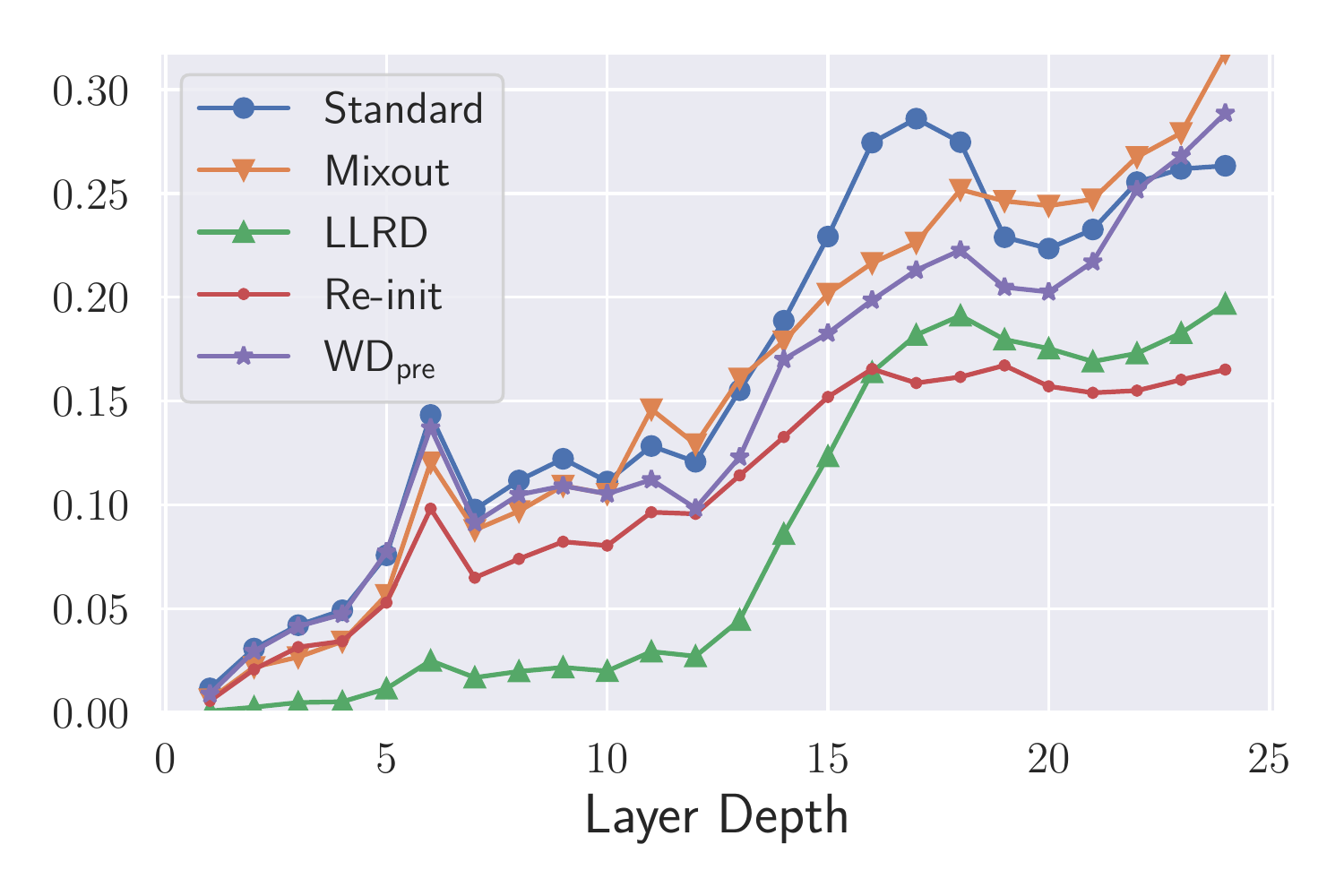}
        \caption{MRPC}
        \label{subfig:reassess_mrpc}
    \end{subfigure}
    \hfill
    \begin{subfigure}[b]{0.32\textwidth}
        \centering
        \includegraphics[width=\textwidth]{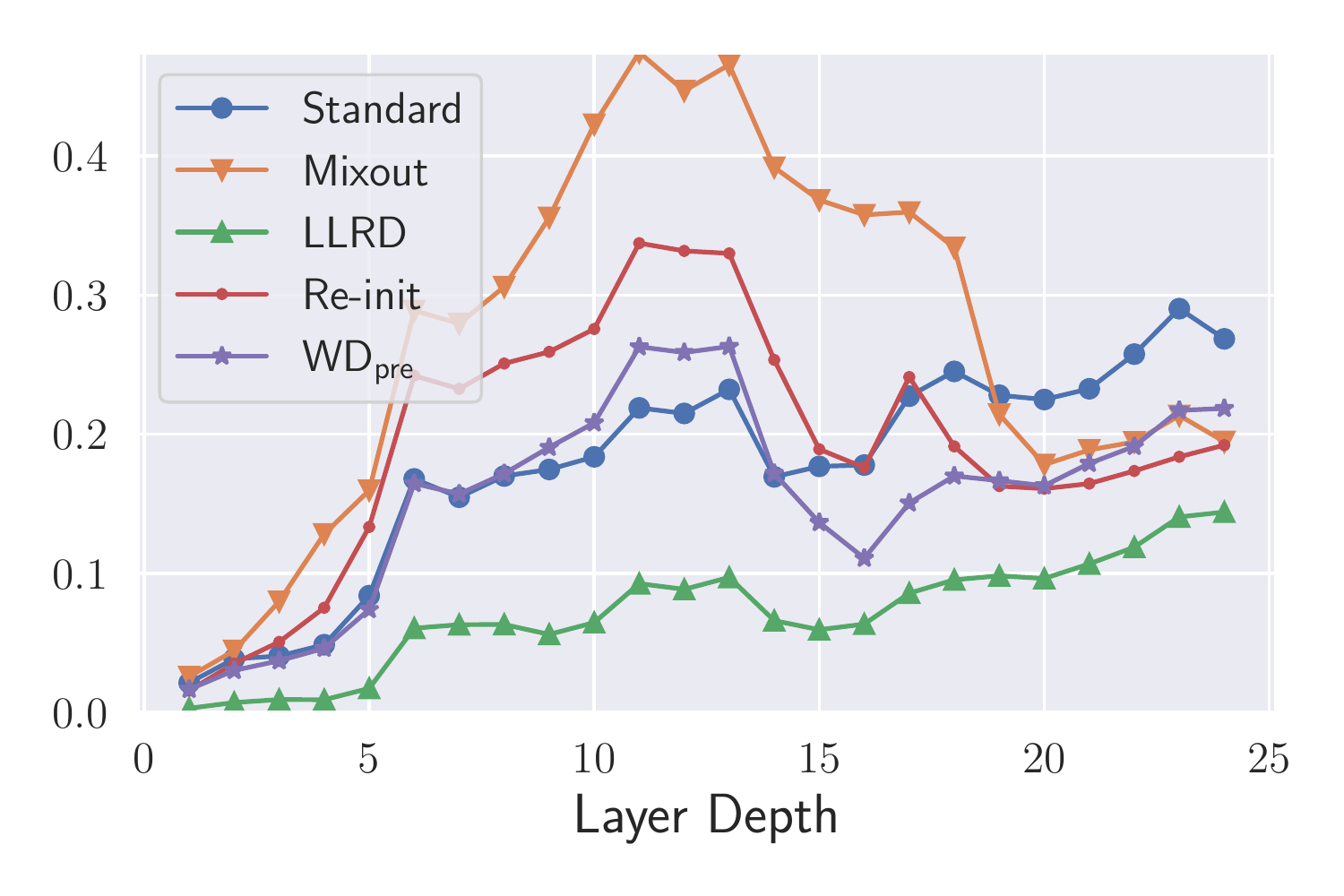}
        \caption{CoLA}
        \label{subfig:reassess_cola}
    \end{subfigure}
    \caption{
        $\mathcal{I}_{\mathrm{CKA}}$ scores of fine-tuned BERT models on all three datasets  
        after applying different IMMs. 
        We observe highly similar trends with RoBERTa and OP. Higher values indicate higher instability.
    }
    \label{fig:reassess_rep}
\end{figure*}

\begin{figure*}
    \centering
    \begin{subfigure}[b]{0.33\textwidth}
        \centering
        \includegraphics[width=\textwidth]{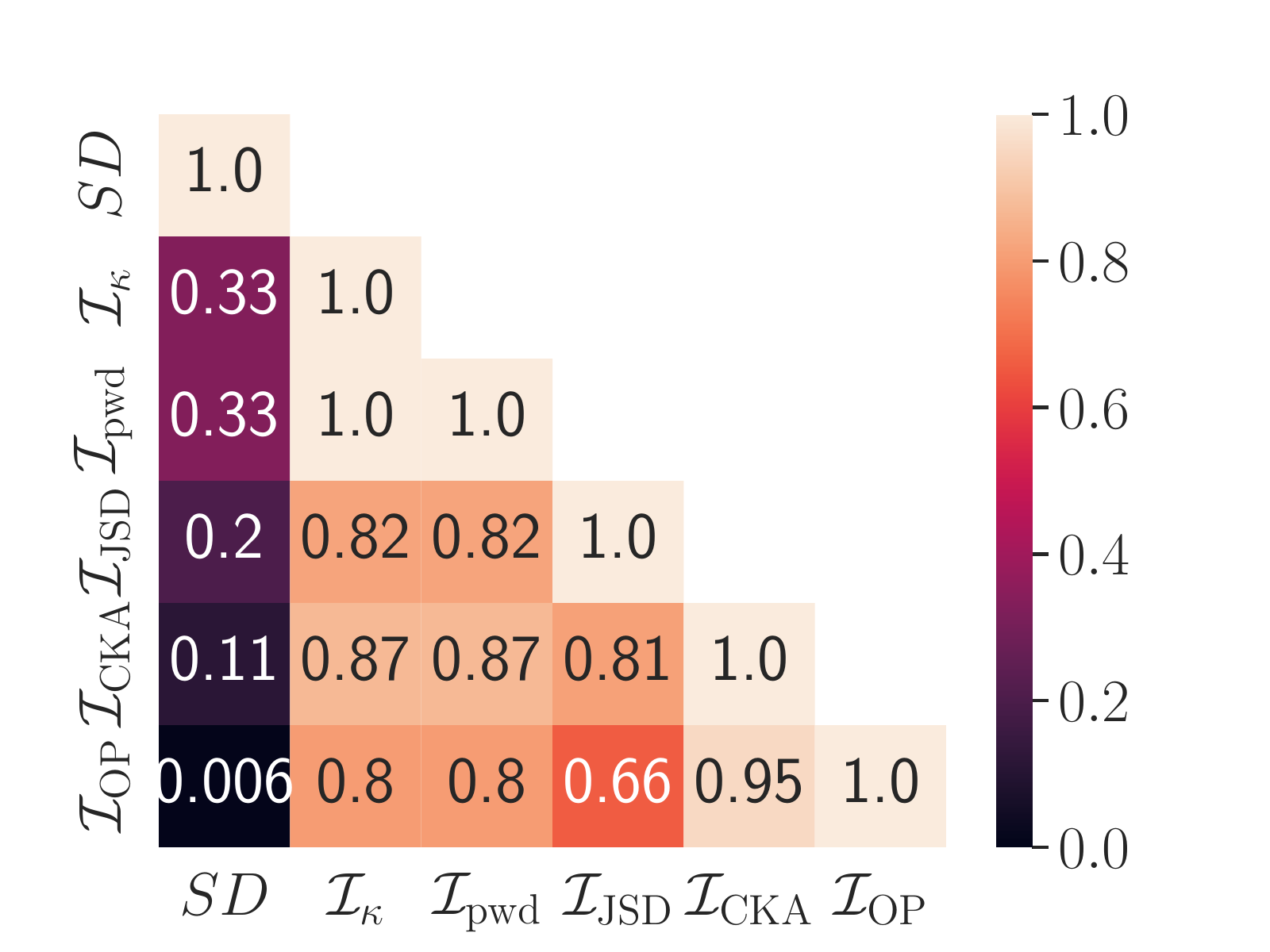}
        \caption{\emph{Standard} BERT on MRPC}
        \label{subfig:bs_mrpc_original_bert}
    \end{subfigure}
    \hfill
    \begin{subfigure}[b]{0.33\textwidth}
        \centering
        \includegraphics[width=\textwidth]{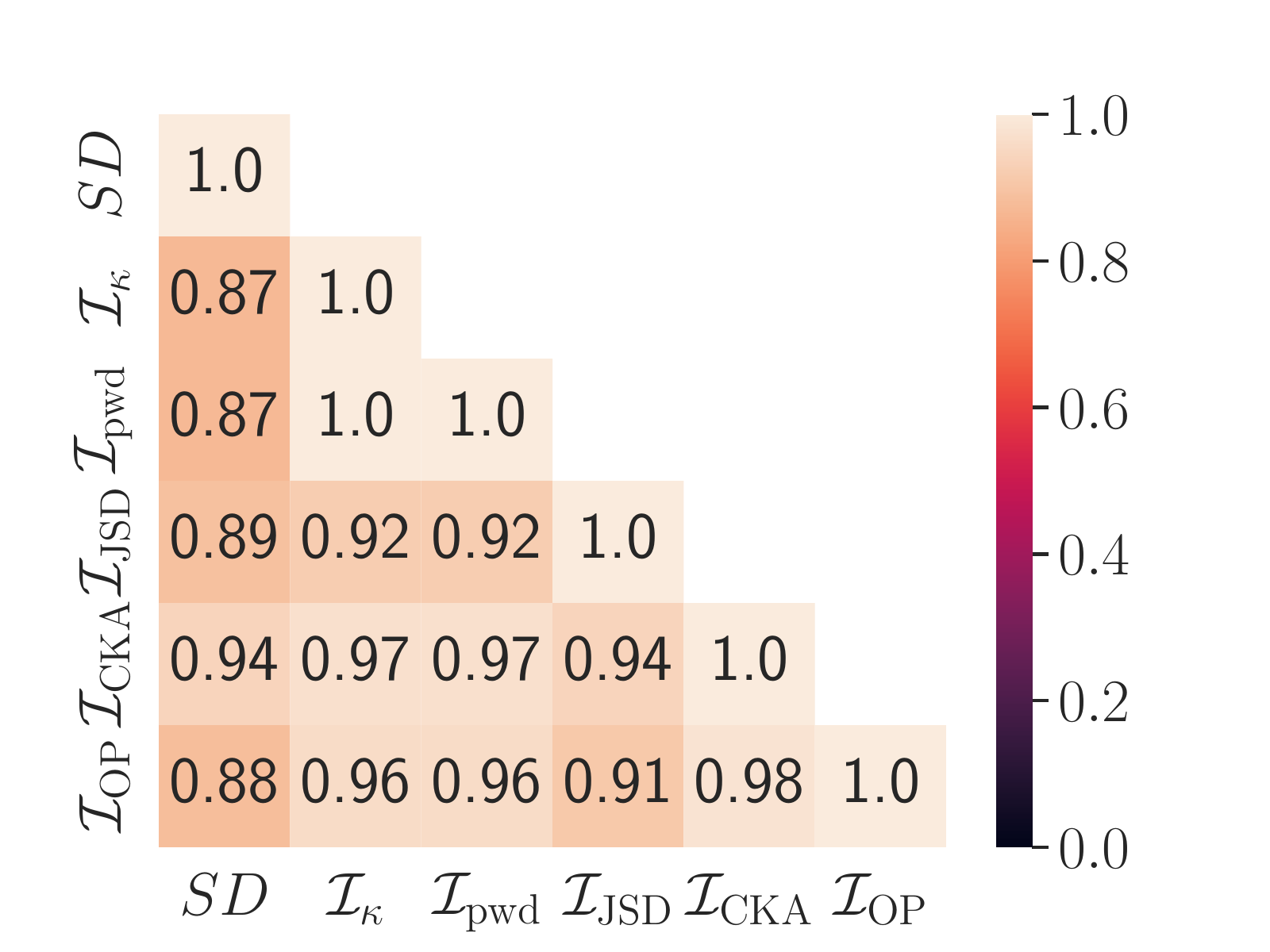}
        \caption{\emph{Standard} BERT on CoLA}
        \label{subfig:bs_cola_original_bert}
    \end{subfigure}
    \hfill
    \begin{subfigure}[b]{0.32\textwidth}
        \centering
        \includegraphics[width=\textwidth]{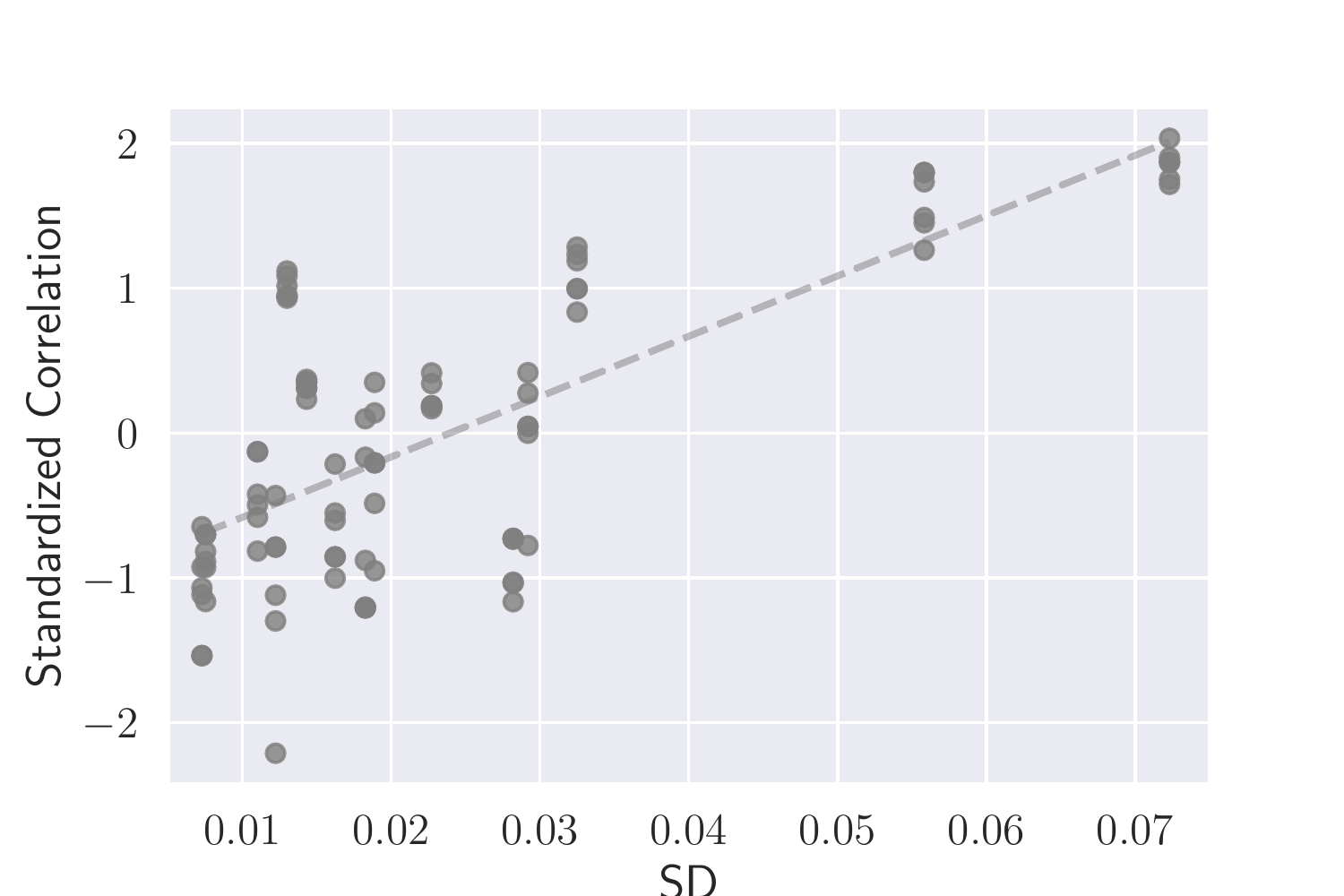}
        \caption{Correlations on BERT}
        \label{subfig:bs_correlations}
    \end{subfigure}
    \caption{
        Figures~\ref{subfig:bs_mrpc_original_bert} and \ref{subfig:bs_cola_original_bert}
        show the bootstrapping results 
        for BERT \emph{Standard} baseline on MRPC and CoLA.
        Figure~\ref{subfig:bs_correlations} shows the correlation between
        1) the average standardized correlation between each measure and other measures
        2) the corresponding SD value,
        for BERT on each dataset/IMM combination.
    }
    \label{fig:bs}
\end{figure*}

\section{The Need to Use Different Measures}\label{sec:relationship}

In \S\ref{sec:validity-of-representation-measures},
 all measures discussed in \S\ref{sec:instability_measures}
except for SVCCA showed good validity in our tests,
thus they are capable of measuring fine-tuning instability.
However, the following question remains: 
\emph{when do we need which instability measures}?
In this section, we explore this question via two studies. 
First, we reassess the effectiveness of existing IMMs 
by comparing the results when using different measures (\S\ref{subsec:reassess}).
Second, we further analyze the relationship between different measures 
using bootstrapping analyses (\S\ref{subsec:bootstrap}). 
We observe that measures at different granularity levels show better consistency
when the models are less stable and vice versa.
Moreover, based on our findings, 
we provide two concrete suggestions for selecting instability measures 
for future studies (\S\ref{subsec:implications}). 

\subsection{Reassessing IMMs}\label{subsec:reassess}

To study the relationships between different measures, 
we reassess existing IMMs from \S\ref{subsec:instability-mitigation-methods} 
and compare their instability scores. 
We include all measures discussed in \S\ref{sec:instability_measures}, 
except for SVCCA because it did not show good validity in our tests.

\paragraph{Experimental setup}
For each IMM, 
we train 10 models using different random seeds,
with the same hyper-parameters in \S\ref{sec:experimental-setup}. 
We also adopt the IMM hyper-parameters from \citet{zhang2021revisiting}:
Mixout $p=0.1$, $\mathsf{{WD}_{pre}} \lambda=0.01$, 
LLRD $\eta=0.95$, and Re-init $k=5$.
We compare the IMMs against a baseline (\emph{Standard}),
which does not use any IMM.\footnote{
    Note that our \emph{Standard} baseline is different from the one in \citet{zhang2021revisiting}. 
    Ours uses a longer training time (five epochs) and a standard weight decay regularization, 
    because these two choices can be seen as hyper-parameter selections. 
    As observed by \citet{mosbach2021on}, 
    these modifications make our baseline stronger. 
} 

\paragraph{Measures of different granularity levels are not always consistent with each other}
We show the results of prediction measures 
and $\mathcal{I}_{\mathrm{CKA}}$ on BERT 
in Table~\ref{tab:reassess_pred} and Figure~\ref{fig:reassess_rep}
(other results show similar trends, 
see Appendix~\ref{app:additional_results}). 
We observe that 
different measures are not always consistent with each other.
For example, when using BERT on MRPC (Table~\ref{tab:reassess_pred}), 
SD shows Mixout to be the most stable training scheme. 
However,
the other three prediction measures
and $\mathcal{I}_{\mathrm{CKA}}$ (the top layer, Figure~\ref{subfig:reassess_mrpc})
rank Mixout to be the (second) least stable one.

To better quantify the inconsistency, 
we calculate Kendall's $\tau$ between 
the rankings of IMMs based on different measures, 
on each dataset/PLM. 
We include the full results in Appendix~\ref{app:additional_results} 
and make two observations. 
First, measures of similar granularity level 
tend to be consistent with each other.  
For example, 
$\mathcal{I}_{\mathrm{CKA}} \sim \mathcal{I}_{\mathrm{OP}}$ 
(both representation measures)
and 
$\mathcal{I}_{\mathrm{pwd}} \sim \mathcal{I}_{\mathrm{\kappa}}$ 
(both based on discrete predictions)
show good consistency (i.e. $\tau \geq 0.8$) 
on each combination of models and datasets. 
Also, $\mathcal{I}_{\mathrm{pwd}}$ and $\mathcal{I}_{\mathrm{\kappa}}$ 
show better consistency with $\mathcal{I}_{\mathrm{jsd}}$ ($\tau \geq 0.6$)
than with SD ($\tau = -0.2$ for BERT on MRPC). 
Second, 
the consistency among measures differs across datasets and models. 
For example, all measures correlate well for BERT on RTE,
with a minimum $\tau \approx 0.6$. 
In contrast, the correlations derived from MRPC are much smaller, 
with close-to-zero $\tau$ values between SD and other measures. 

\paragraph{Most IMMs are not always effective}
Our results also show that 
most IMMs are not always effective:
they sometimes fail to improve stability compared to the \emph{Standard} baseline,
which is consistent with the observations of \citet{zhang2021revisiting}.
In fact, Re-init is the only IMM  
that consistently improves over \emph{Standard} according to all measures. 
Also, for BERT on RTE, 
\emph{Standard} is the third most stable training method 
according to all prediction (Table~\ref{tab:reassess_pred})
and representation (Figure~\ref{fig:reassess_rep}) measures.
Generally, 
models trained with $\mathsf{WD_{pre}}$ and Mixout are less stable 
compared to models trained with LLRD and Re-init.
Because both $\mathsf{WD_{pre}}$ and Mixout 
aim to stabilize fine-tuning by resolving catastrophic forgetting, 
these results suggest that
catastrophic forgetting may not be the actual or sole cause of instability, 
which is consistent with the observations of \citet{mosbach2021on}.

\subsection{Bootstrapping Analyses}\label{subsec:bootstrap}

In \S\ref{subsec:reassess}, 
we computed Kendall's $\tau$ between the rankings of different IMMs obtained using different instability measures. 
However, because we only have five groups of models 
(i.e. each group consists of 10 models trained with the same IMM/\emph{Standard} baseline
but different random seeds), 
the results we obtain may be less accurate. 
To mitigate this issue, in this section 
we focus on generating more groups for a specific IMM-dataset combination. 
Unfortunately, generating more groups 
is extremely expensive, as each group consists of 
10 models trained with different random seeds.\footnote{
For example, the minimum recommended sample size for Pearson's $r=0.5$ is 99
\citep{bonett2000sample}. 
In our case, this means training 990 models (i.e. 99 groups). }

To avoid the high training cost,
we instead use \textbf{bootstrapping} to \emph{generate more groups of models}.
Concretely, 
for each group of 10 models
(i.e. 10 different random seeds),
we sample 10 models with replacement for 1000 times 
to obtain 1000 groups of models.
We then compute the Pearson's $r$ between each pair of measures using these groups.
We apply the representation measures on the topmost layer
to make the results more comparable with the prediction measures. 

We show the results for the BERT \emph{Standard} baseline on MRPC 
and CoLA in Figure~\ref{fig:bs}
and observe similar trends on other datasets/models 
(see Appendix~\ref{app:additional_results}).
We make two observation. 
First, consistent with \S\ref{subsec:reassess}, 
we observe that measures at closer granularity levels 
have higher correlations with each other. 
For example, 
SD has correlations of a decreasing strength with other measures on MRPC: 
from the most similar and coarse-grained Fleiss' Kappa and pairwise disagreement, 
to the pairwise JSD in between, 
and finally to the furthest and the most fine-grained representation measures OP and Linear-CKA. 
Also, the correlations between the two representation measures
($\mathcal{I}_{\mathrm{CKA}}$ and $\mathcal{I}_{\mathrm{OP}}$ )
are much higher than that between them and other measures.
Second, we also observe that correlations obtained from
different combinations of dataset/IMM 
are different from each other, which is expected. 

The second observation points to another question:
when will different measures be more consistent with each other?
Intuitively,
\emph{they will be more consistent when the differences between models are large}.
In this case, 
both coarse-grained and fine-grained measures 
can detect the instability. 
In contrast, when the differences between models are small,
only fine-grained measures can capture these nuances.
In other words,
\emph{instability measures are more consistent
when the models are less stable, and vice versa.}

To quantitatively check this intuition,
on each PLM
(i.e.~BERT and RoBERTa),
using the bootstrapping results on each dataset and IMM,
we compute the Pearson's $r$ between
1) the average correlations between each measure and other measures
and 2) SD values.\footnote{
Although sharing the same range -1--1,
correlations between different measures usually have different scales of values.
In other words, some measures are more consistent with other measures,
and thus have larger correlations.
To balance the weights of different measures,
we standardize the correlations for each measure
according to the its average correlations with other measures on different datasets/IMMs.
} 
We observe strong correlations on
both BERT ($r = 0.734$, Figure~\ref{subfig:bs_correlations})
and RoBERTa ($r = 0.653$, Appendix~\ref{app:additional_results}),
confirming our intuition.

\subsection{Implications}\label{subsec:implications}

In \S\ref{subsec:reassess} and \S\ref{subsec:bootstrap}, 
we investigated the consistency and differences 
between different instability measures.
Based on our observations, 
we provide two practical suggestions for future studies. 

First, we observed that measures are not always consistent with each other, 
despite their good validity in \S\ref{sec:validity-of-representation-measures}. 
This observation suggests that 
different measures focus on different aspects of instability 
and therefore should be used together in future studies. 
Moreover, we observed that 
different measures tend to be less consistent with each other 
when the models themselves are more stable.
This observation further demonstrates the necessity of 
adopting multiple measures 
when the instability assessed by one of the measures is low, 
and that using any measure alone may produce inaccurate conclusions. 

Second, 
we observed measures at similar granularity levels
to be more consistent.
One can therefore start with SD,  
and sequentially add more fine-grained measures 
when previous measures indicate low stability.
Because computing fine-grained instability is often slow, 
 only 
one prediction measure and one representation measure  can be used
when limited computational resources are available.

%% file: app.tex
\section{Experimental Setup}\label{app:experimental_setup}

\paragraph{Running Environment}
All models are trained using a single NVIDIA RTX 6000 graphics card,
with Python 3.7, PyTorch 1.10.1 \citep{pytorch}, 
Hugging Face Transformers 4.14.1 \citep{wolf-etal-2020-transformers}, 
and CUDA 10.2.
The total training time is approximately 70 GPU hours.
We calculate the results of the instability measures on Intel Xeon E5-2699 CPUs,
with Python 3.7 and Numpy 1.19.5 \citep{numpy}, taking approximately 24 CPU hours.

\paragraph{Validation/Test Split}

\begin{table}[h]
    \centering
    \small
    \begin{tabular}{@{}ccccccc@{}}
        \toprule
                 & \multicolumn{2}{l}{RTE} & \multicolumn{2}{l}{MRPC} & \multicolumn{2}{l}{CoLA} \\
                 & Dev        & Test       & Dev        & Test        & Dev        & Test        \\
        \midrule
        Positive & 59         & 72         & 142        & 137         & 375        & 346         \\
        Negative & 79         & 67         & 62         & 67          & 146        & 176         \\
        \bottomrule
    \end{tabular}
    \caption{Statistics of the test-validation split}
    \label{tab:dev_test_split}
\end{table}

The GLUE benchmark is one of the most popular benchmarks in NLP research ~\citep{wang-etal-2018-glue},
which consists of 11 different tasks, 
including RTE, MRPC, and CoLA. 
We download and process the datasets using Hugging Face Datasets \citep{lhoest-etal-2021-datasets}.
Following \citet{zhang2021revisiting},
we split the original validation dataset into two parts of (almost) equal sizes,
because we have no access to the test data.
We then use one part as the new validation data
to select checkpoints with the best performance,
and we use the other part as the new test data
to compute all instability measures.
We provide the statistics of the splits in Table~\ref{tab:dev_test_split}.

\section{Details of Fleiss' Kappa}\label{app:feliss_kappa}

Consider a $k$-class classification task, $m$ different models, and a test dataset size of $n$.
We denote the number of models which predict the $i$-th data point as the $j$-th class as $x_{ij}$.
Clearly, we have $\sum_{j=1}^{k} x_{i j}=m$, 
because each of the $m$ models will make a prediction on $x_i$.

We estimate the proportion of \emph{pairs of models}
that agree on the $i$-th data point by
\begin{IEEEeqnarray}{rCl}
    p_{i}=\frac{\sum_{j=1}^{k} C\left(x_{i j}, 2\right)}{C(m, 2)}
    =\frac{\sum_{j=1}^{k} x_{i j}^{2}-m}{m(m-1)}, \notag
\end{IEEEeqnarray}
where $C$ means the combination.
We can then calculate the mean value of $p_i$ as
\begin{IEEEeqnarray}{ll}
    p_{a}&=\frac{1}{n} \sum_{i=1}^{n} p_{i}\notag\\
    &= \frac{1}{m n(m-1)}\left[\sum_{i=1}^{n} \sum_{j=1}^{k} x_{i j}^{2}-m n\right].\notag
\end{IEEEeqnarray}
Moreover, we estimate the error term as
\begin{IEEEeqnarray}{ll}
    p_{\epsilon}=\sum_{j=1}^{k} (\frac{1}{n m} \sum_{i=1}^{n} x_{i j})^2.\notag
\end{IEEEeqnarray}
After obtaining $p_a$ and $p_{\epsilon}$, we can calculate Fleiss' Kappa as
\begin{IEEEeqnarray}{ll}
    \kappa=\frac{p_{a}-p_{\epsilon}}{1-p_{\epsilon}}. \notag
\end{IEEEeqnarray}

\begin{table}[ht]
    \small
    \centering
    \begin{tabular}{@{}ccc@{}}
    \toprule
    & BERT & RoBERTa \\ \midrule
    \begin{tabular}[c]{@{}c@{}}$\mathcal{I}_{\mathrm{CKA}} \sim \mathcal{I}_{\mathrm{OP}}$\end{tabular}    & 0.78  & 0.94    \\
    \begin{tabular}[c]{@{}c@{}}$\mathcal{I}_{\mathrm{CKA}} \sim \mathcal{I}_{\mathrm{SVCCA}}$\end{tabular} & -0.24 & 0.41    \\
    \begin{tabular}[c]{@{}c@{}}$\mathcal{I}_{\mathrm{OP}} \sim \mathcal{I}_{\mathrm{SVCCA}}$\end{tabular}  & 0.14  & 0.51    \\

    \midrule

    \begin{tabular}[c]{@{}c@{}} Acc $\pm$ SD                    \end{tabular}  & 92.6 $\pm$ 0.8 & 94.4 $\pm$ 0.9    \\
    \begin{tabular}[c]{@{}c@{}} $\mathcal{I}_{\mathrm{JSD}}$    \end{tabular}  & 2.3            & 2.5    \\
    \begin{tabular}[c]{@{}c@{}} $\mathcal{I}_{\mathrm{\kappa}}$ \end{tabular}  & 4.5            & 4.3    \\
    \begin{tabular}[c]{@{}c@{}} $\mathcal{I}_{\mathrm{pwd}}$    \end{tabular}  & 4.5            & 4.3    \\
    \bottomrule
    \end{tabular}
    \caption{Correlations between representation measures,
        and instability scores computed by different prediction measures, on SST-2.}
    \label{tab:sst2_rm_correlation}
\end{table}

\begin{figure*}[ht]
    \centering
    \begin{subfigure}[b]{0.45\textwidth}
        \centering
        \includegraphics[width=\textwidth]{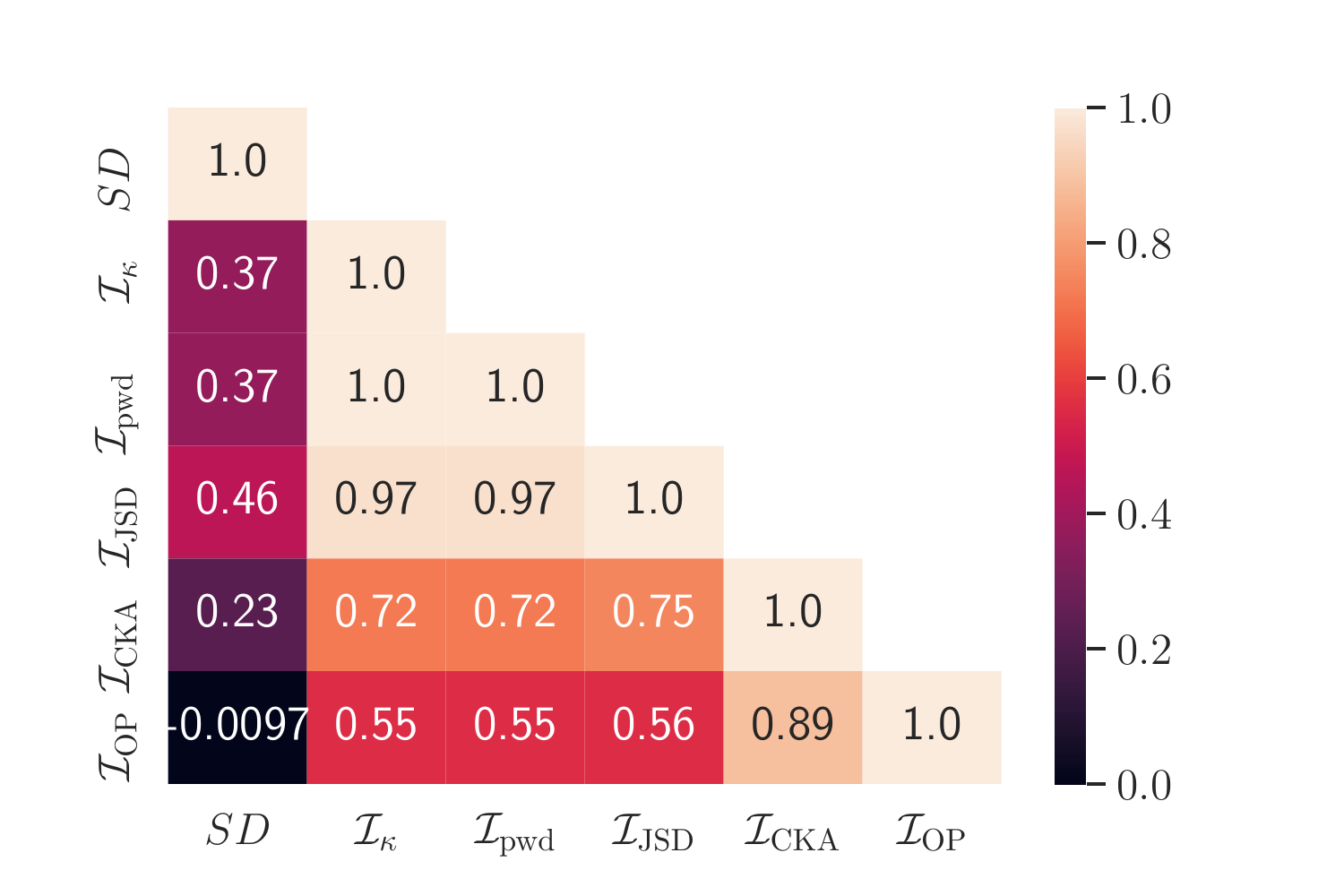}
        \caption{BERT}
        \label{subfig:app_sst2_bs_bert}
    \end{subfigure}
    \hfill
    \begin{subfigure}[b]{0.45\textwidth}
        \centering
        \includegraphics[width=\textwidth]{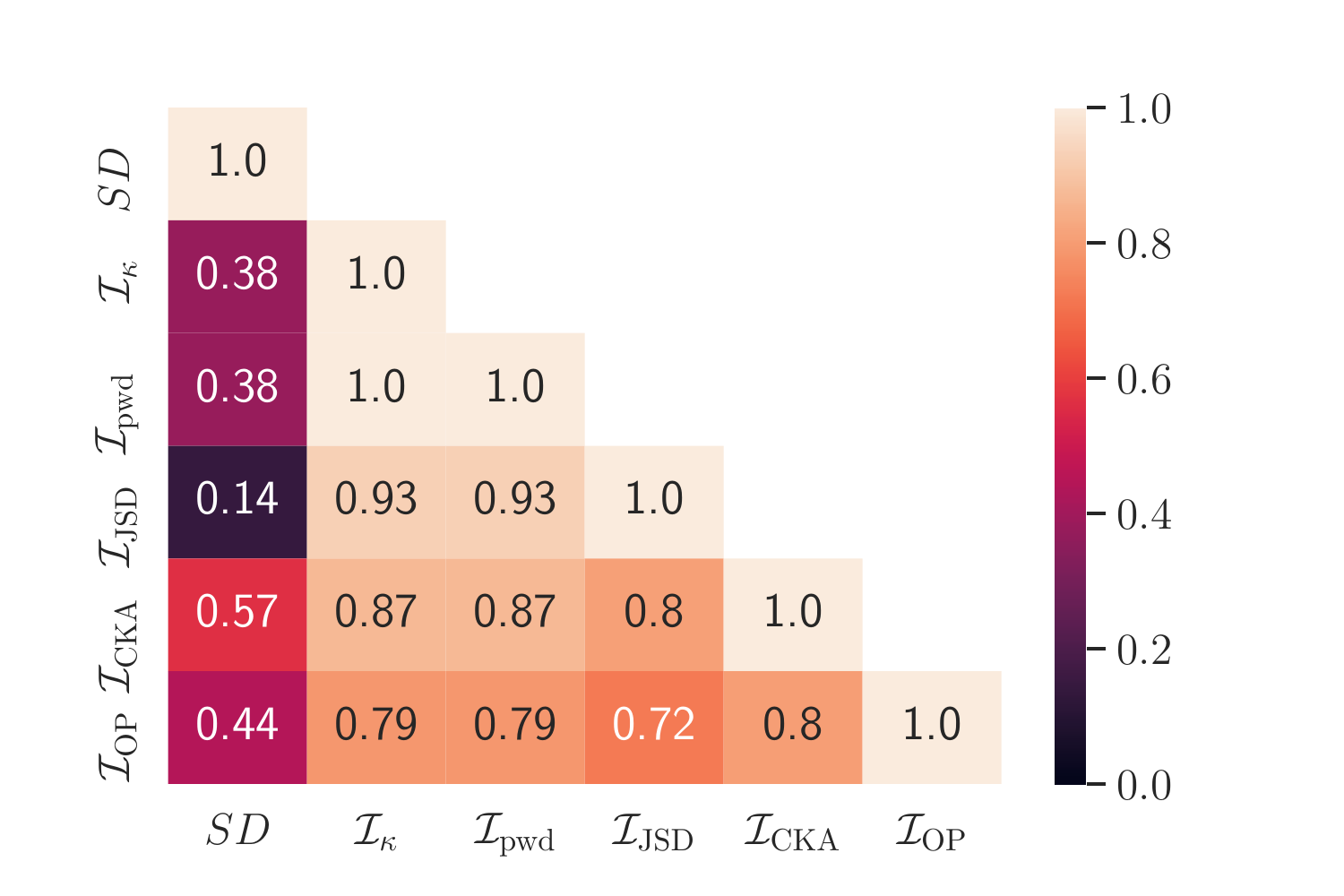}
        \caption{RoBERTa}
        \label{subfig:app_sst2_bs_roberta}
    \end{subfigure}
    \caption{Bootstrapping analyses on SST-2.}
    \label{fig:app_sst2_bs}
\end{figure*}

\begin{figure*}[ht]
    \centering
    \begin{subfigure}[b]{0.32\textwidth}
        \centering
        \includegraphics[width=\textwidth]{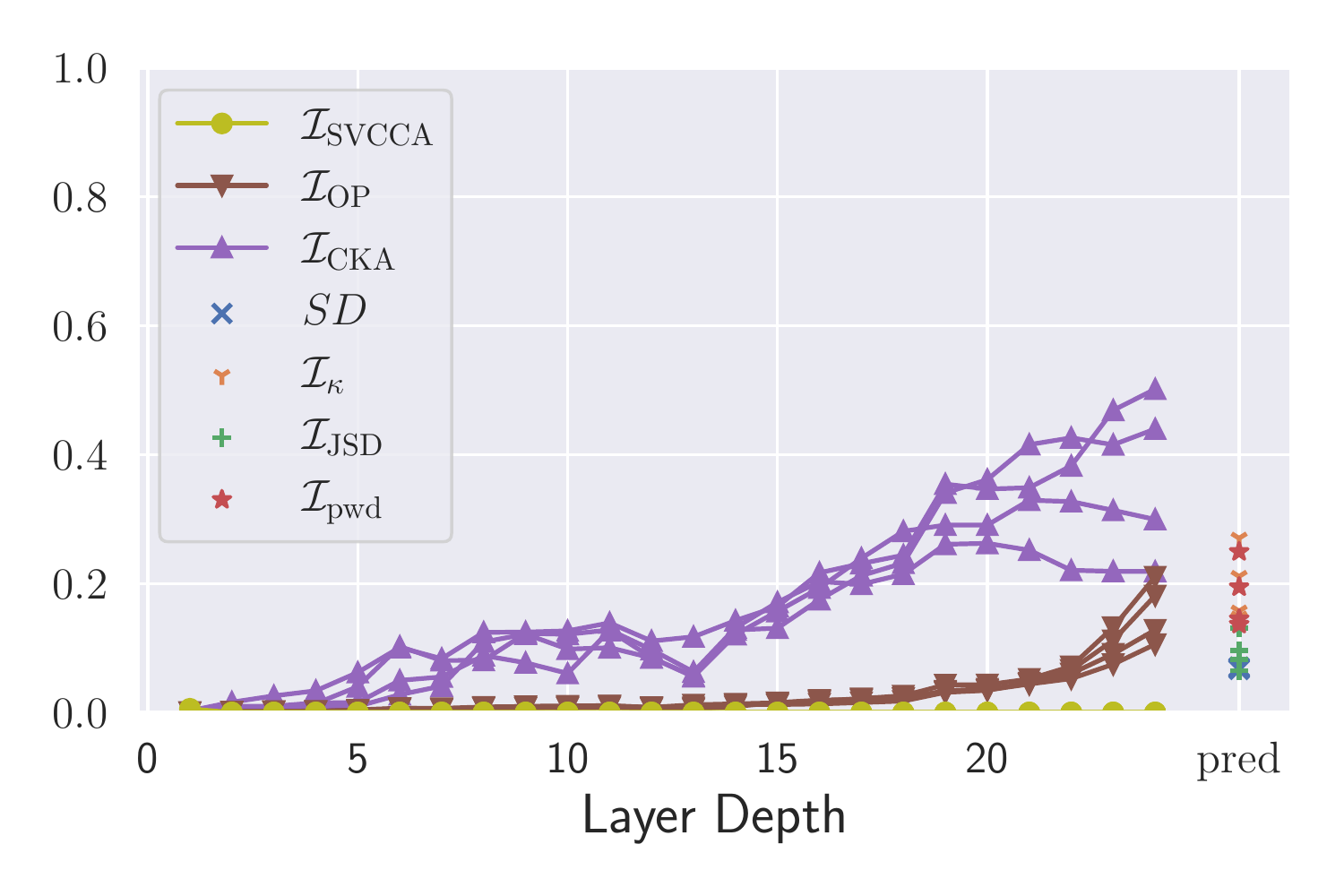}
        \caption{RTE}
        \label{subfig:subsample_bert_rte_1}
    \end{subfigure}
    \hfill
    \begin{subfigure}[b]{0.32\textwidth}
        \centering
        \includegraphics[width=\textwidth]{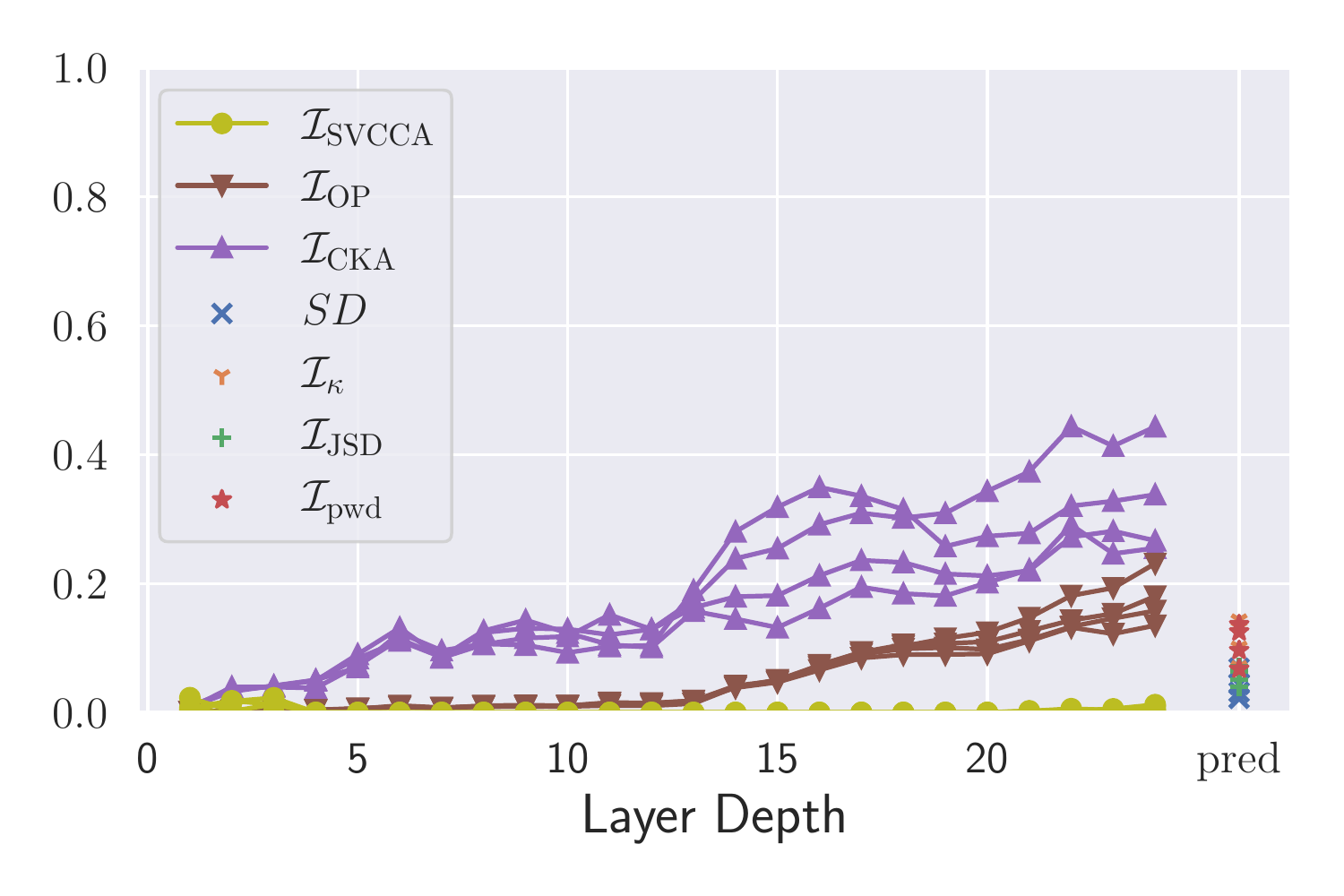}
        \caption{MRPC}
        \label{subfig:subsample_bert_mrpc_1}
    \end{subfigure}
    \hfill
    \begin{subfigure}[b]{0.32\textwidth}
        \centering
        \includegraphics[width=\textwidth]{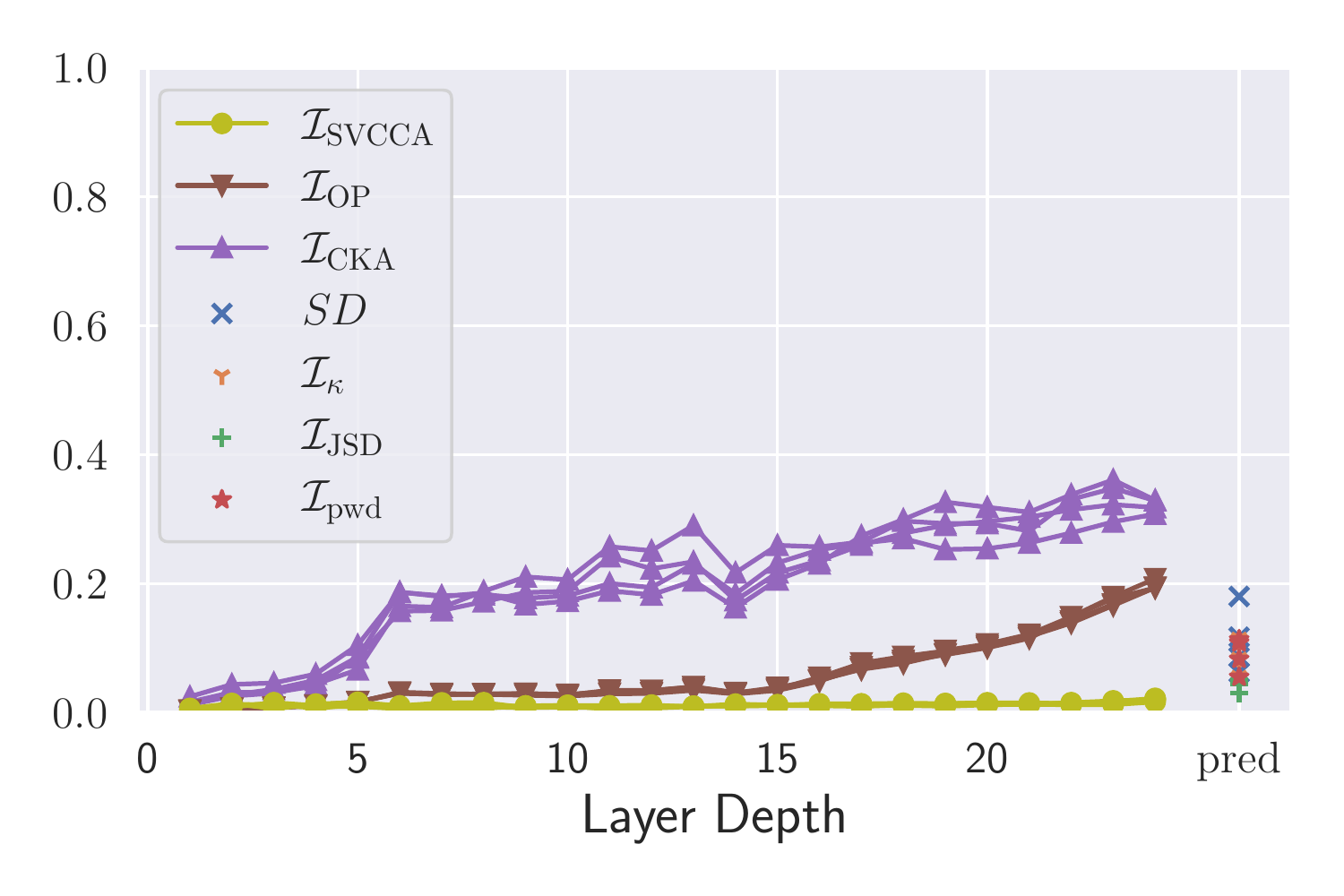}
        \caption{CoLA}
        \label{subfig:subsample_bert_cola_1}
    \end{subfigure}
    \caption{Consistency among sub-samples on BERT, sample rate 0.1.}
    \label{fig:subsample_bert_1}
    \vfill
    \begin{subfigure}[b]{0.32\textwidth}
        \centering
        \includegraphics[width=\textwidth]{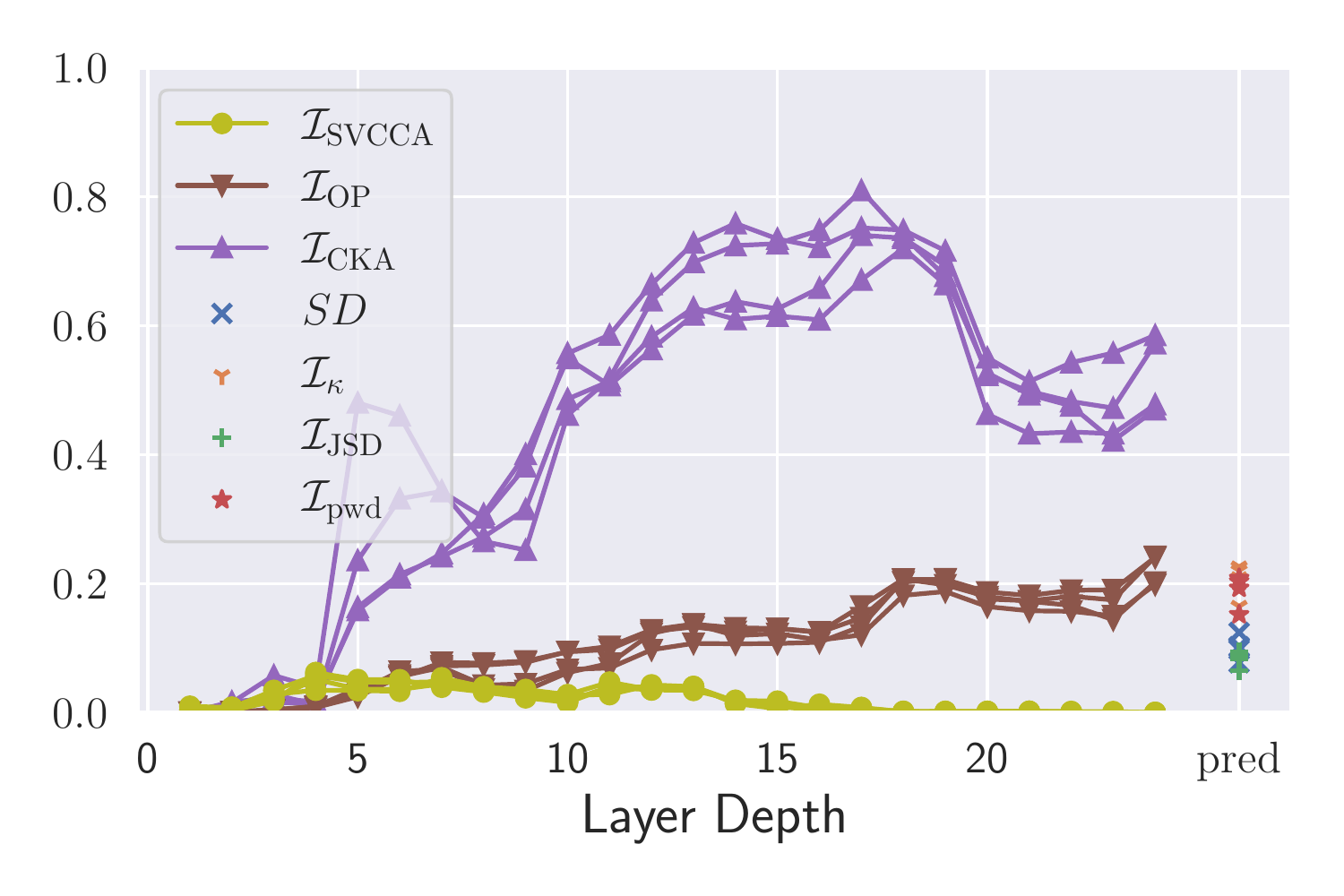}
        \caption{RTE}
        \label{subfig:subsample_roberta_rte_1}
    \end{subfigure}
    \hfill
    \begin{subfigure}[b]{0.32\textwidth}
        \centering
        \includegraphics[width=\textwidth]{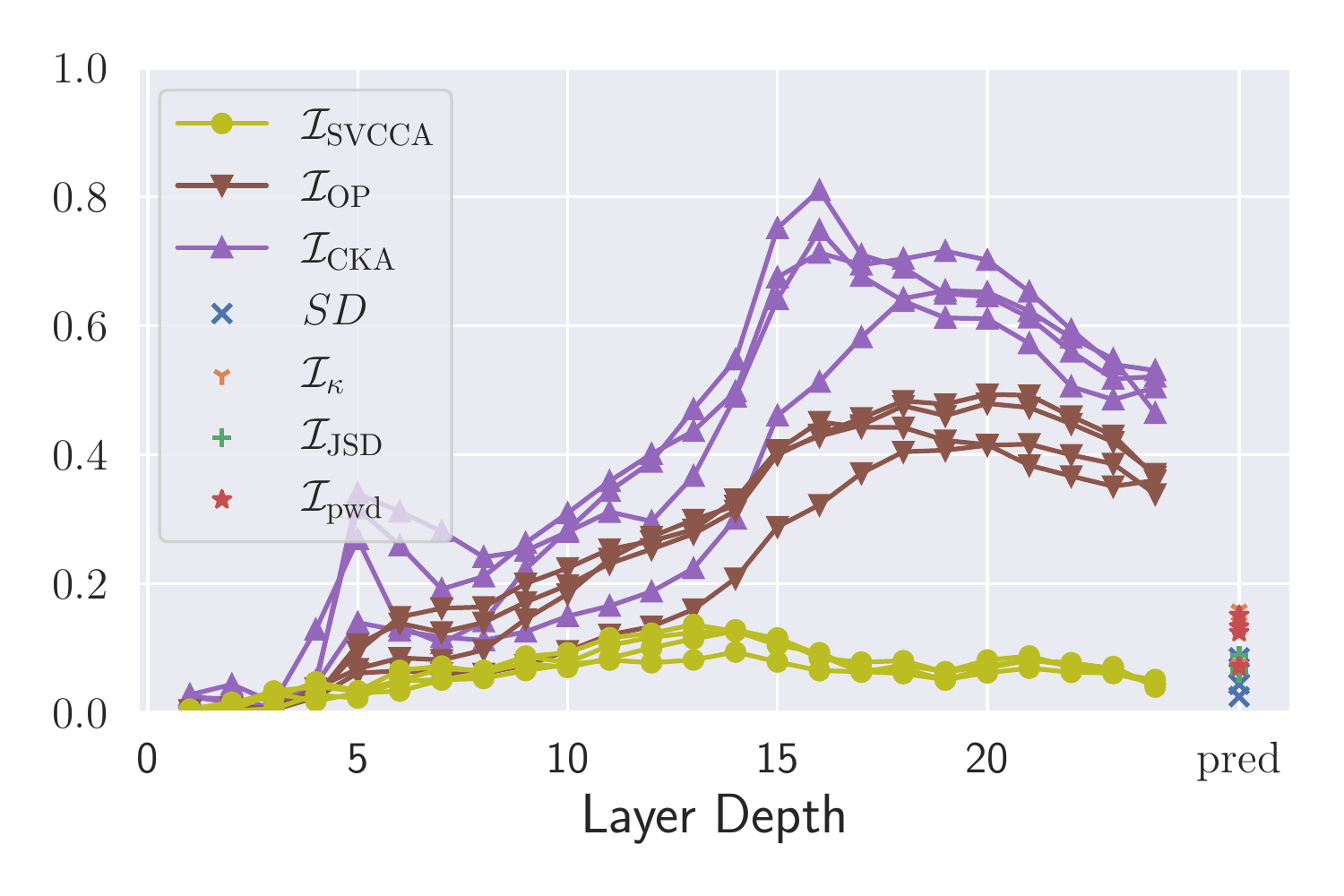}
        \caption{MRPC}
        \label{subfig:subsample_roberta_mrpc_1}
    \end{subfigure}
    \hfill
    \begin{subfigure}[b]{0.32\textwidth}
        \centering
        \includegraphics[width=\textwidth]{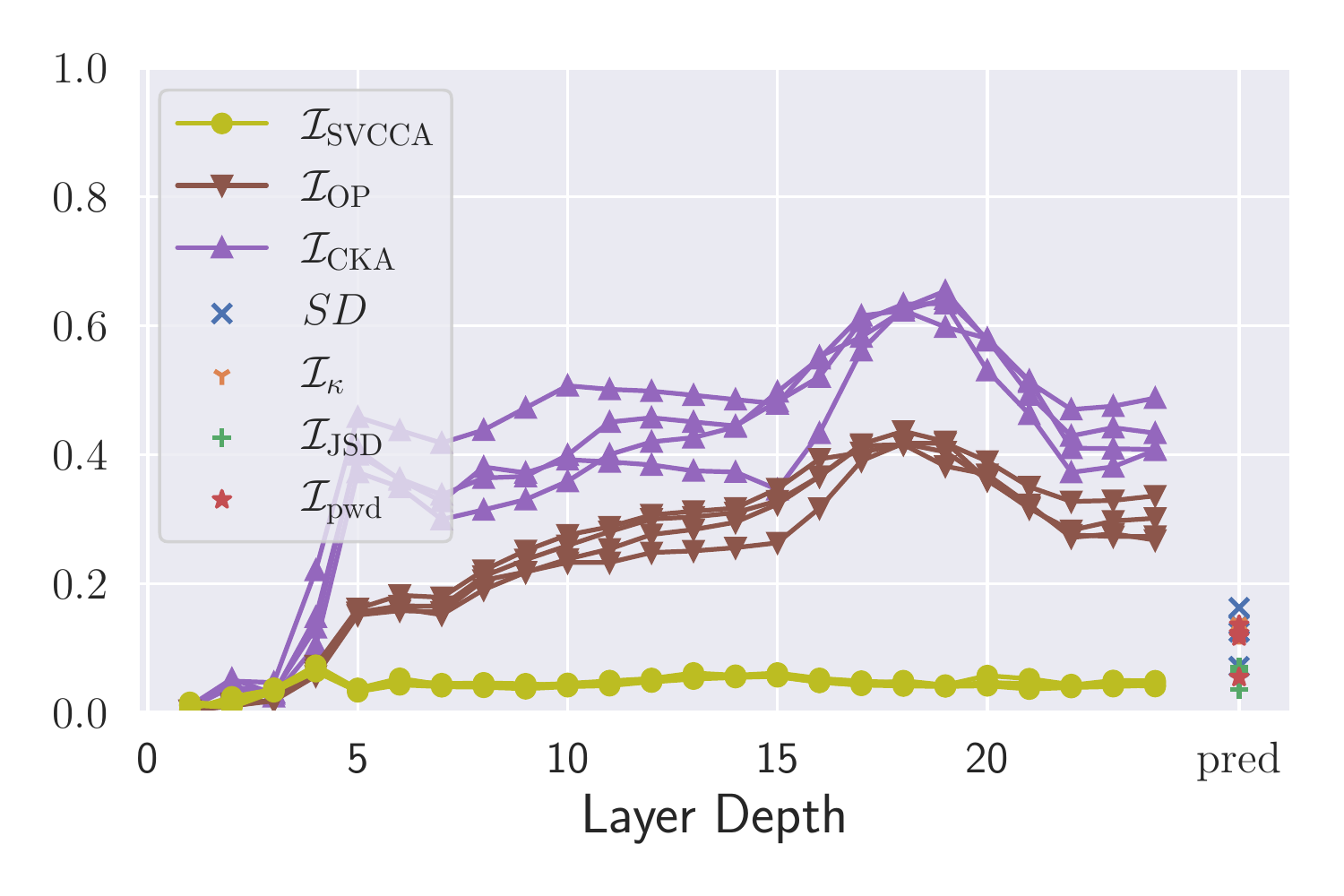}
        \caption{CoLA}
        \label{subfig:subsample_roberta_cola_1}
    \end{subfigure}
    \caption{Consistency among sub-samples on RoBERTa, sample rate 0.1.}
    \label{fig:subsample_roberta_1}
\end{figure*}

\section{Impact of Dataset Size}\label{app:sst2}

To better understand the impact of using only small train datasets,
we perform a preliminary study on SST-2~\citep{sst2},
which consists of 67,000 training samples,
around eight times larger than the size of CoLA (8,000).
We use the same hyper-parameter settings as in Section~\ref{sec:experimental-setup},
namely a 16 batch size, a 0.1 Dropout rate, a $2 \times 10^{-5}$ learning rate,
using 20 different random seeds and de-biased Adam, without IMMs.

We computed the instability scores using different prediction measures.
We also computed the correlations between representation measures
and performed bootstrapping analyses.
We show the results in
Table~\ref{tab:sst2_rm_correlation} and Figure~\ref{fig:app_sst2_bs}.
We make three observations.
First, as expected, we observe lower instability from models trained on SST-2
compared with models trained on the three small datasets we used in the main text.
Second, consistent with our observations in Section~\ref{sec:relationship},
we observe that the correlations between different measures on SST-2 are lower,
because the models are more stable.
Third, also consistent with our observations in Section~\ref{sec:relationship},
Figure~\ref{fig:app_sst2_bs} shows that
measures at similar granularity levels are more consistent with each other.
Our results on SST-2 suggest that
our previous observations are generalizable to larger datasets.

\section{Impact of Subsample Size}\label{app:sample_size}

To investigate the impact of sample sizes
regarding the differences among different i.i.d. datasets,
we also experimented with sampling only 10\% of the test samples.
We show the results in Figure~\ref{fig:subsample_bert_1}--\ref{fig:subsample_roberta_1}.
Sampling only 10\% of the test samples does bring larger variances (compared with sampling 50\%),
but results on different samples are mostly still consistent,
especially in the lower layers.

\section{Additional Results}\label{app:additional_results}
\begin{figure*}
    \centering
    \begin{subfigure}[b]{0.45\textwidth}
        \centering
        \includegraphics[width=\textwidth]{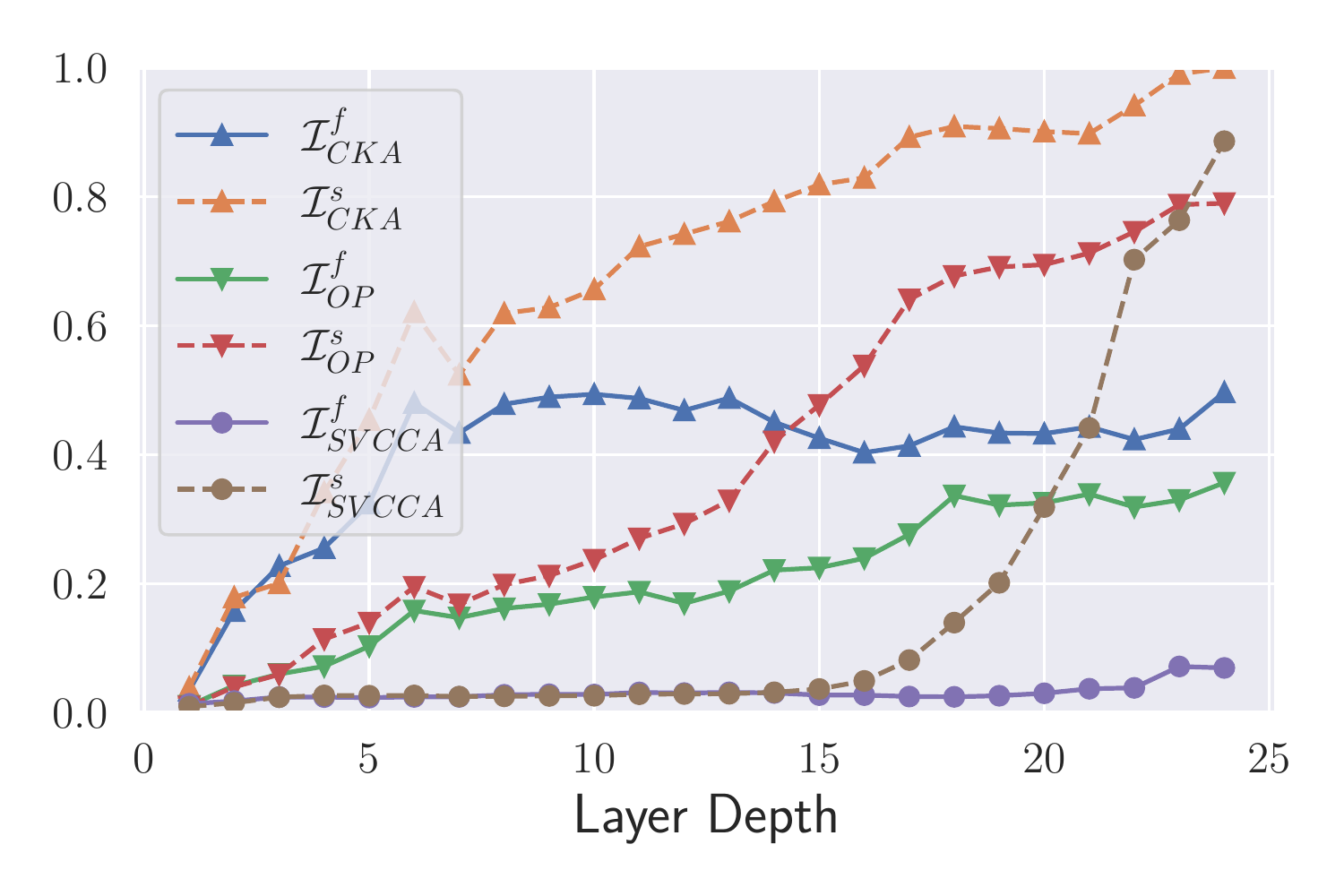}
        \caption{MRPC}
        \label{subfig:app_sf_bert_mrpc}
    \end{subfigure}
    \hfill
    \begin{subfigure}[b]{0.45\textwidth}
        \centering
        \includegraphics[width=\textwidth]{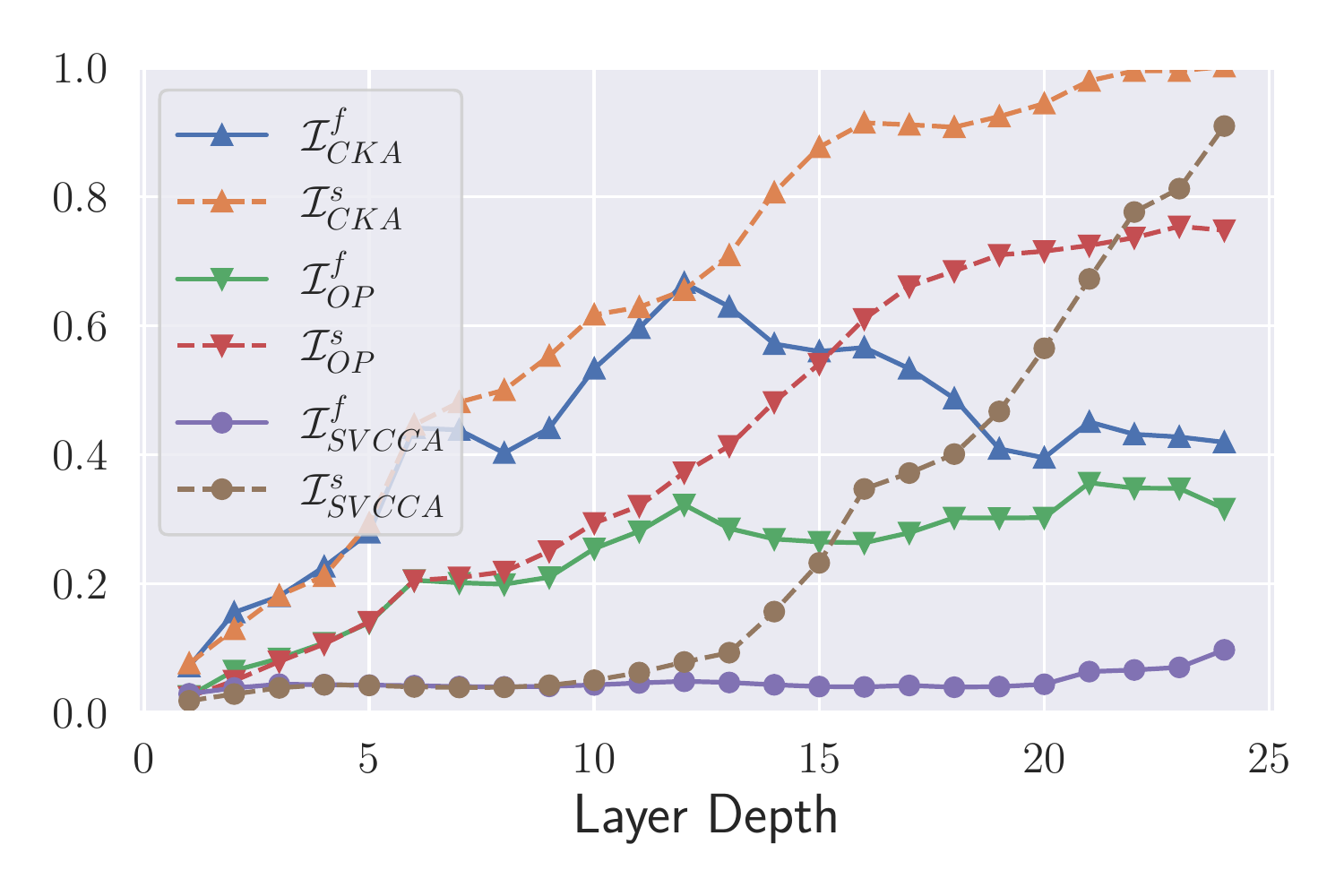}
        \caption{CoLA}
        \label{subfig:app_sf_bert_cola}
    \end{subfigure}
    \caption{Results of successful vs. failed runs, 
        BERT on MRPC and CoLA.}
    \label{fig:app_sf_bert}
\end{figure*}

\begin{figure*}
    \centering
    \begin{subfigure}[b]{0.32\textwidth}
        \centering
        \includegraphics[width=\textwidth]{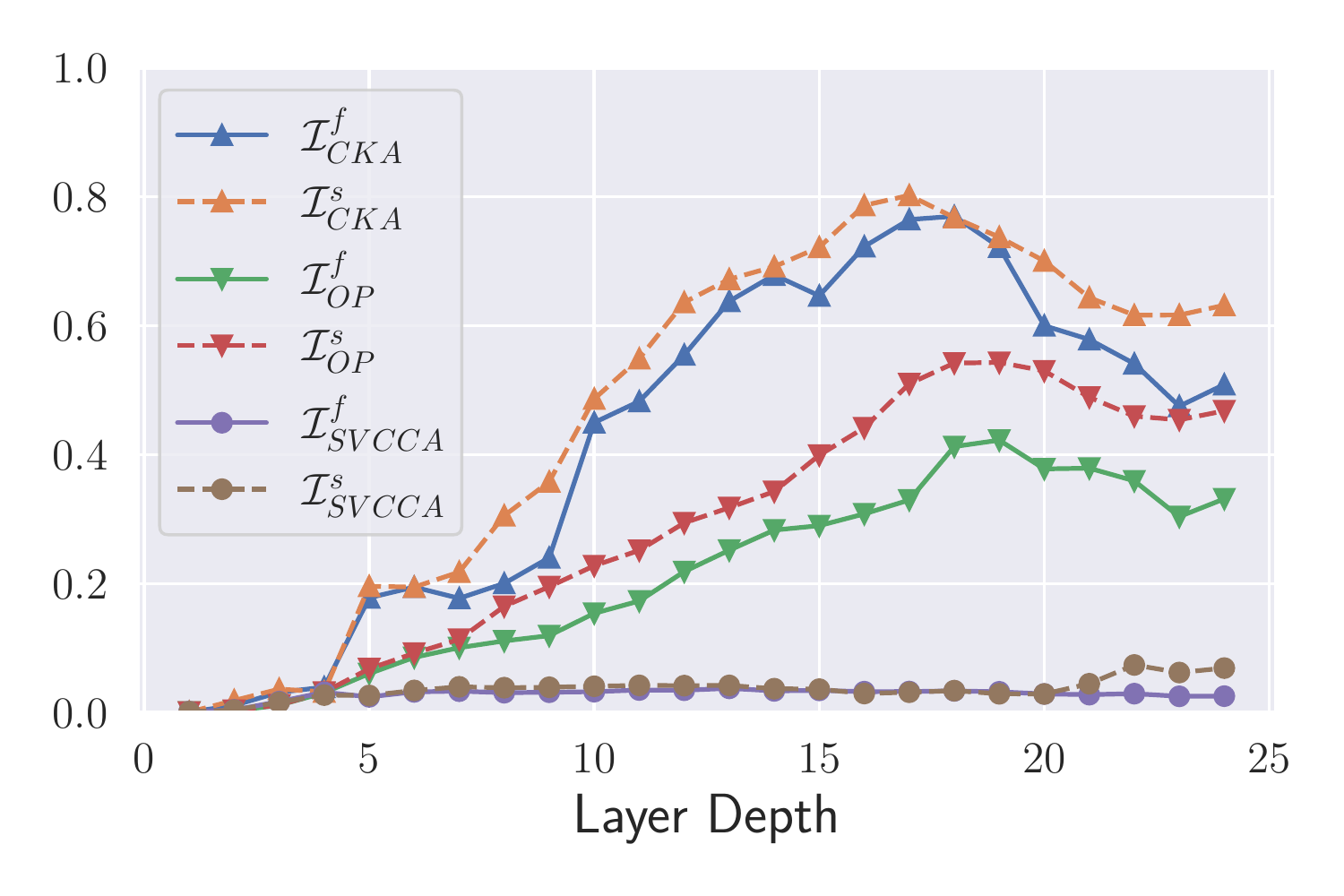}
        \caption{RTE}
        \label{subfig:app_sf_roberta_rte}
    \end{subfigure}
    \hfill
    \begin{subfigure}[b]{0.32\textwidth}
        \centering
        \includegraphics[width=\textwidth]{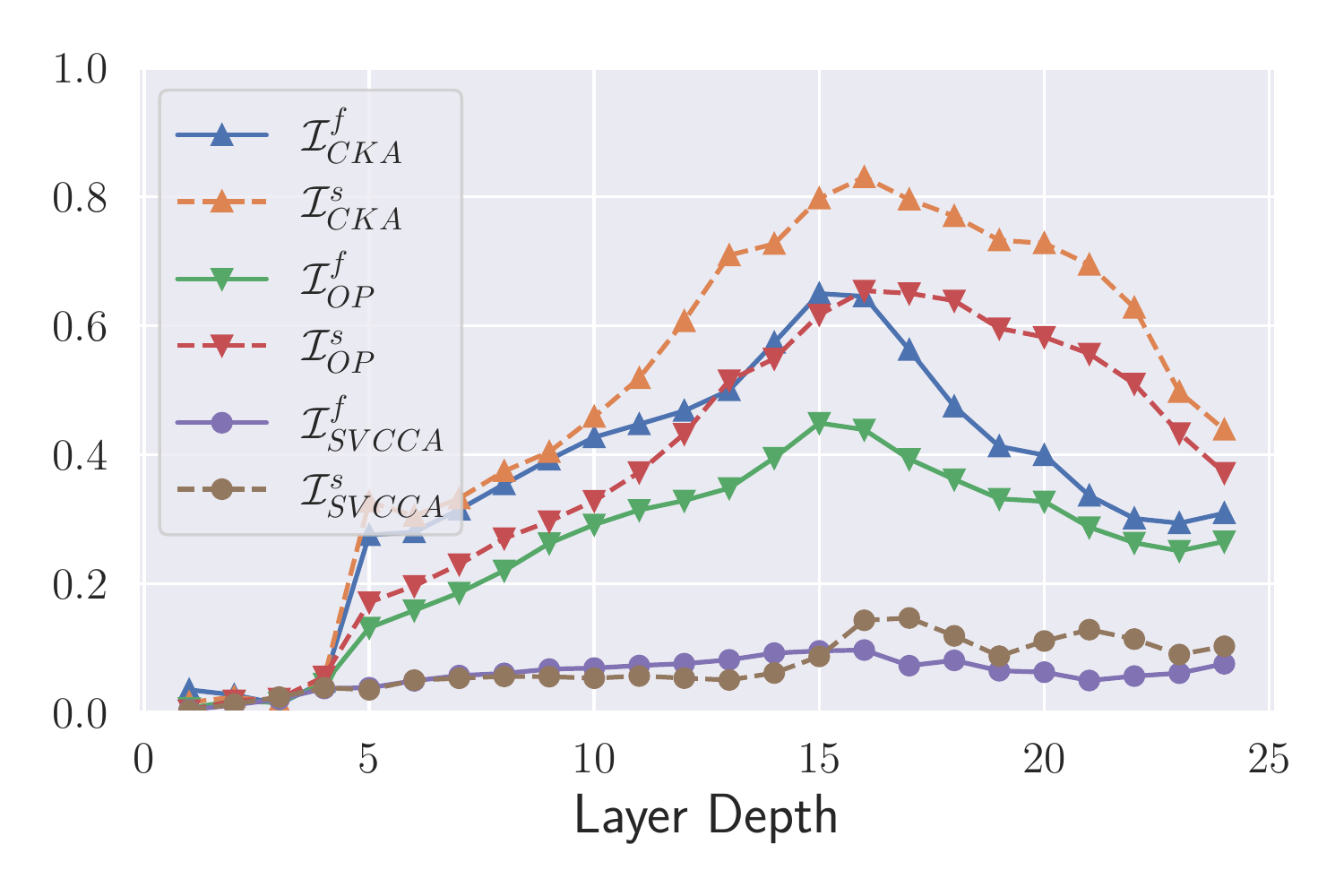}
        \caption{MRPC}
        \label{subfig:app_sf_roberta_mrpc}
    \end{subfigure}
    \hfill
    \begin{subfigure}[b]{0.32\textwidth}
        \centering
        \includegraphics[width=\textwidth]{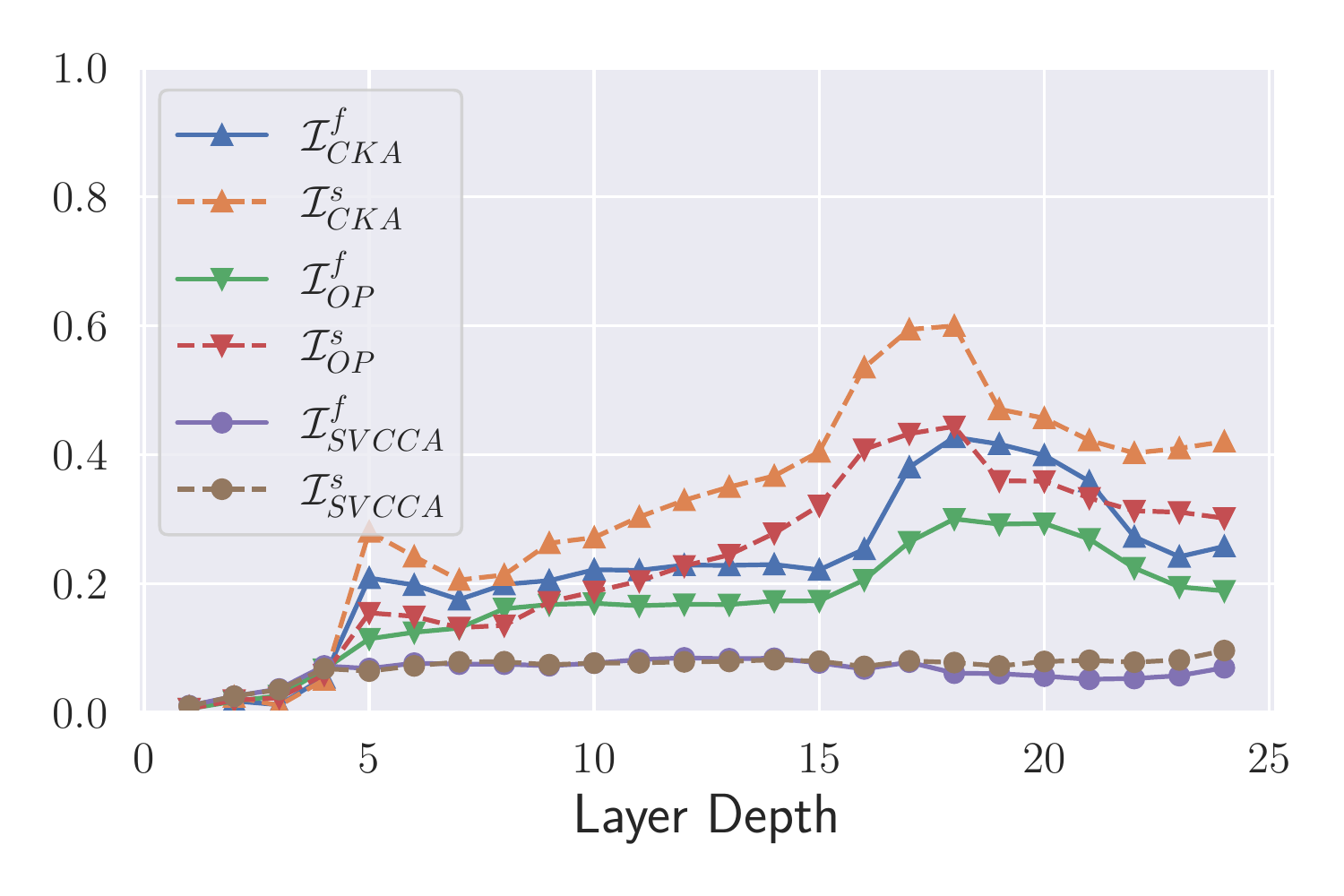}
        \caption{CoLA}
        \label{subfig:app_sf_roberta_cola}
    \end{subfigure}
    \caption{Results of successful vs. failed runs, 
        RoBERTa on MRPC and CoLA.}
    \label{fig:app_sf_roberta}
\end{figure*}

\begin{figure*}
    \centering
    \begin{subfigure}[b]{0.45\textwidth}
        \centering
        \includegraphics[width=\textwidth]{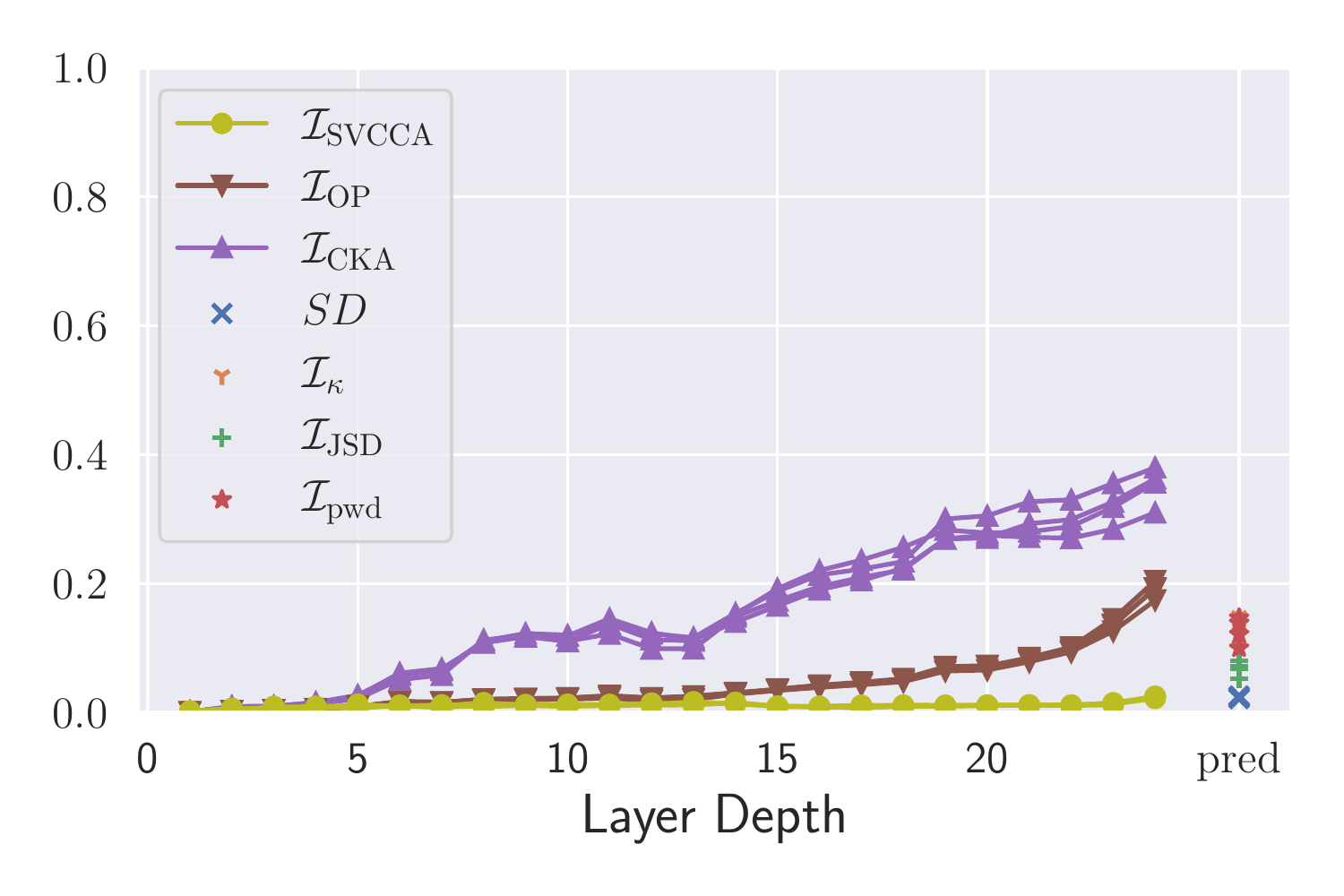}
        \caption{RTE}
        \label{subfig:app_subsample_bert_rte}
    \end{subfigure}
    \hfill
    \begin{subfigure}[b]{0.45\textwidth}
        \centering
        \includegraphics[width=\textwidth]{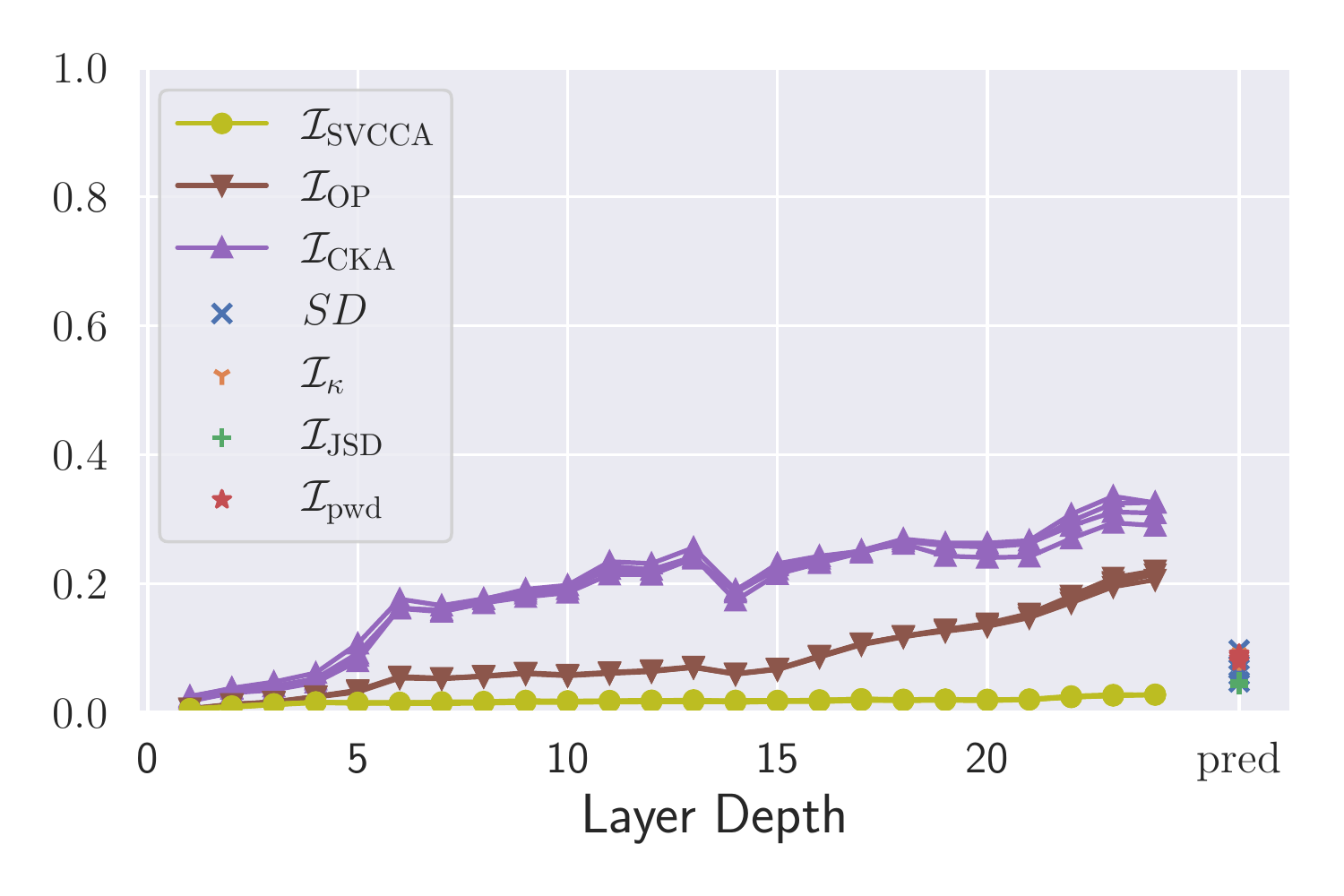}
        \caption{MRPC}
        \label{subfig:app_subsample_bert_cola}
    \end{subfigure}
    \caption{
        Consistency among sub-samples on BERT (on RTE and CoLA), sample rate 0.5.
    }
    \label{fig:app_subsample_bert}
\end{figure*}

\begin{figure*}
    \centering
    \begin{subfigure}[b]{0.32\textwidth}
        \centering
        \includegraphics[width=\textwidth]{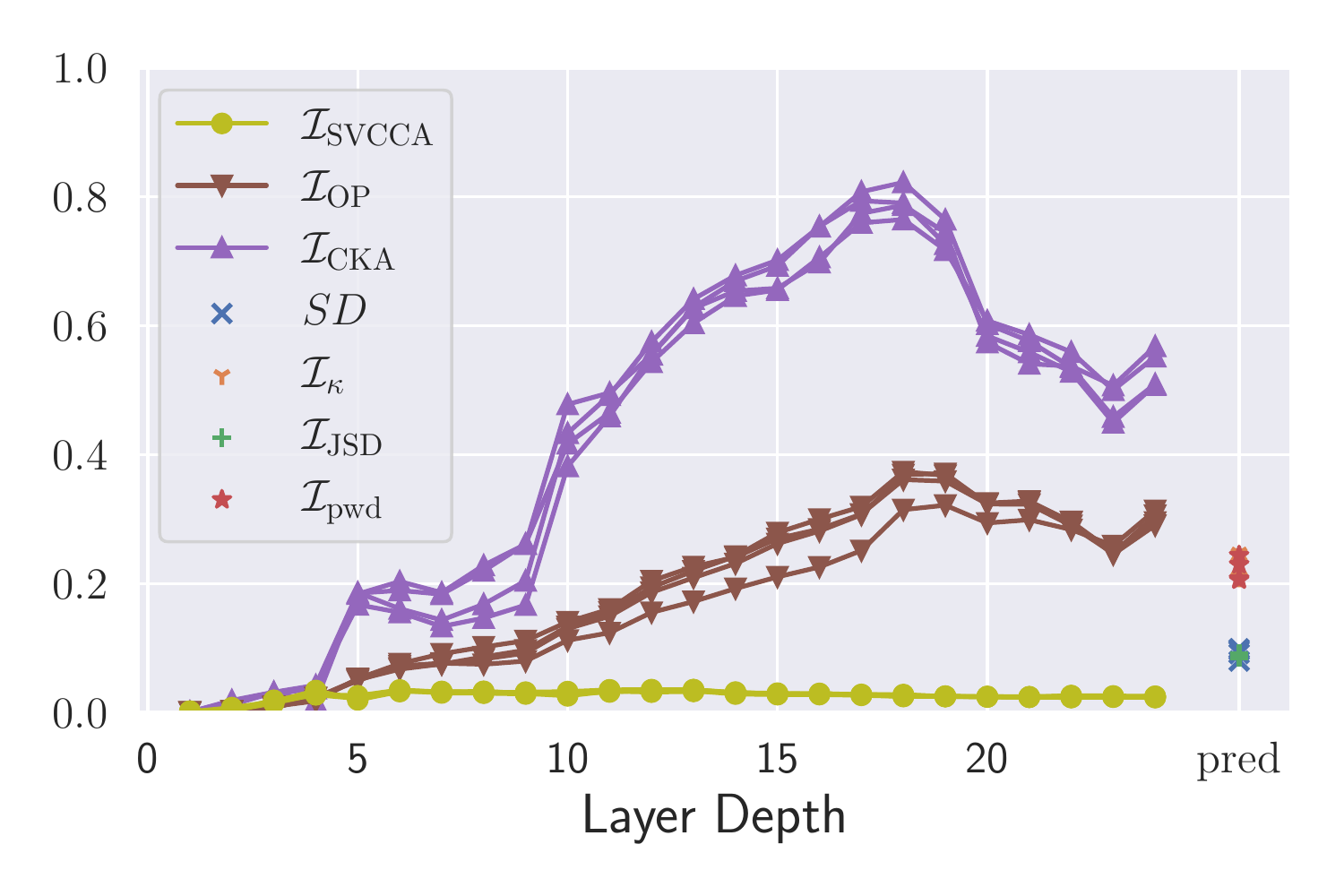}
        \caption{RTE}
        \label{subfig:app_subsample_roberta_rte}
    \end{subfigure}
    \hfill
    \begin{subfigure}[b]{0.32\textwidth}
        \centering
        \includegraphics[width=\textwidth]{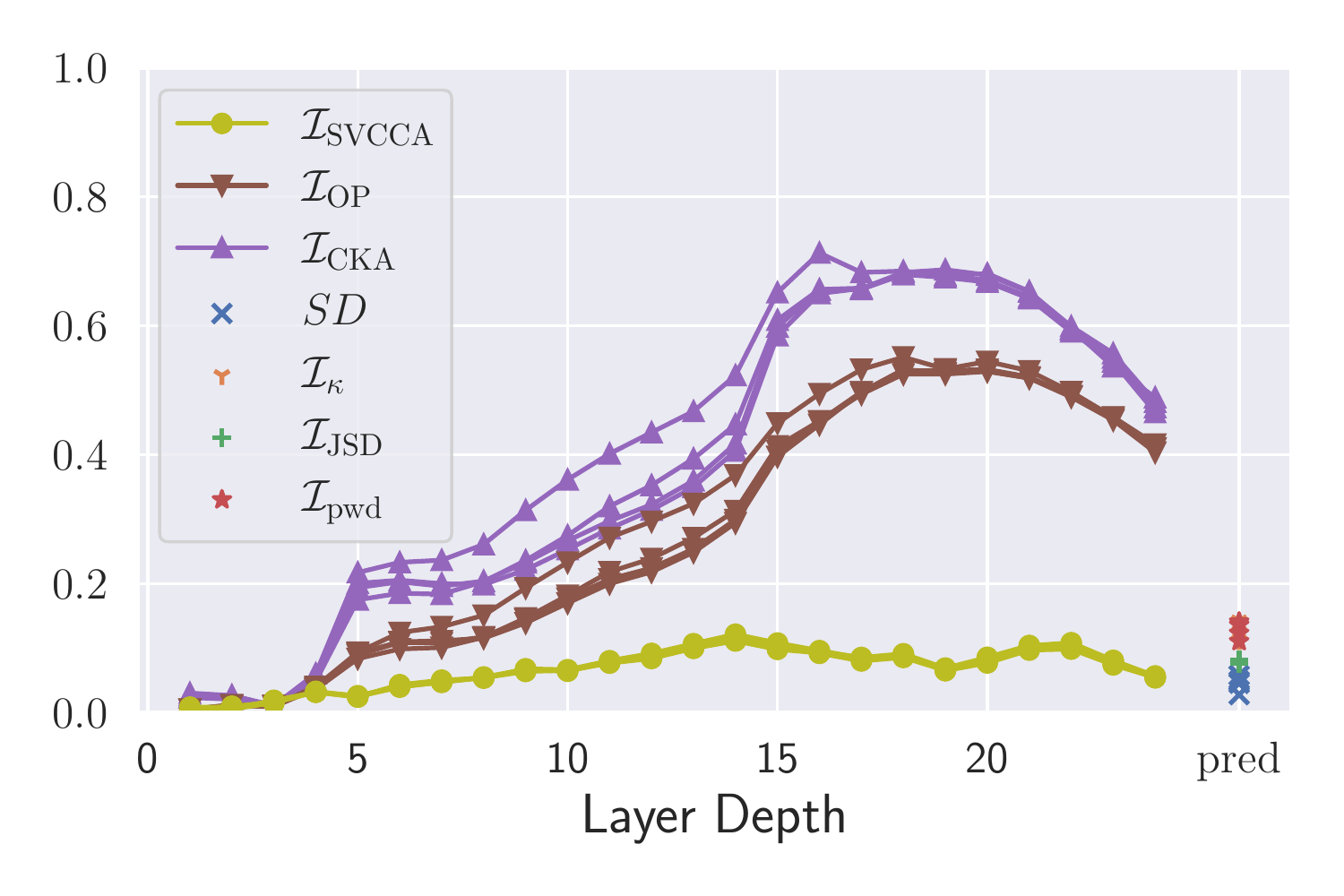}
        \caption{MRPC}
        \label{subfig:app_subsample_roberta_mrpc}
    \end{subfigure}
    \hfill
    \begin{subfigure}[b]{0.32\textwidth}
        \centering
        \includegraphics[width=\textwidth]{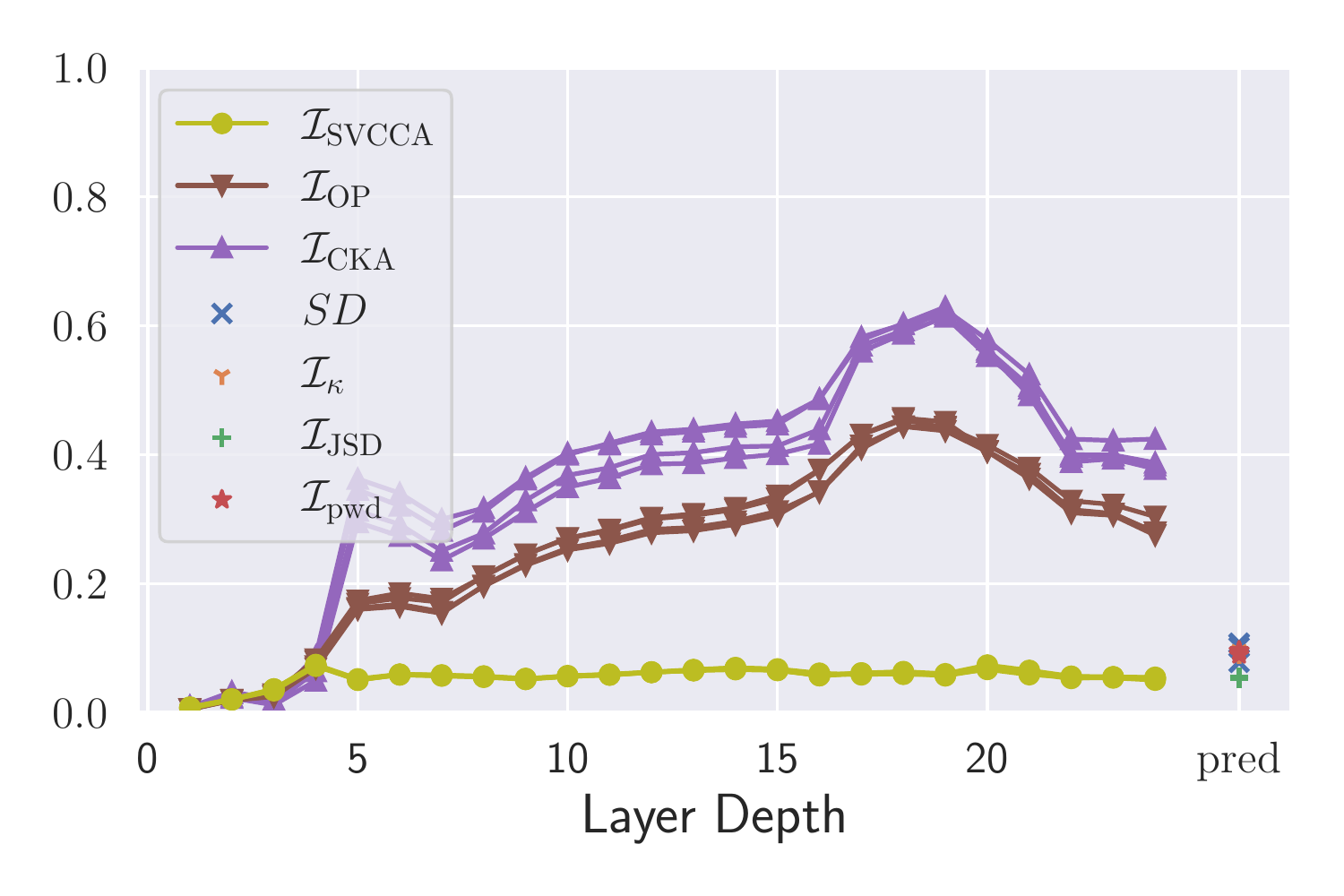}
        \caption{CoLA}
        \label{subfig:app_subsample_roberta_cola}
    \end{subfigure}
    \caption{
        Consistency among sub-samples on RoBERTa, sample rate 0.1.}
    \label{fig:app_subsample_roberta}
\end{figure*}

\begin{figure*}
    \centering
    \begin{subfigure}[b]{0.32\textwidth}
        \centering
        \includegraphics[width=\textwidth]{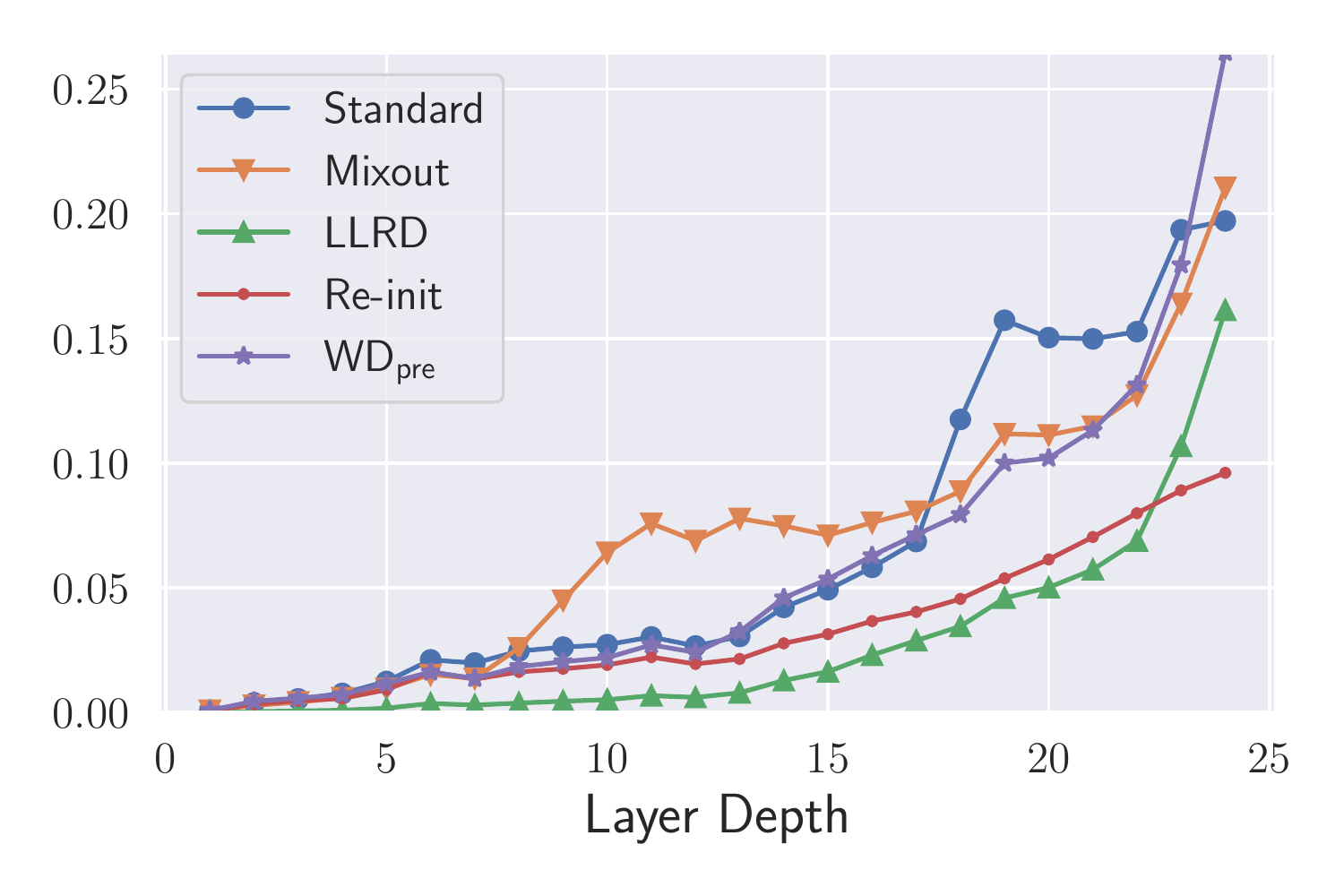}
        \caption{RTE}
        \label{subfig:reassess_rte_bert_op}
    \end{subfigure}
    \hfill
    \begin{subfigure}[b]{0.32\textwidth}
        \centering
        \includegraphics[width=\textwidth]{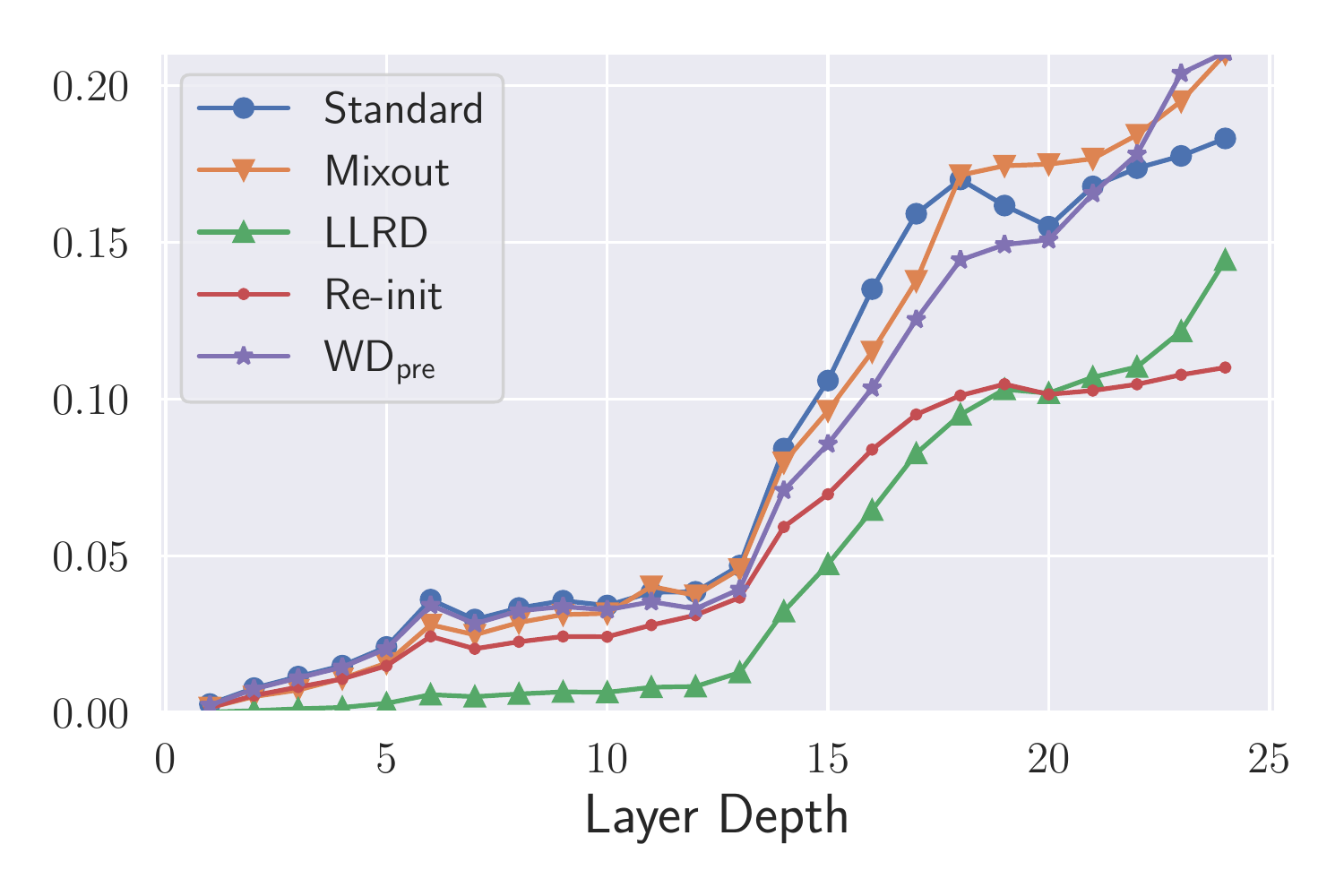}
        \caption{MRPC}
        \label{subfig:reassess_mrpc_bert_op}
    \end{subfigure}
    \hfill
    \begin{subfigure}[b]{0.32\textwidth}
        \centering
        \includegraphics[width=\textwidth]{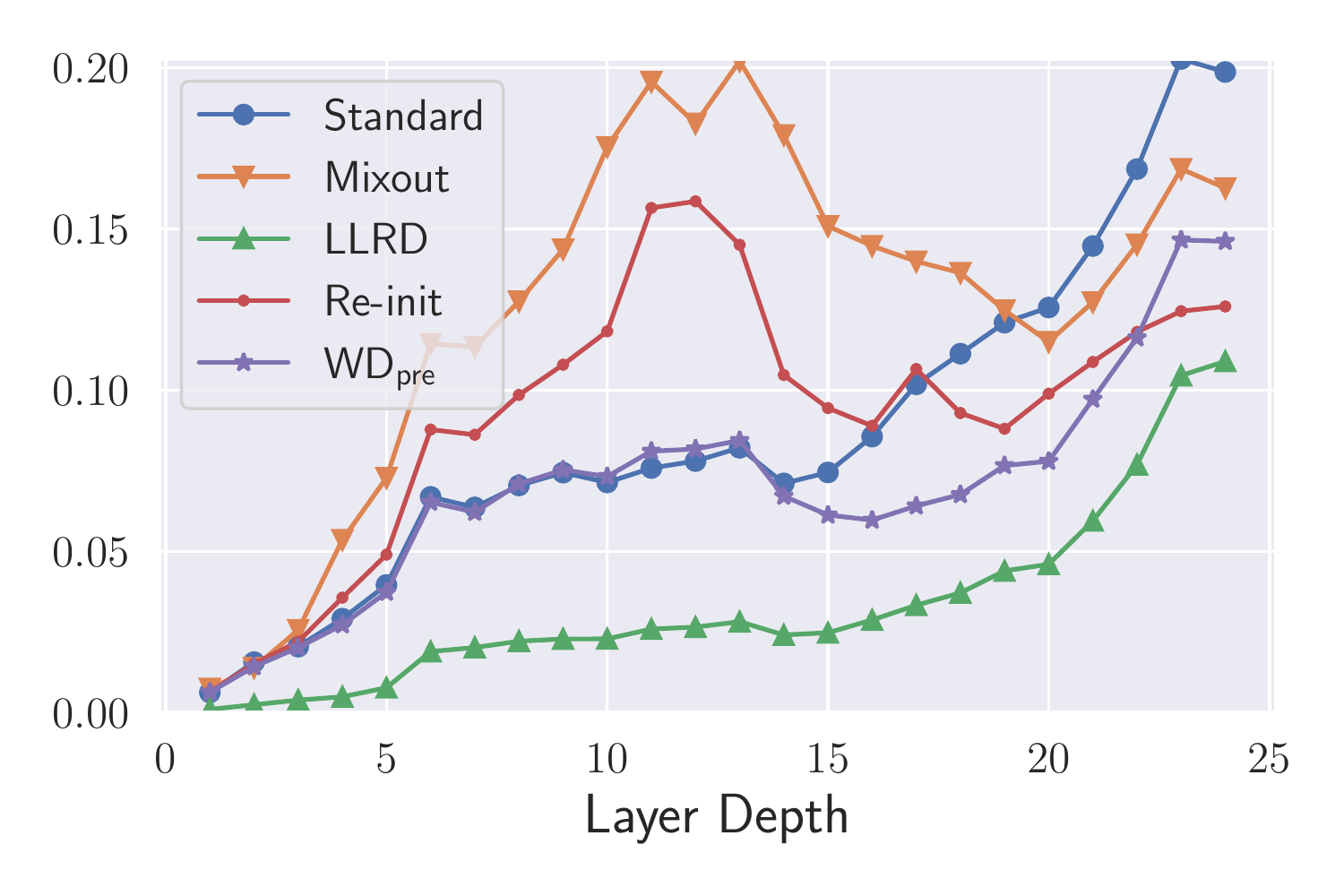}
        \caption{CoLA}
        \label{subfig:reassess_cola_bert_op}
    \end{subfigure}
    \caption{Representation instability for BERT after applying
        different instability mitigation methods on all three datasets, 
        measured by OP.}
    \label{fig:app_reassess_bert_op}
\end{figure*}

\begin{figure*}
    \centering
    \begin{subfigure}[b]{0.32\textwidth}
        \centering
        \includegraphics[width=\textwidth]{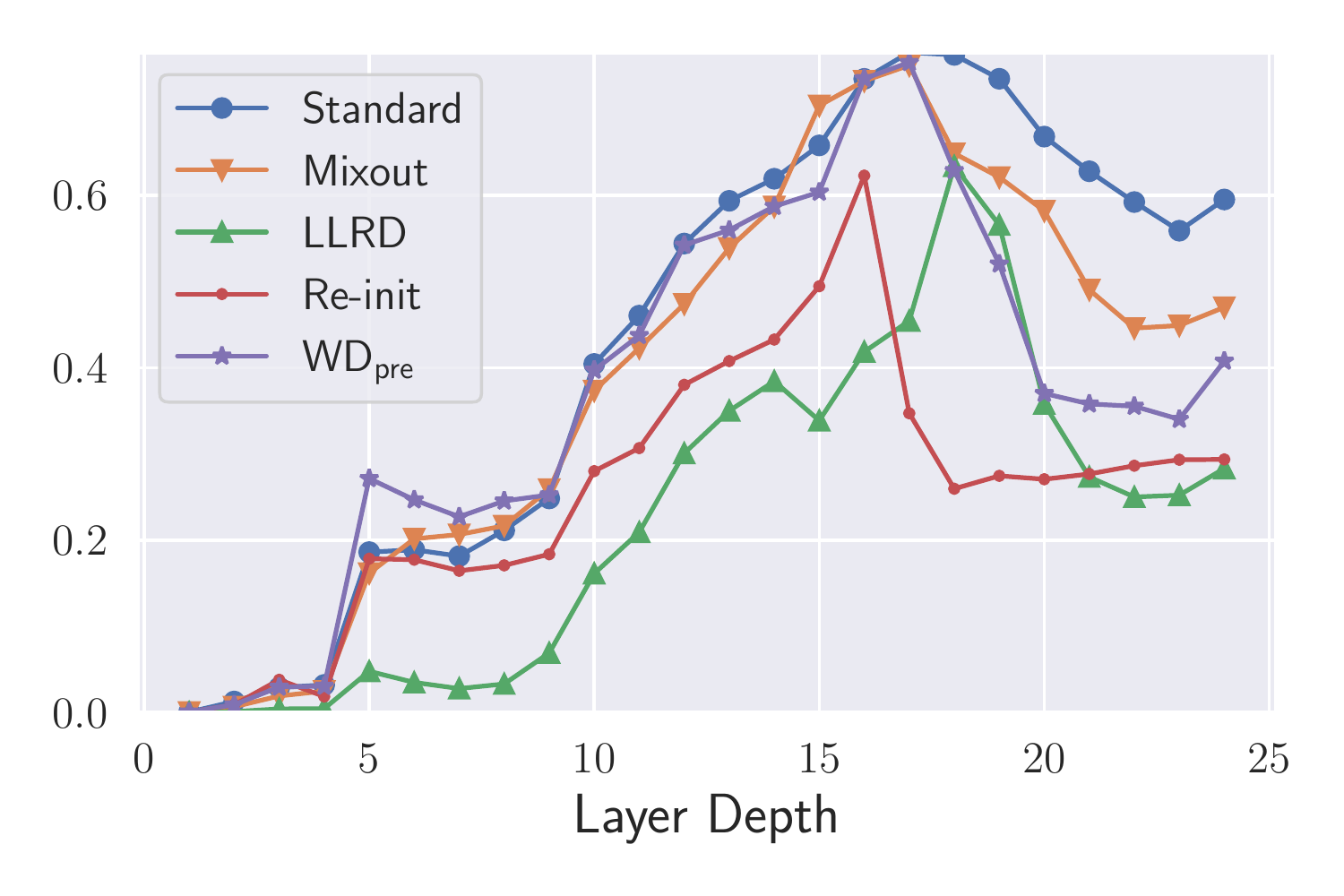}
        \caption{RTE}
        \label{subfig:reassess_rte_roberta_cka}
    \end{subfigure}
    \hfill
    \begin{subfigure}[b]{0.32\textwidth}
        \centering
        \includegraphics[width=\textwidth]{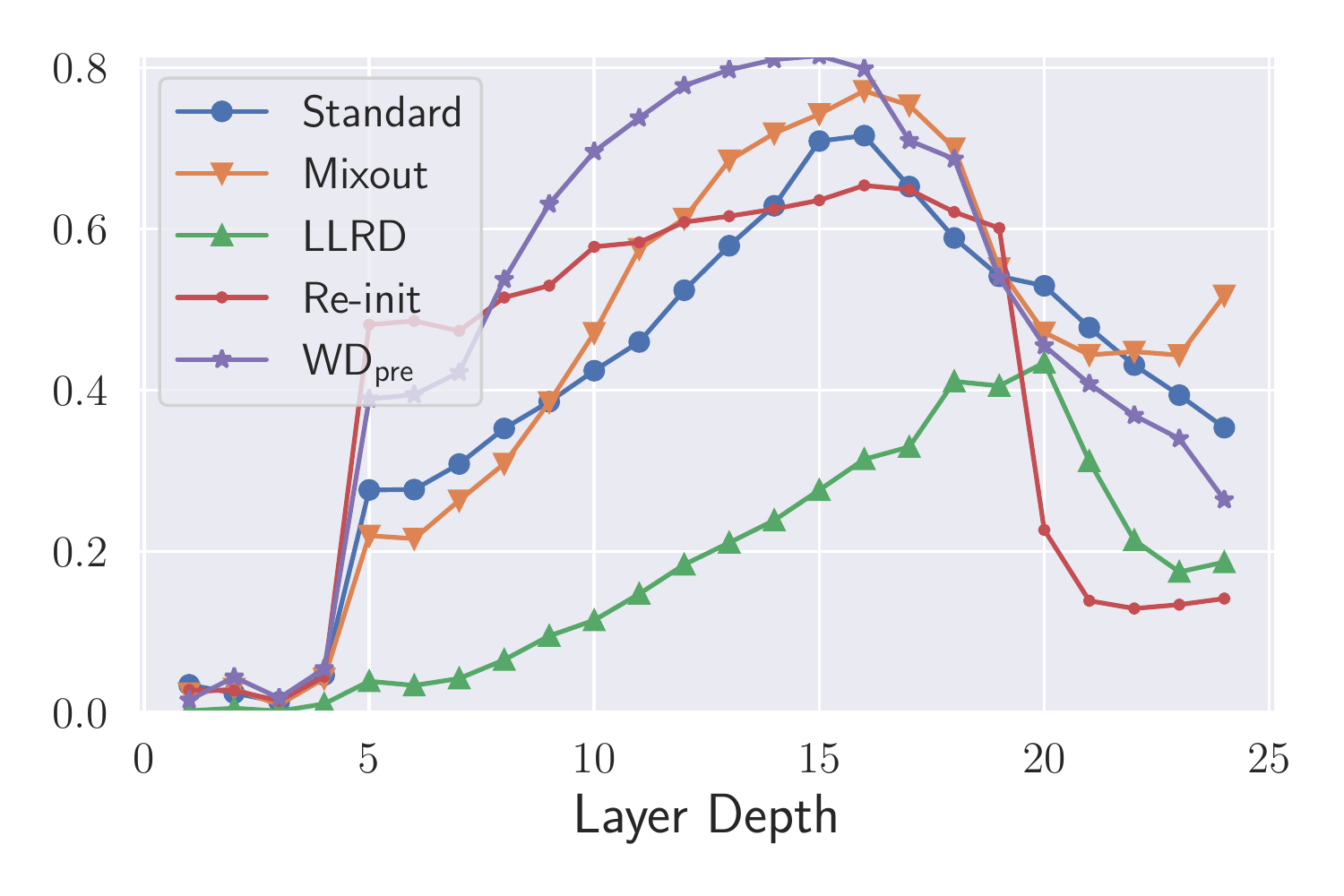}
        \caption{MRPC}
        \label{subfig:reassess_mrpc_roberta_cka}
    \end{subfigure}
    \hfill
    \begin{subfigure}[b]{0.32\textwidth}
        \centering
        \includegraphics[width=\textwidth]{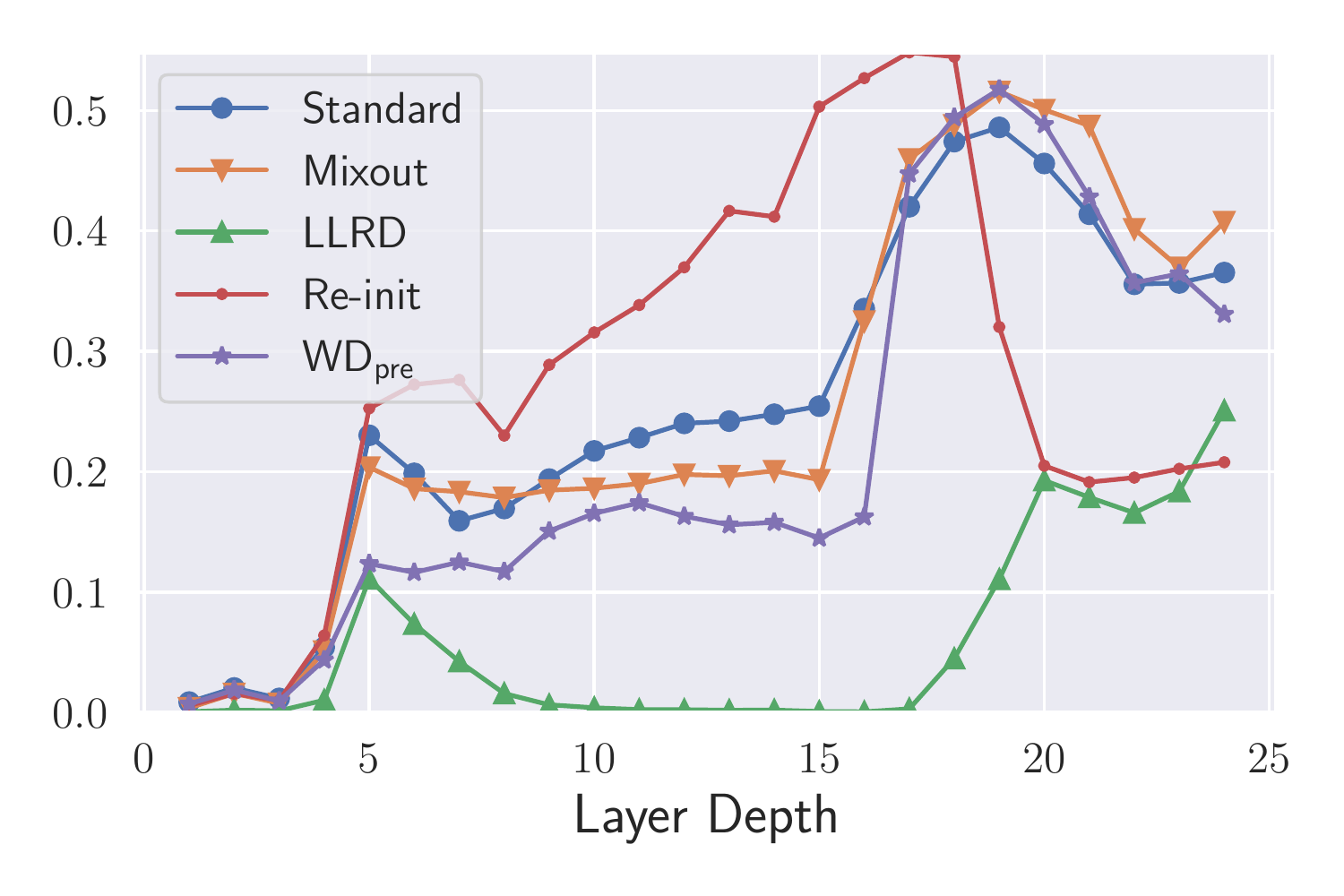}
        \caption{CoLA}
        \label{subfig:reassess_cola_roberta_cka}
    \end{subfigure}
    \caption{Representation instability for RoBERTa after applying
        different instability mitigation methods on all three datasets, 
        measured by Linear-CKA.}
    \label{fig:app_reassess_roberta_cka}
\end{figure*}

\begin{figure*}
    \centering
    \begin{subfigure}[b]{0.32\textwidth}
        \centering
        \includegraphics[width=\textwidth]{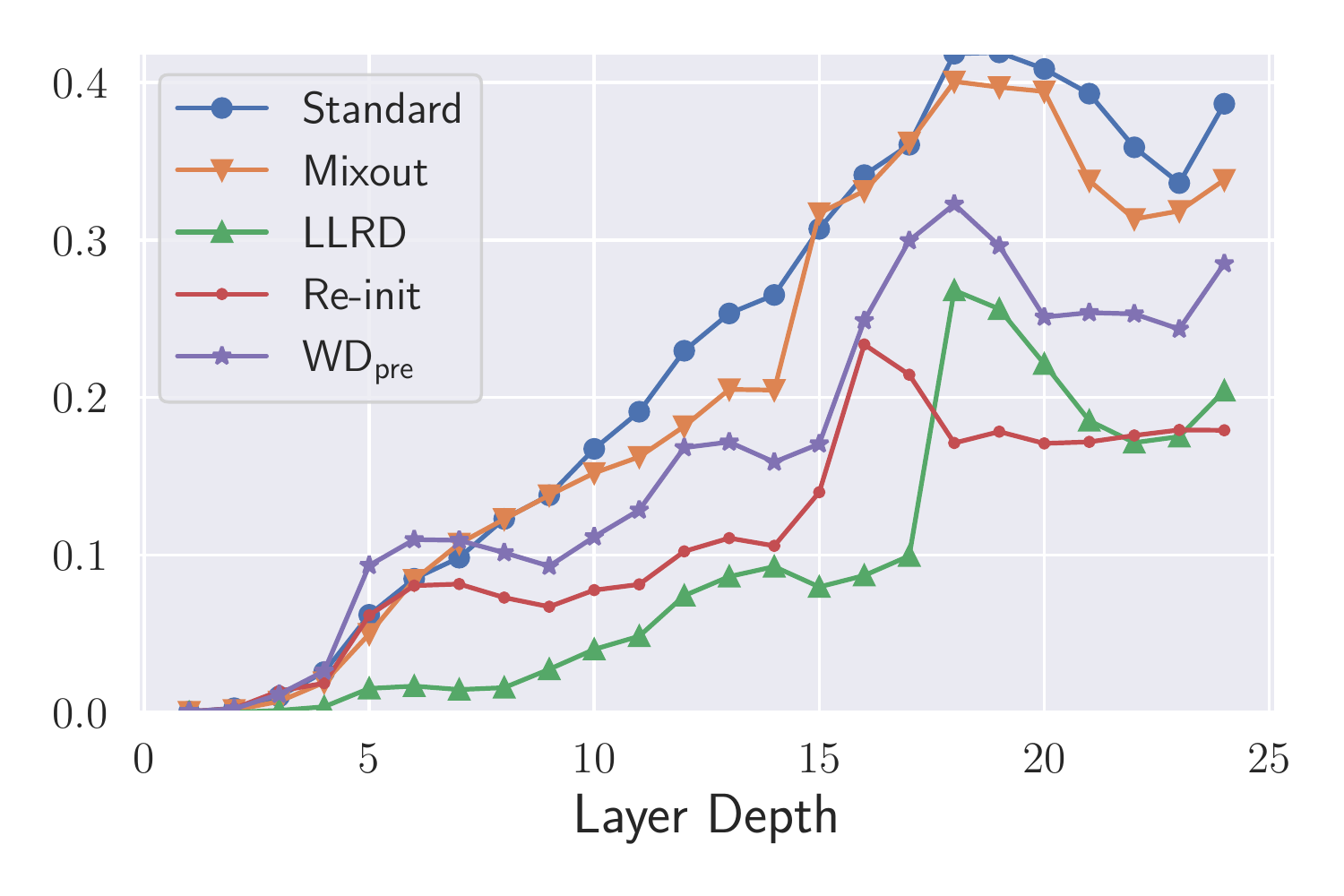}
        \caption{RTE}
        \label{subfig:reassess_rte_roberta_op}
    \end{subfigure}
    \hfill
    \begin{subfigure}[b]{0.32\textwidth}
        \centering
        \includegraphics[width=\textwidth]{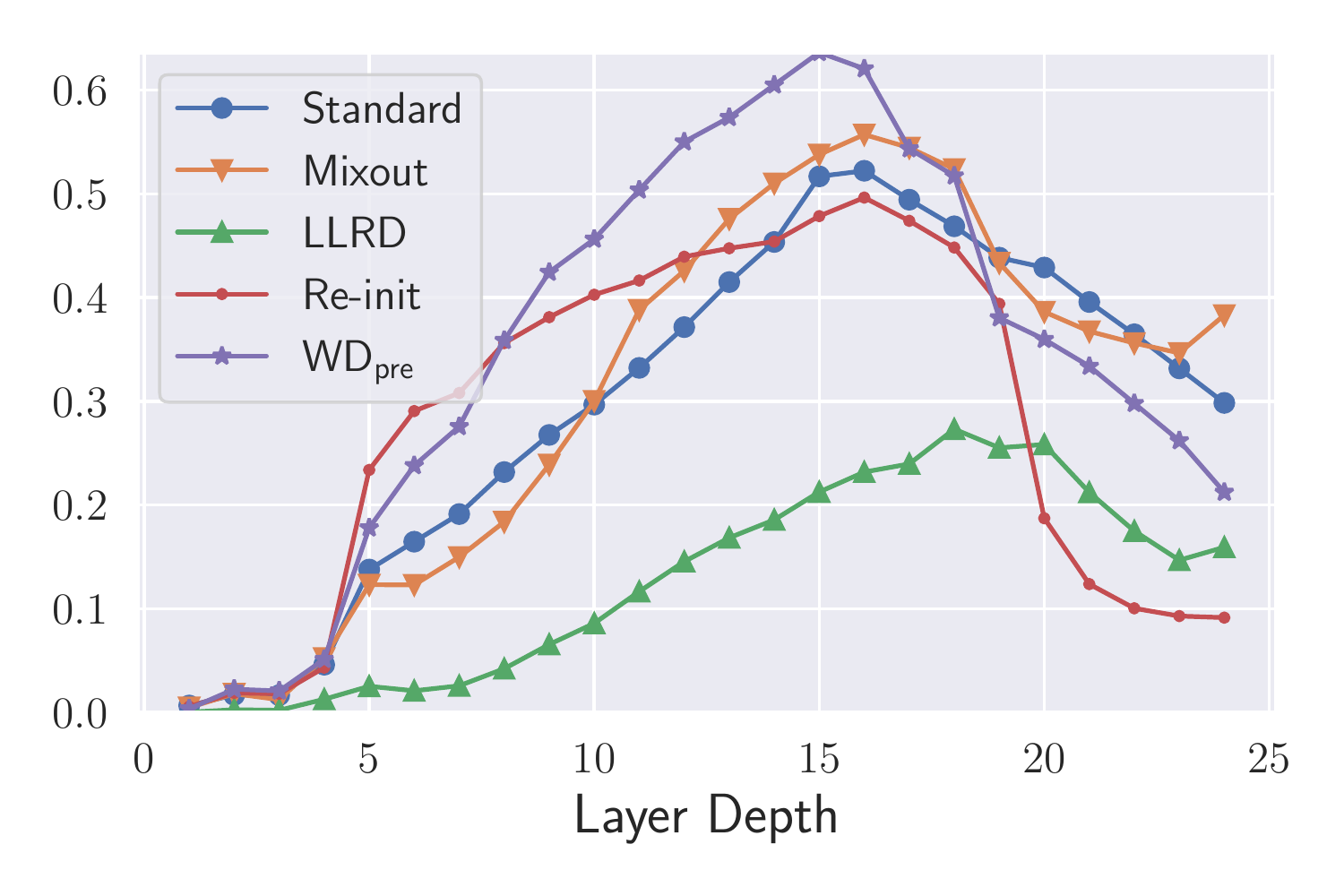}
        \caption{MRPC}
        \label{subfig:reassess_mrpc_roberta_op}
    \end{subfigure}
    \hfill
    \begin{subfigure}[b]{0.32\textwidth}
        \centering
        \includegraphics[width=\textwidth]{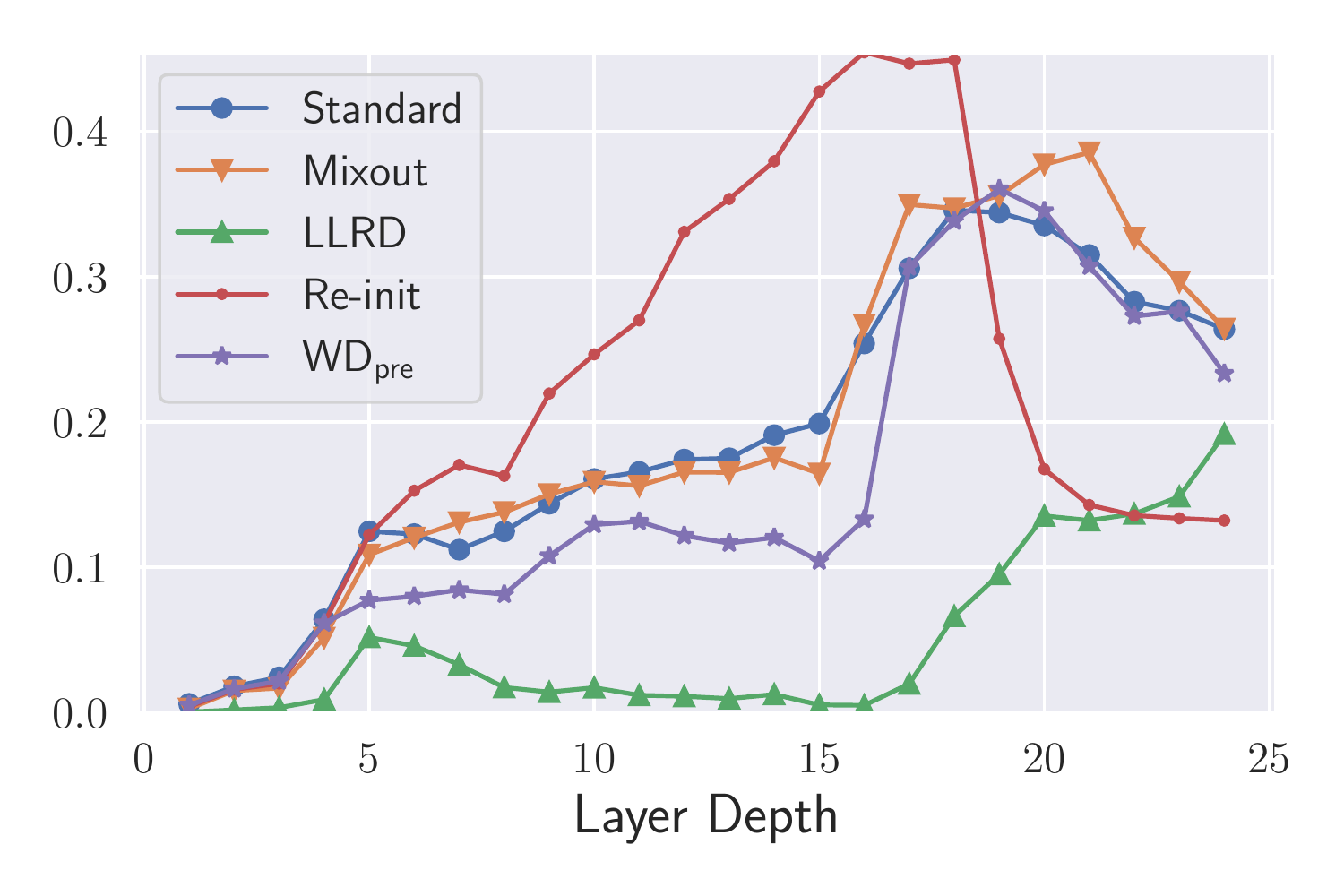}
        \caption{CoLA}
        \label{subfig:reassess_cola_roberta_op}
    \end{subfigure}
    \caption{Representation instability for RoBERTa after applying
        different instability mitigation methods on all three datasets, 
        measured by OP.}
    \label{fig:app_reassess_roberta_op}
\end{figure*}

Figures are on the next page. 
In Figures~\ref{fig:app_sf_bert} -- \ref{fig:app_reassess_roberta_op}, 
the Y-axis refers to the instability scores computed by different measures. 

\begin{table*}[ht]
    \centering
    {\fontsize{8}{11}\selectfont
    \begin{tabular}{@{}ccccccccccccc@{}}
    \toprule
                                             & \multicolumn{4}{c}{RTE}                                                                                                                      & \multicolumn{4}{c}{MRPC}                                                                                                                     & \multicolumn{4}{c}{CoLA}                                                                                                \\
    \multicolumn{1}{c|}{}                    & Acc $\pm$ SD                     & $\mathcal{I}_{\mathrm{JSD}}$ & $\mathcal{I}_{\mathrm{\kappa}}$ & \multicolumn{1}{c|}{$\mathcal{I}_{\mathrm{pwd}}$} & F1 $\pm$ SD             & $\mathcal{I}_{\mathrm{JSD}}$ & $\mathcal{I}_{\mathrm{\kappa}}$ & \multicolumn{1}{c|}{$\mathcal{I}_{\mathrm{pwd}}$} & MCC $\pm$ SD            & $\mathcal{I}_{\mathrm{JSD}}$ & $\mathcal{I}_{\mathrm{\kappa}}$ & $\mathcal{I}_{\mathrm{pwd}}$ \\ \midrule
    \multicolumn{1}{c|}{Standard}            & 74.4 $\pm$ 12.2                  & 9.8                          & 25.6                            & \multicolumn{1}{c|}{25.4}                         & 90.8 $\pm$ 3.6          & 5.9                          & 10.2                            & \multicolumn{1}{c|}{10.1}                         & 65.6 $\pm$ 7.8          & 4.9                          & 8.9                             & 8.9                          \\
    \multicolumn{1}{c|}{Mixout}              & 79.3 $\pm$ 4.4                   & 9.3                          & 16.7                            & \multicolumn{1}{c|}{16.6}                         & 89.4 $\pm$ 3.2          & 6.2                          & 13.2                            & \multicolumn{1}{c|}{13.1}                         & 68.1 $\pm$ \textbf{2.2} & 4.6                          & 8.7                             & 8.7                          \\
    \multicolumn{1}{c|}{LLRD}                & \textbf{81.3} $\pm$ \textbf{1.8} & \textbf{5.7}                 & \textbf{11.2}                   & \multicolumn{1}{c|}{\textbf{11.2}}                & 91.3 $\pm$ \textbf{0.6} & 3.3                          & 6.2                             & \multicolumn{1}{c|}{6.2}                          & \textbf{69.7} $\pm$ 4.1 & \textbf{3.0}                 & \textbf{6.2}                    & \textbf{6.2}                 \\
    \multicolumn{1}{c|}{Re-init}             & 79.6 $\pm$ 2.0                   & 7.2                          & 12.7                            & \multicolumn{1}{c|}{12.6}                         & \textbf{92.5} $\pm$ 0.8 & \textbf{3.0}                 & \textbf{5.3}                    & \multicolumn{1}{c|}{\textbf{5.3}}                 & 69.2 $\pm$ 2.7          & 3.8                          & 7.1                             & 7.1                          \\
    \multicolumn{1}{c|}{$\mathsf{WD_{pre}}$} & \textbf{81.3} $\pm$ 2.8          & 6.6                          & 13.0                            & \multicolumn{1}{c|}{12.9}                         & 92.0 $\pm$ 1.0          & 3.6                          & 6.7                             & \multicolumn{1}{c|}{6.6}                          & 66.6 $\pm$ 2.5          & 4.4                          & 8.4                             & 8.4                          \\ \bottomrule
    \end{tabular}
    }
    \caption{Prediction instability scores of RoBERTa 
    after applying different IMMs.
    To obtain better readability, all values shown here are multiplied by 100. 
    Higher values indicate higher instability for instability measures 
    (i.e. except for performance metrics Acc, F1, and MCC).}
    \label{tab:reassess_pred_roberta}
\end{table*}

\begin{figure*}
    \centering
    \begin{subfigure}[b]{0.32\textwidth}
        \centering
        \includegraphics[width=\textwidth]{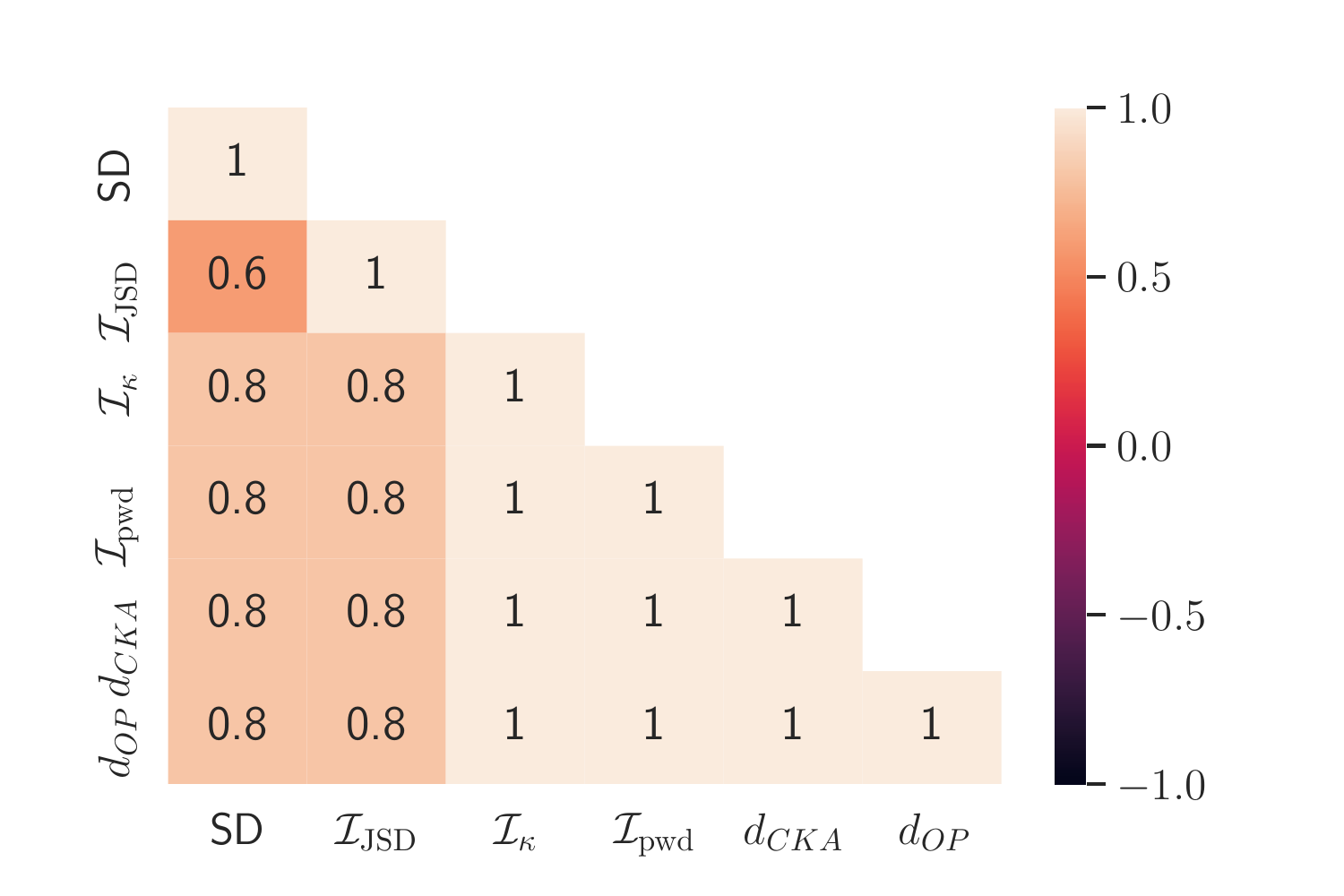}
        \caption{RTE}
        \label{subfig:bert_rte_tau}
    \end{subfigure}
    \hfill
    \begin{subfigure}[b]{0.32\textwidth}
        \centering
        \includegraphics[width=\textwidth]{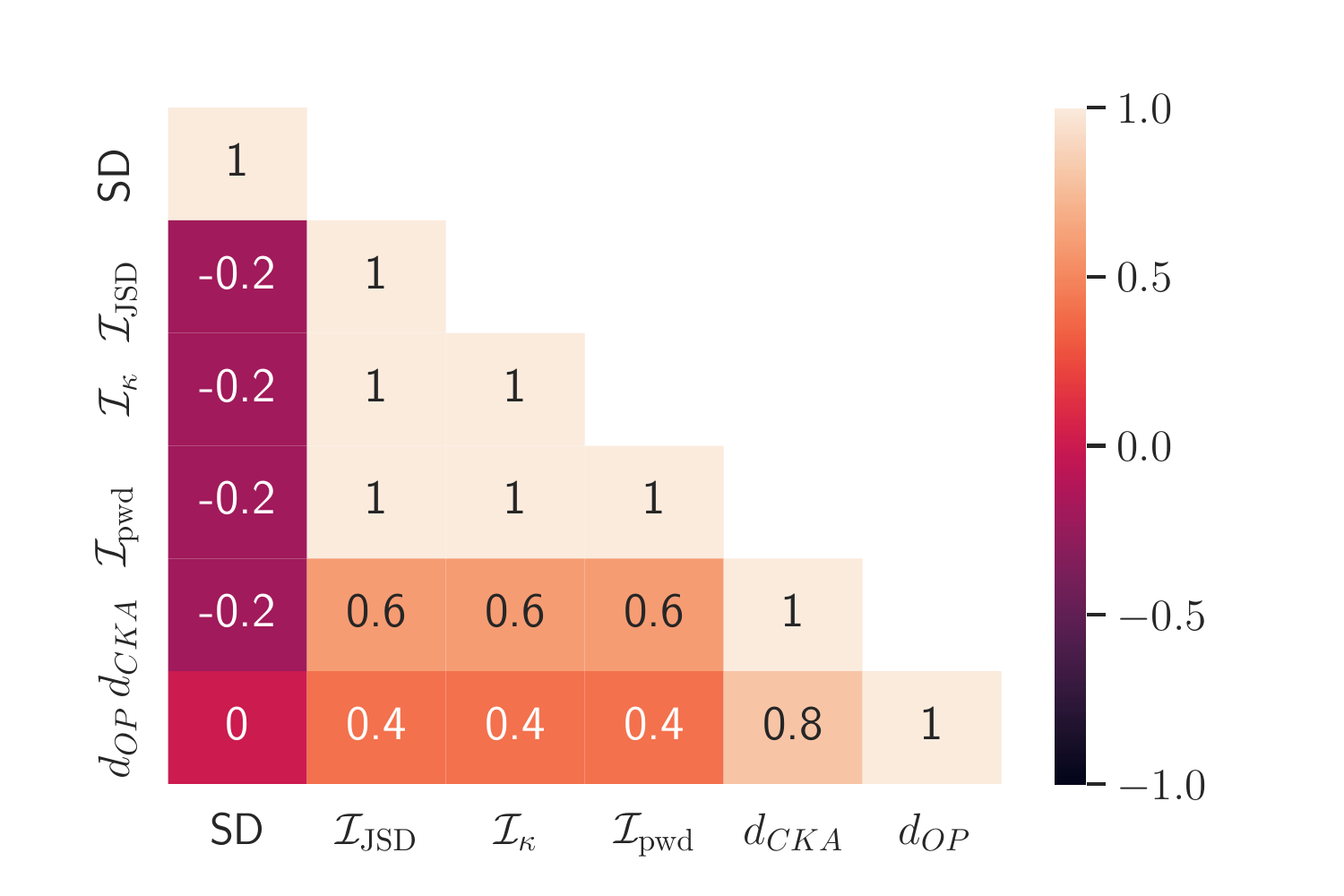}
        \caption{MRPC}
        \label{subfig:bert_mrpc_tau}
    \end{subfigure}
    \hfill
    \begin{subfigure}[b]{0.32\textwidth}
        \centering
        \includegraphics[width=\textwidth]{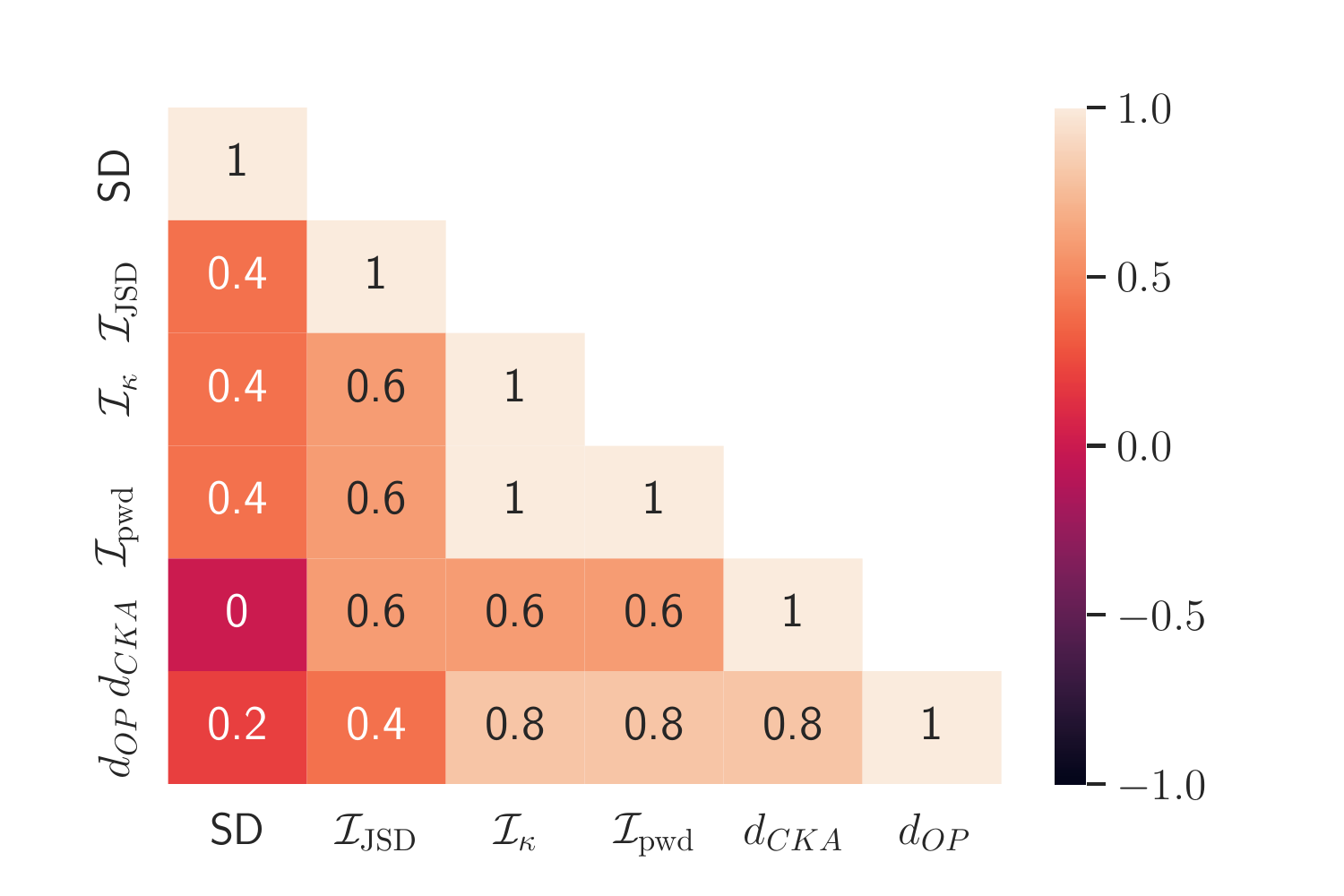}
        \caption{CoLA}
        \label{subfig:bert_cola_tau}
    \end{subfigure}
    \caption{
        Kendall's $\tau$ values after applying different instability mitigation methods on BERT,
        between each pair of measures.
        For representation measures, we take the value of the topmost layer.
    }
    \label{fig:app_bert_tau}
\end{figure*}

\begin{figure*}
    \centering
    \begin{subfigure}[b]{0.32\textwidth}
        \centering
        \includegraphics[width=\textwidth]{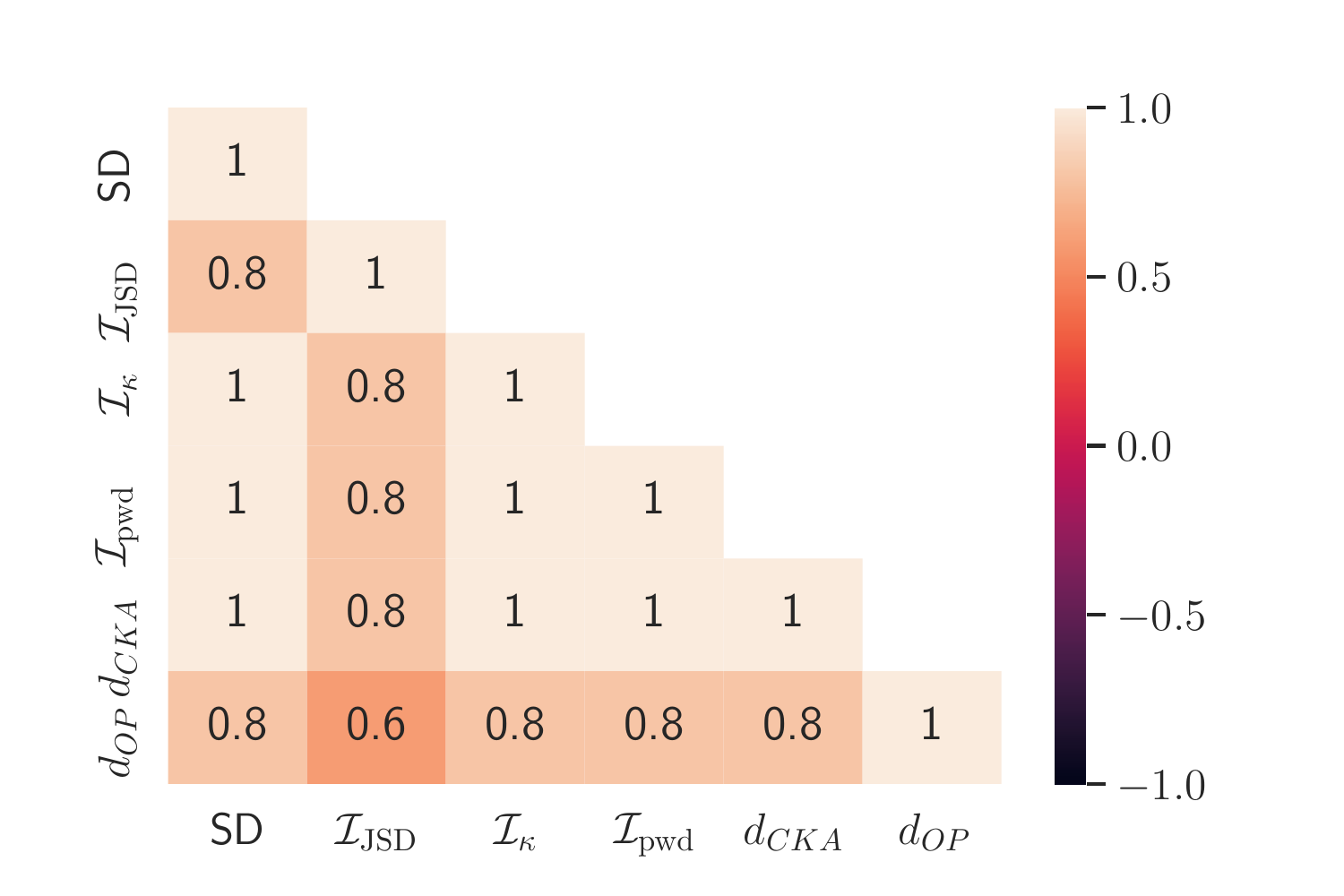}
        \caption{RTE}
        \label{subfig:roberta_rte_tau}
    \end{subfigure}
    \hfill
    \begin{subfigure}[b]{0.32\textwidth}
        \centering
        \includegraphics[width=\textwidth]{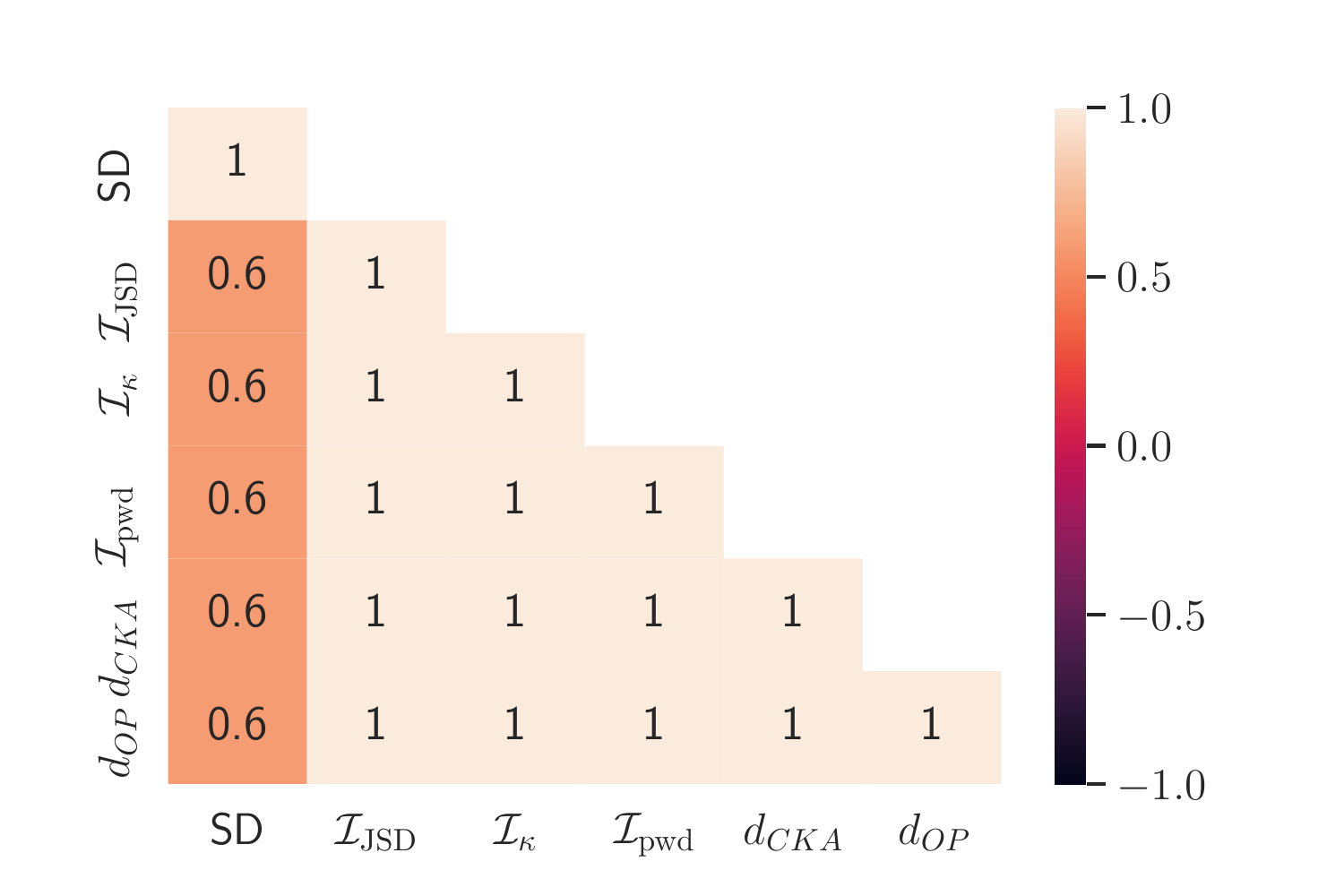}
        \caption{MRPC}
        \label{subfig:roberta_mrpc_tau}
    \end{subfigure}
    \hfill
    \begin{subfigure}[b]{0.32\textwidth}
        \centering
        \includegraphics[width=\textwidth]{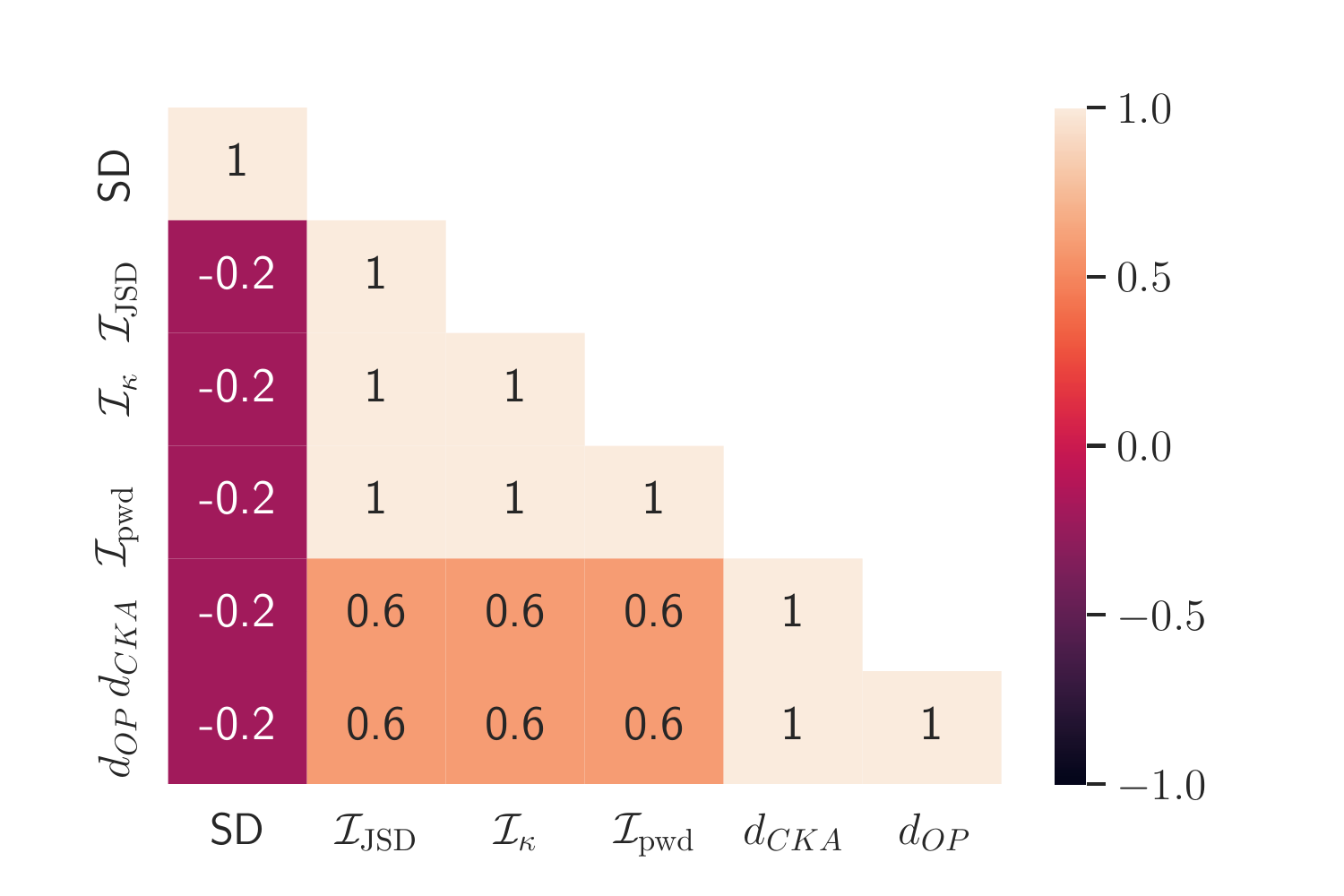}
        \caption{CoLA}
        \label{subfig:roberta_cola_tau}
    \end{subfigure}
    \caption{
        Kendall's $\tau$ values after applying different instability mitigation methods on RoBERTa,
        between each pair of measures.
        For representation measures, we take the value of the topmost layer.
    }
    \label{fig:app_roberta_tau}
\end{figure*}

\begin{figure*}
    \centering
    \begin{subfigure}[b]{0.32\textwidth}
        \centering
        \includegraphics[width=\textwidth]{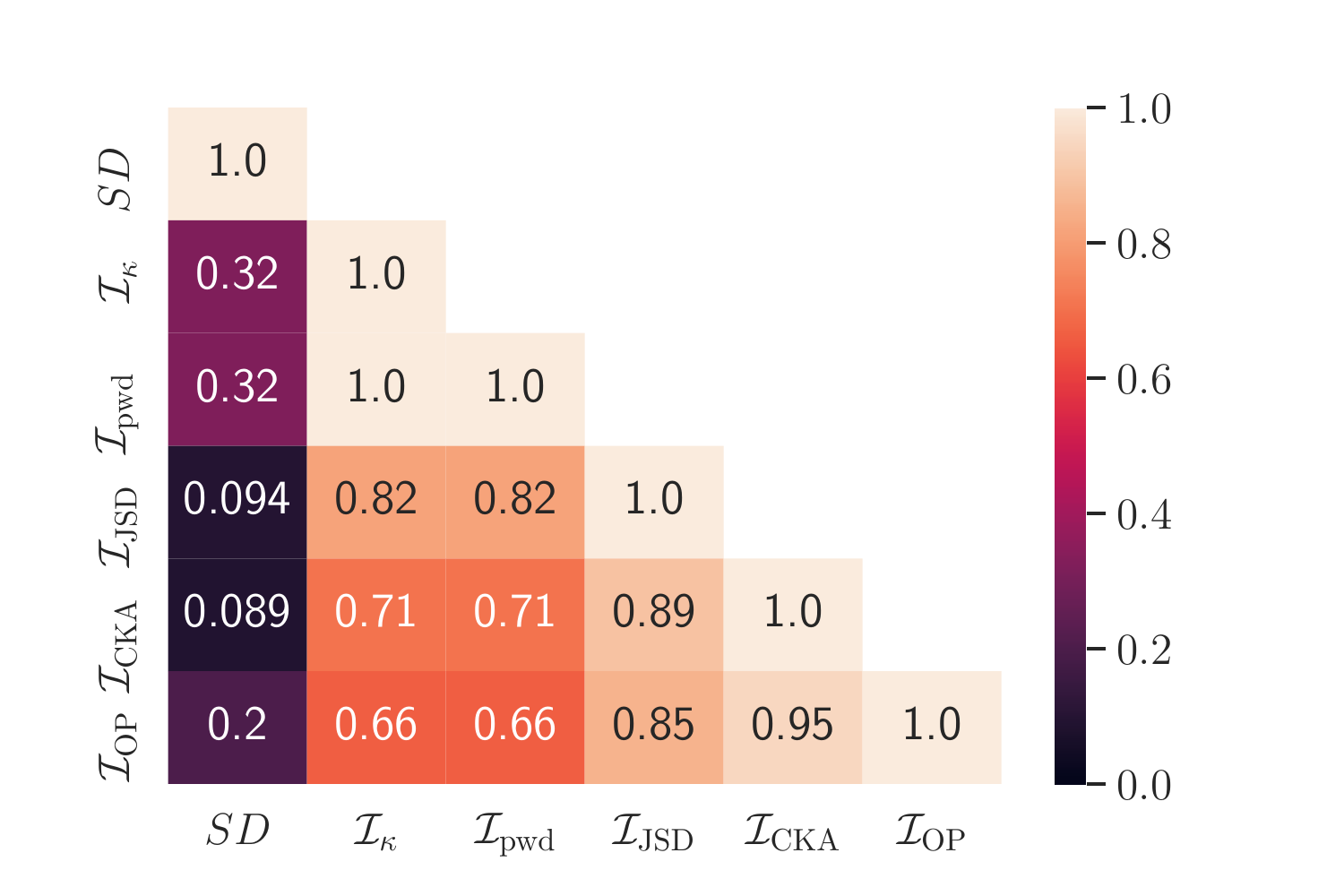}
        \caption{RTE}
        \label{subfig:bs_rte_bert_original}
    \end{subfigure}
    \hfill
    \begin{subfigure}[b]{0.32\textwidth}
        \centering
        \includegraphics[width=\textwidth]{fig/bs/mrpc_original_bert}
        \caption{MRPC}
        \label{subfig:bs_mrpc_bert_original}
    \end{subfigure}
    \hfill
    \begin{subfigure}[b]{0.32\textwidth}
        \centering
        \includegraphics[width=\textwidth]{fig/bs/cola_original_bert}
        \caption{CoLA}
        \label{subfig:bs_cola_bert_original}
    \end{subfigure}
    \caption{
        Bootstrapping results of \emph{Standard} BERT.
    }
    \label{fig:app_bs_bert_original}
\end{figure*}

\begin{figure*}
    \centering
    \begin{subfigure}[b]{0.32\textwidth}
        \centering
        \includegraphics[width=\textwidth]{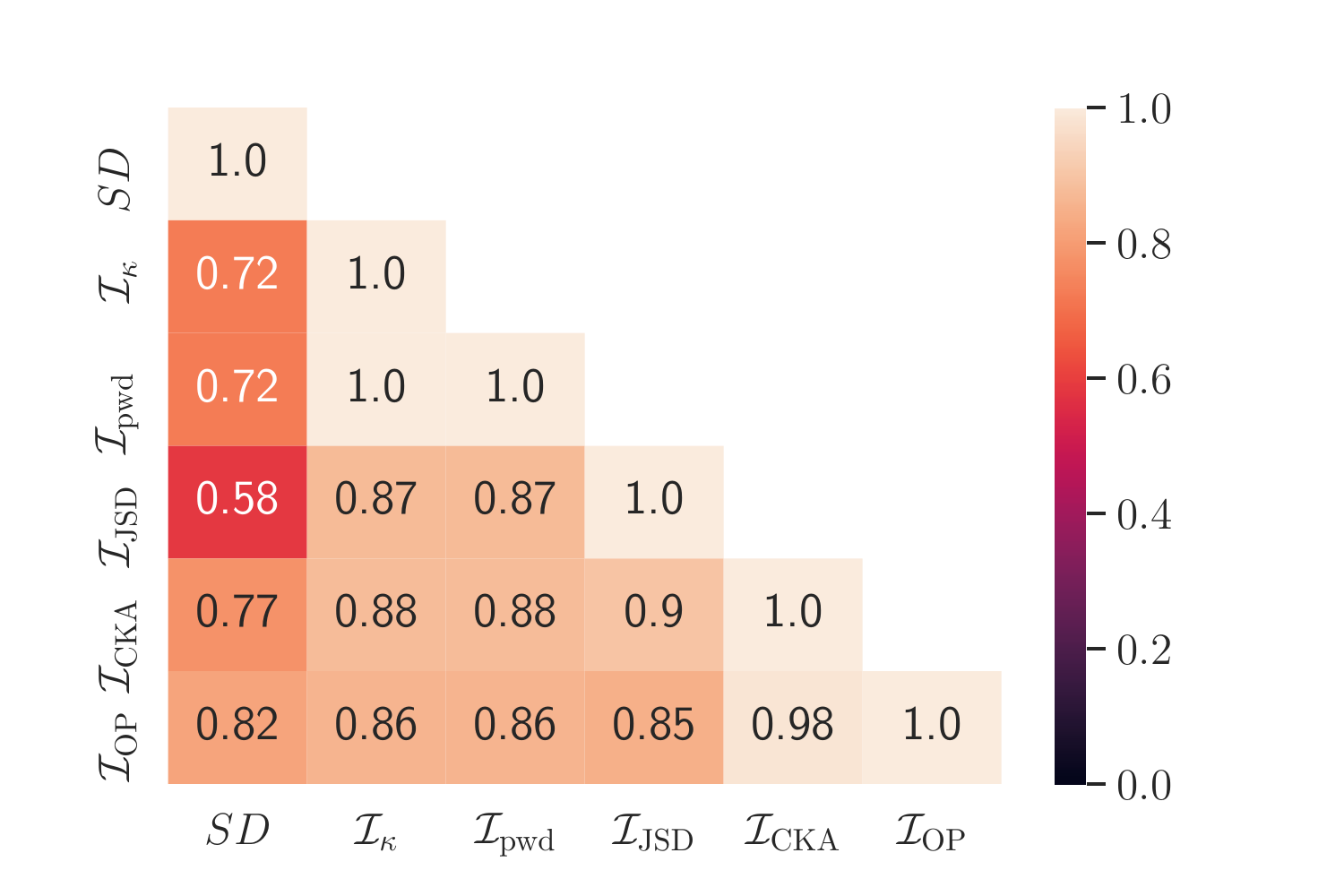}
        \caption{RTE}
        \label{subfig:bs_rte_mixout_bert}
    \end{subfigure}
    \hfill
    \begin{subfigure}[b]{0.32\textwidth}
        \centering
        \includegraphics[width=\textwidth]{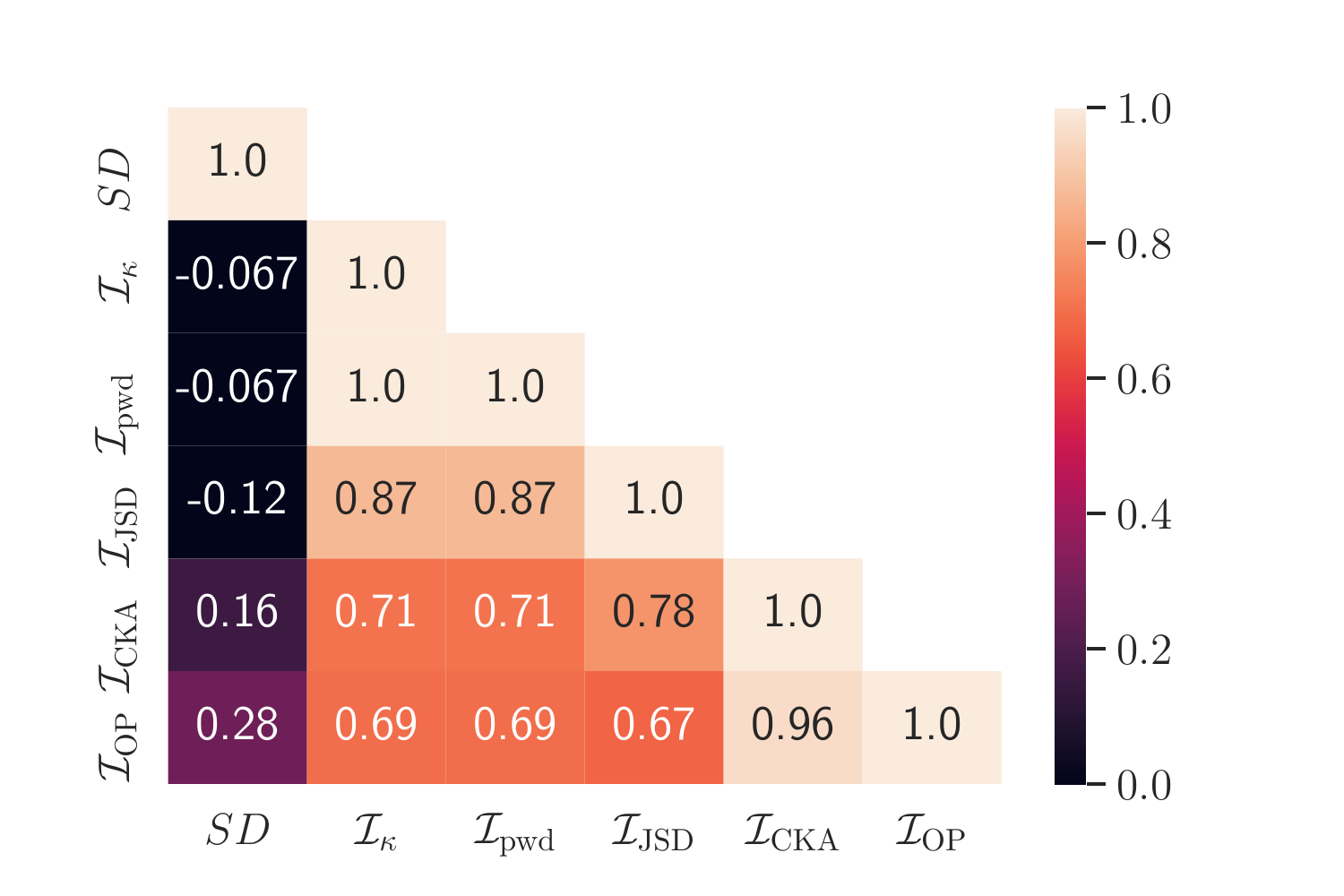}
        \caption{MRPC}
        \label{subfig:bs_mrpc_mixout_bert}
    \end{subfigure}
    \hfill
    \begin{subfigure}[b]{0.32\textwidth}
        \centering
        \includegraphics[width=\textwidth]{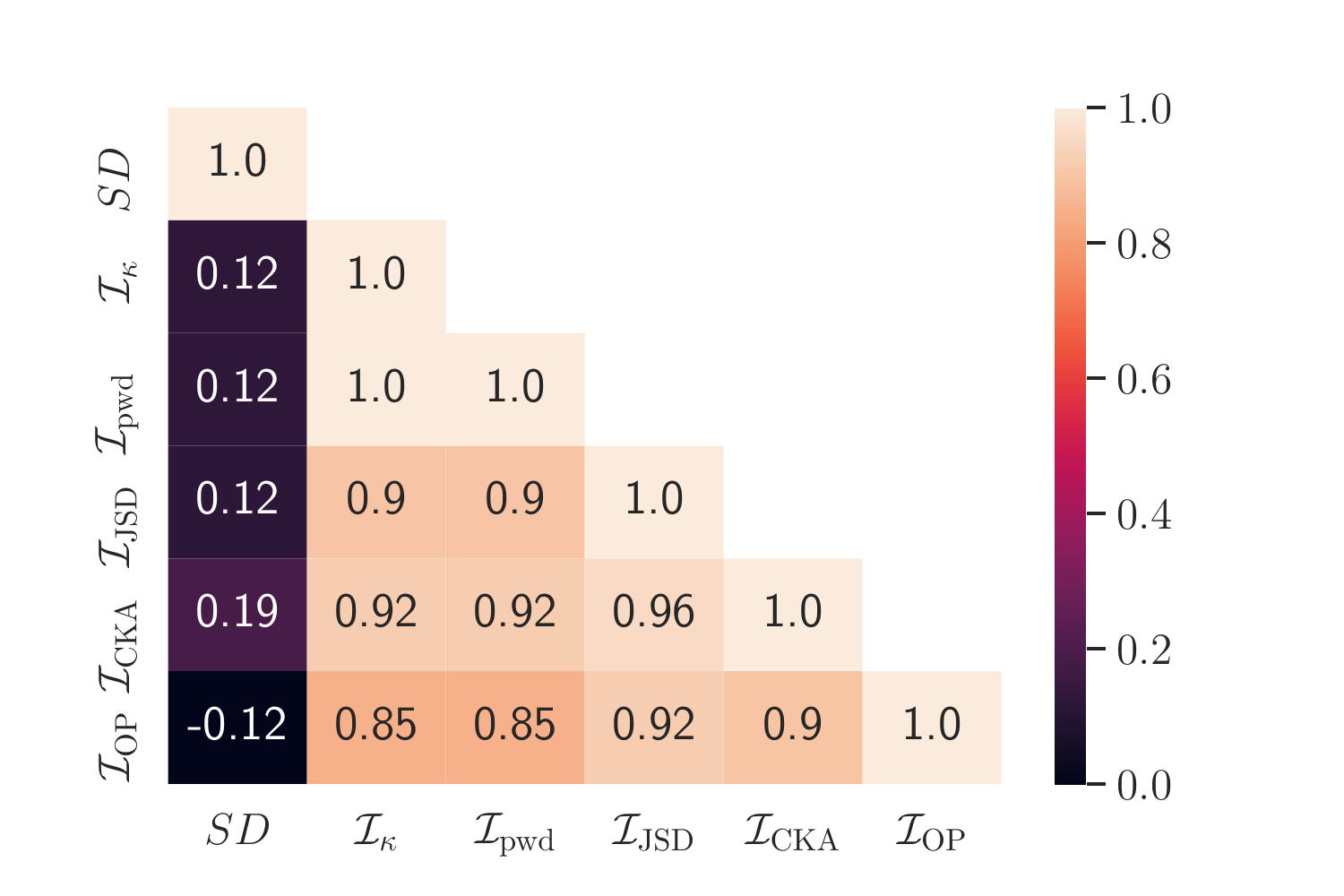}
        \caption{CoLA}
        \label{subfig:bs_cola_mixout_bert}
    \end{subfigure}
    \caption{
        Bootstrapping results of Mixout BERT.
    }
    \label{fig:app_bs_bert_mixout}
\end{figure*}

\begin{figure*}
    \centering
    \begin{subfigure}[b]{0.32\textwidth}
        \centering
        \includegraphics[width=\textwidth]{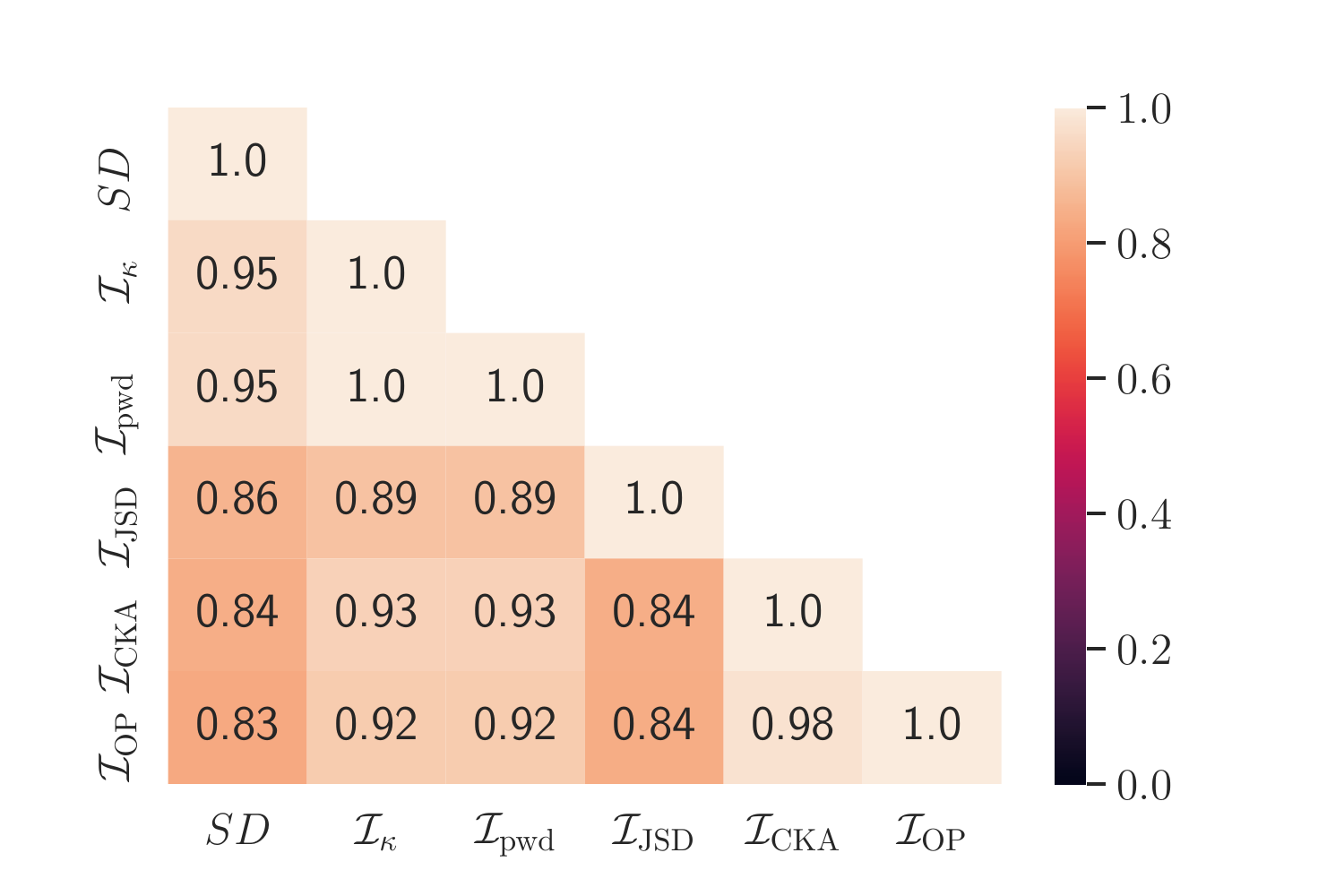}
        \caption{RTE}
        \label{subfig:bs_rte_wd_pre_bert}
    \end{subfigure}
    \hfill
    \begin{subfigure}[b]{0.32\textwidth}
        \centering
        \includegraphics[width=\textwidth]{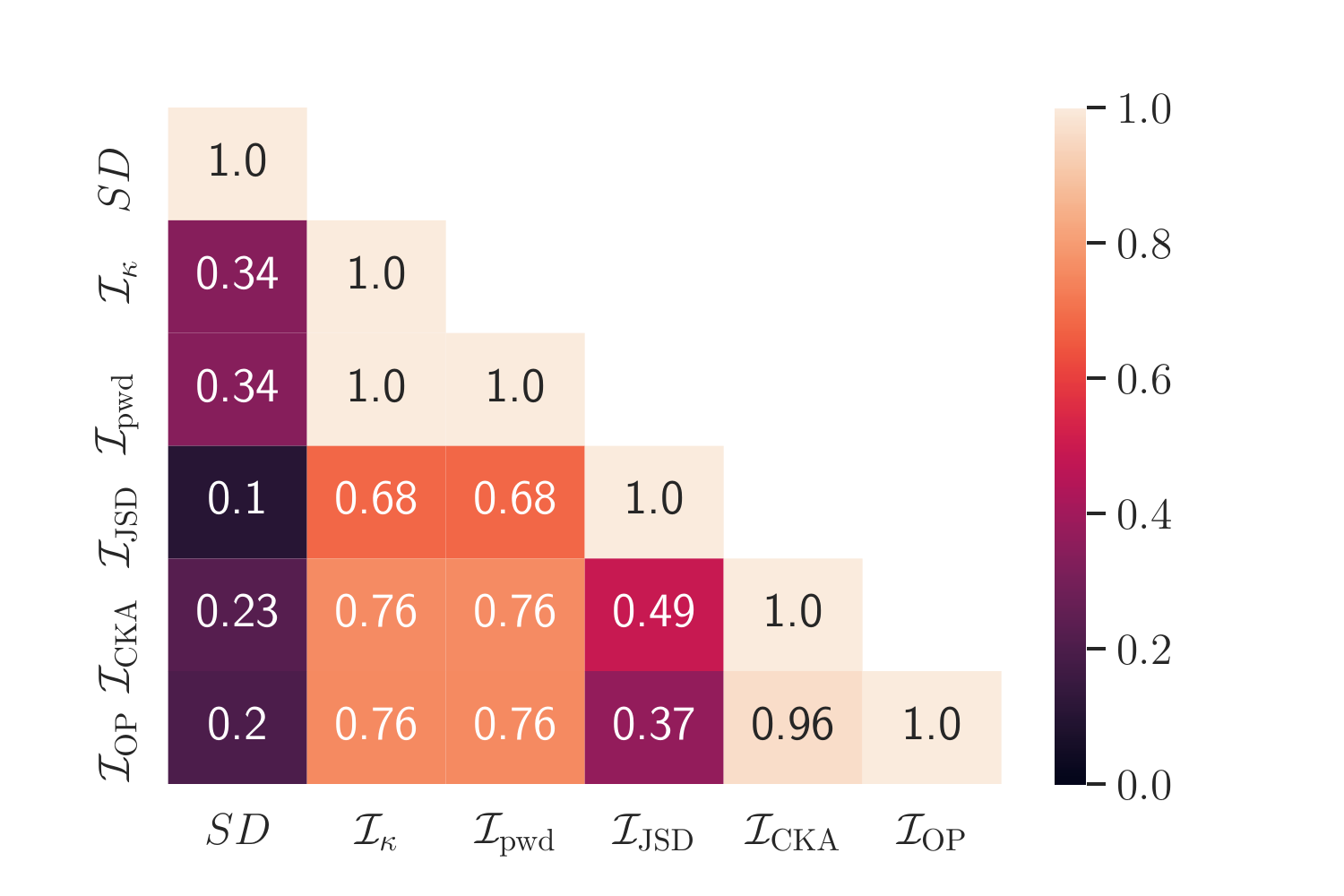}
        \caption{MRPC}
        \label{subfig:bs_mrpc_wd_pre_bert}
    \end{subfigure}
    \hfill
    \begin{subfigure}[b]{0.32\textwidth}
        \centering
        \includegraphics[width=\textwidth]{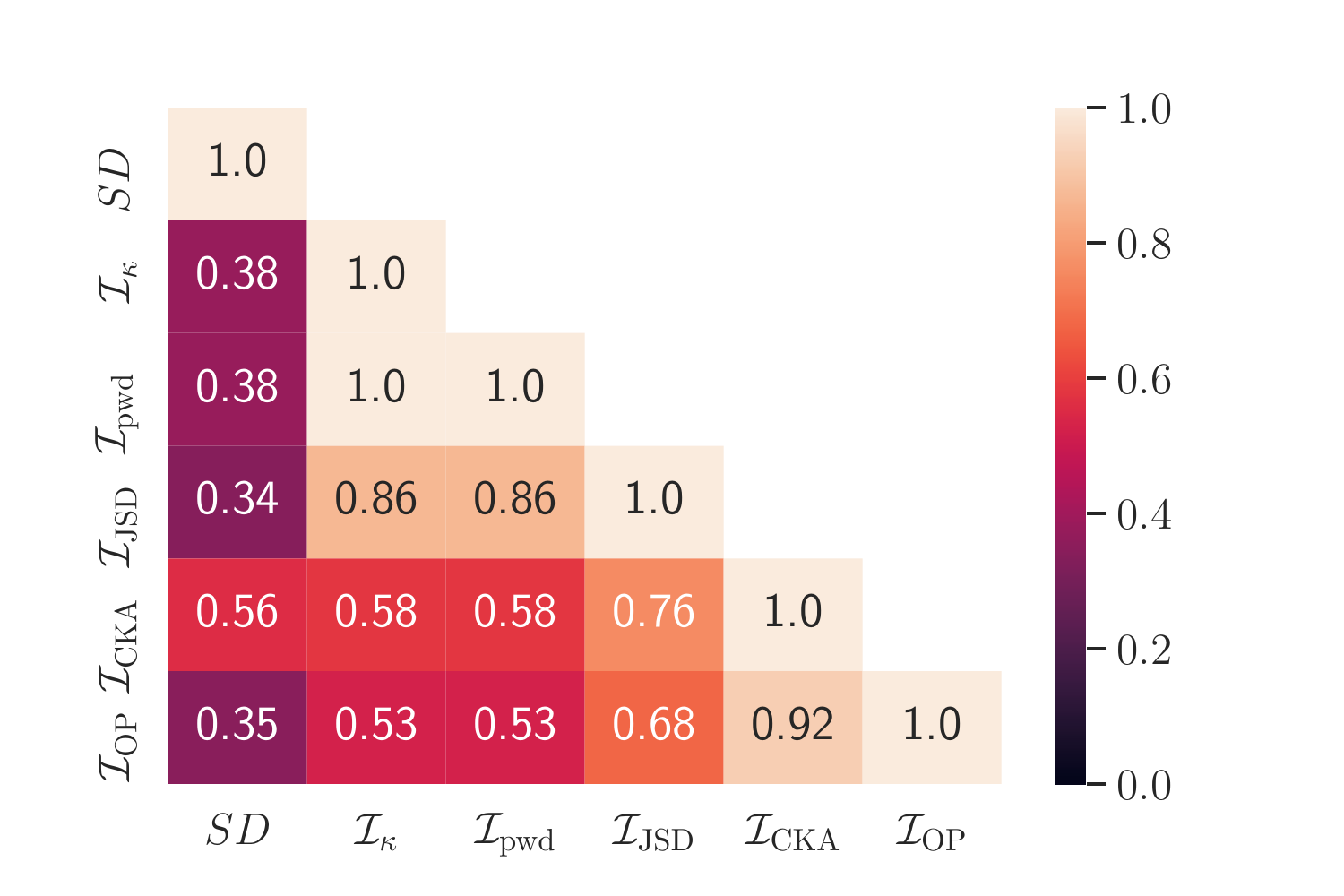}
        \caption{CoLA}
        \label{subfig:bs_cola_wd_pre_bert}
    \end{subfigure}
    \caption{
        Bootstrapping results of $\mathsf{WD_{pre}}$ BERT.
    }
    \label{fig:app_bs_bert_wd_pre}
\end{figure*}

\begin{figure*}
    \centering
    \begin{subfigure}[b]{0.32\textwidth}
        \centering
        \includegraphics[width=\textwidth]{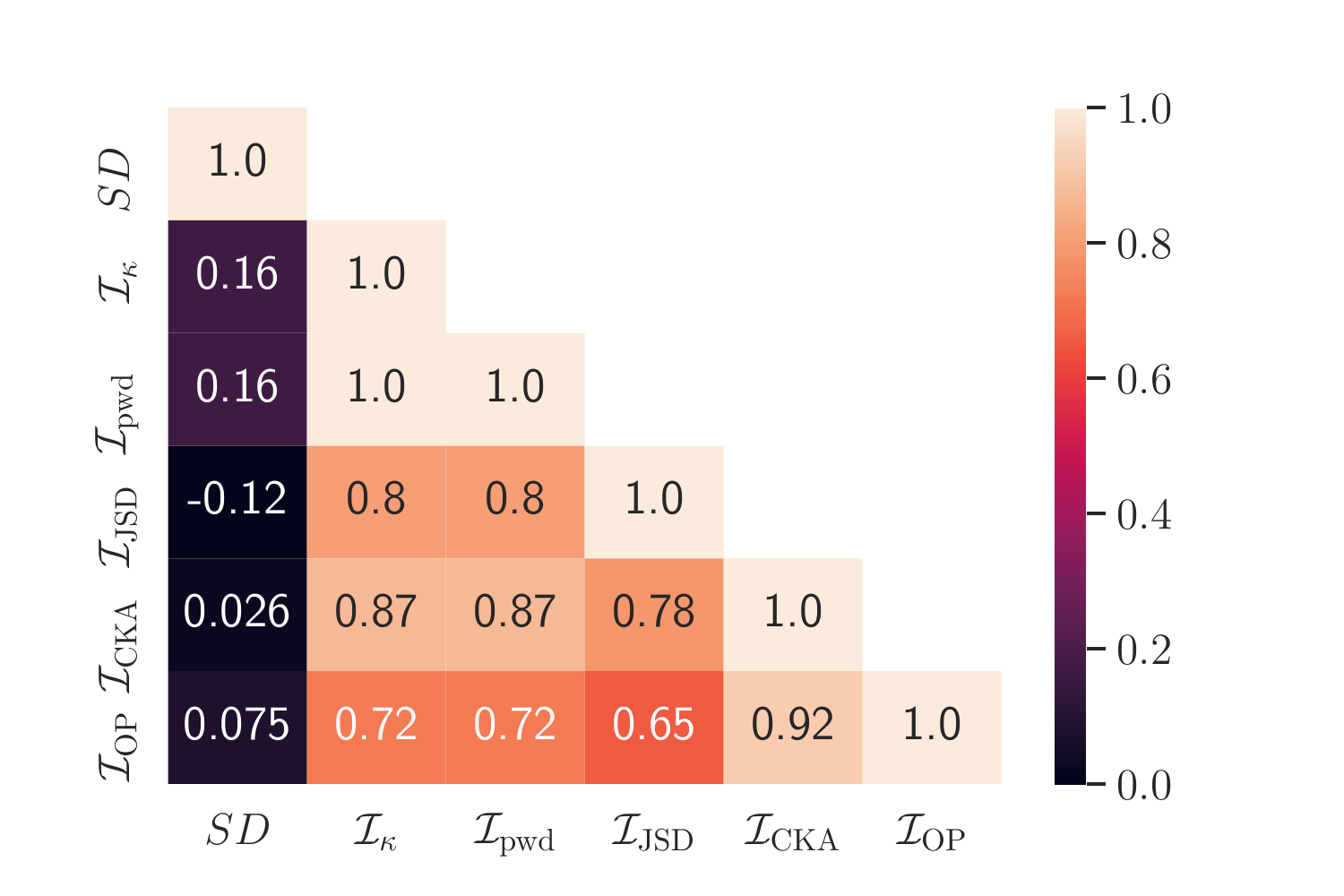}
        \caption{RTE}
        \label{subfig:bs_rte_llrd_bert}
    \end{subfigure}
    \hfill
    \begin{subfigure}[b]{0.32\textwidth}
        \centering
        \includegraphics[width=\textwidth]{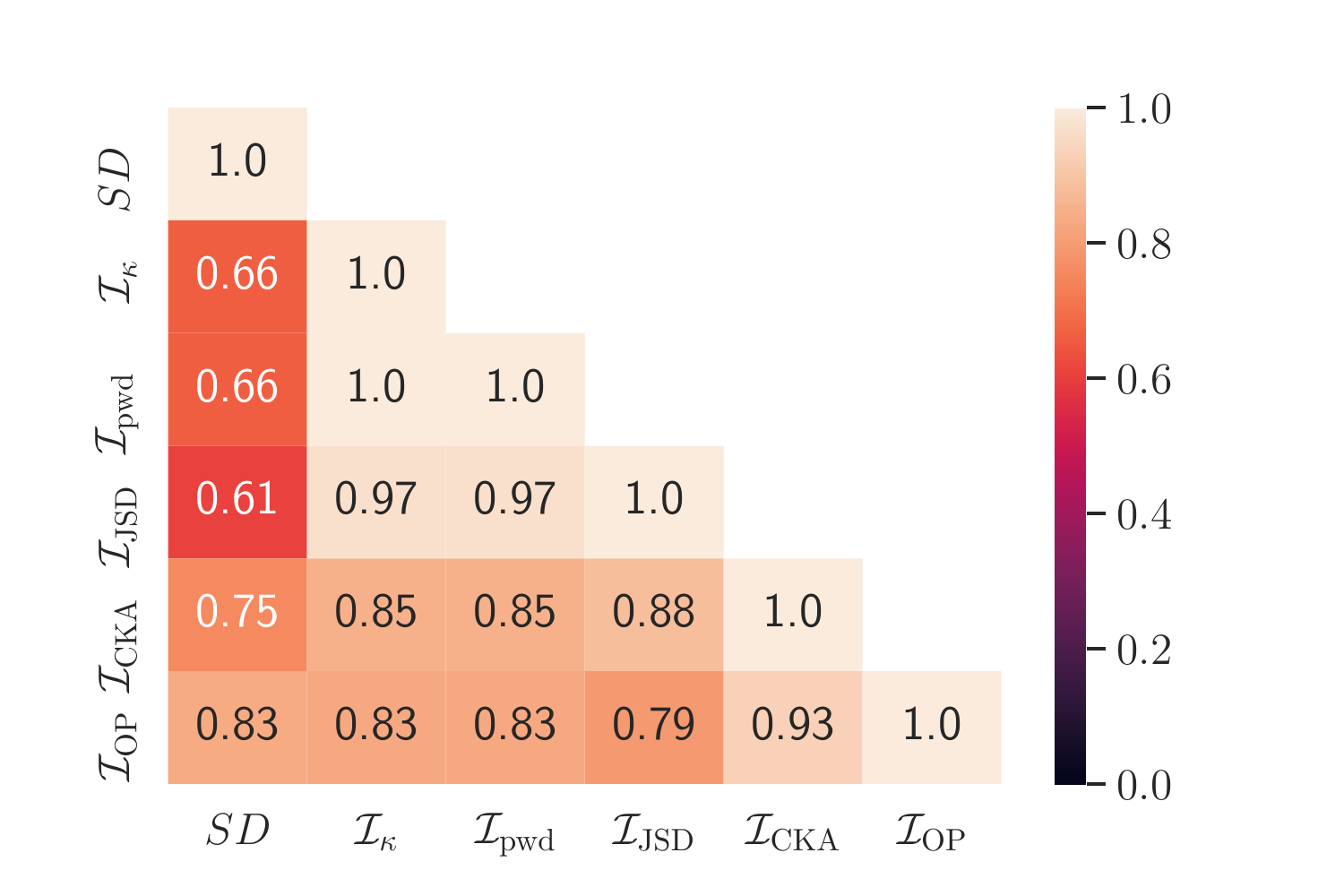}
        \caption{MRPC}
        \label{subfig:bs_mrpc_llrd_bert}
    \end{subfigure}
    \hfill
    \begin{subfigure}[b]{0.32\textwidth}
        \centering
        \includegraphics[width=\textwidth]{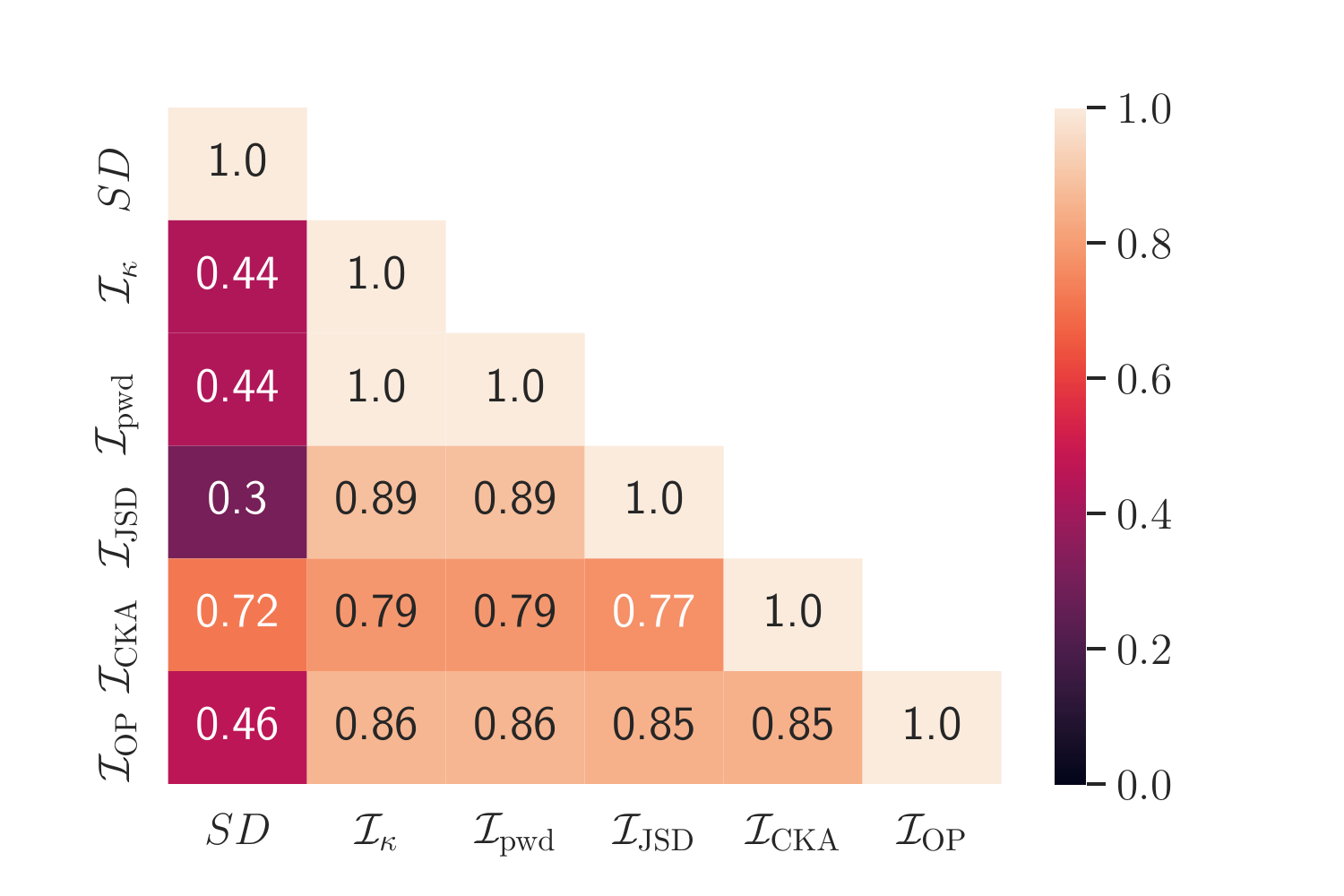}
        \caption{CoLA}
        \label{subfig:bs_cola_llrd_bert}
    \end{subfigure}
    \caption{
        Bootstrapping results of LLRD BERT.
    }
    \label{fig:app_bs_bert_llrd}
\end{figure*}

\begin{figure*}
    \centering
    \begin{subfigure}[b]{0.32\textwidth}
        \centering
        \includegraphics[width=\textwidth]{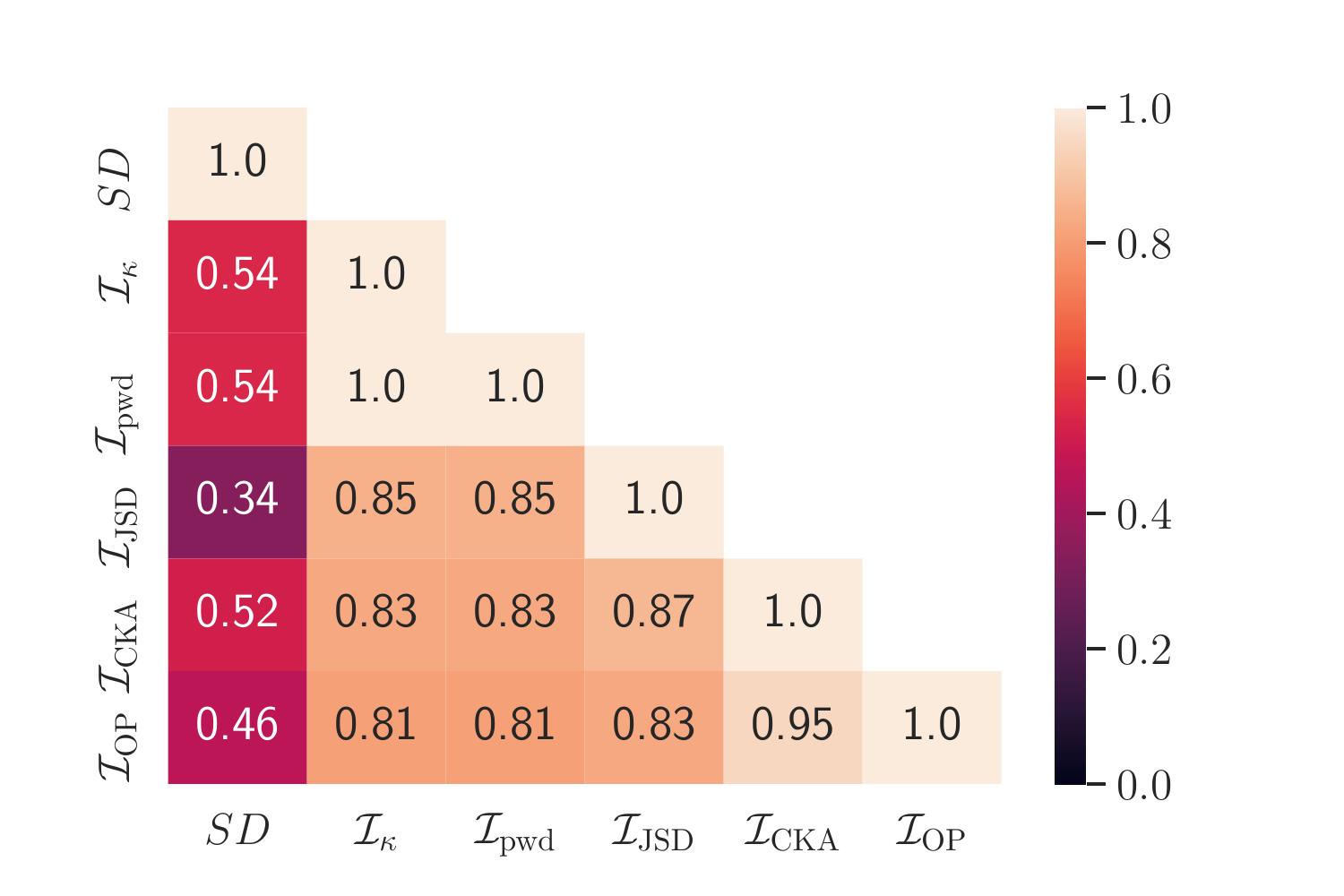}
        \caption{RTE}
        \label{subfig:bs_rte_reinit_bert}
    \end{subfigure}
    \hfill
    \begin{subfigure}[b]{0.32\textwidth}
        \centering
        \includegraphics[width=\textwidth]{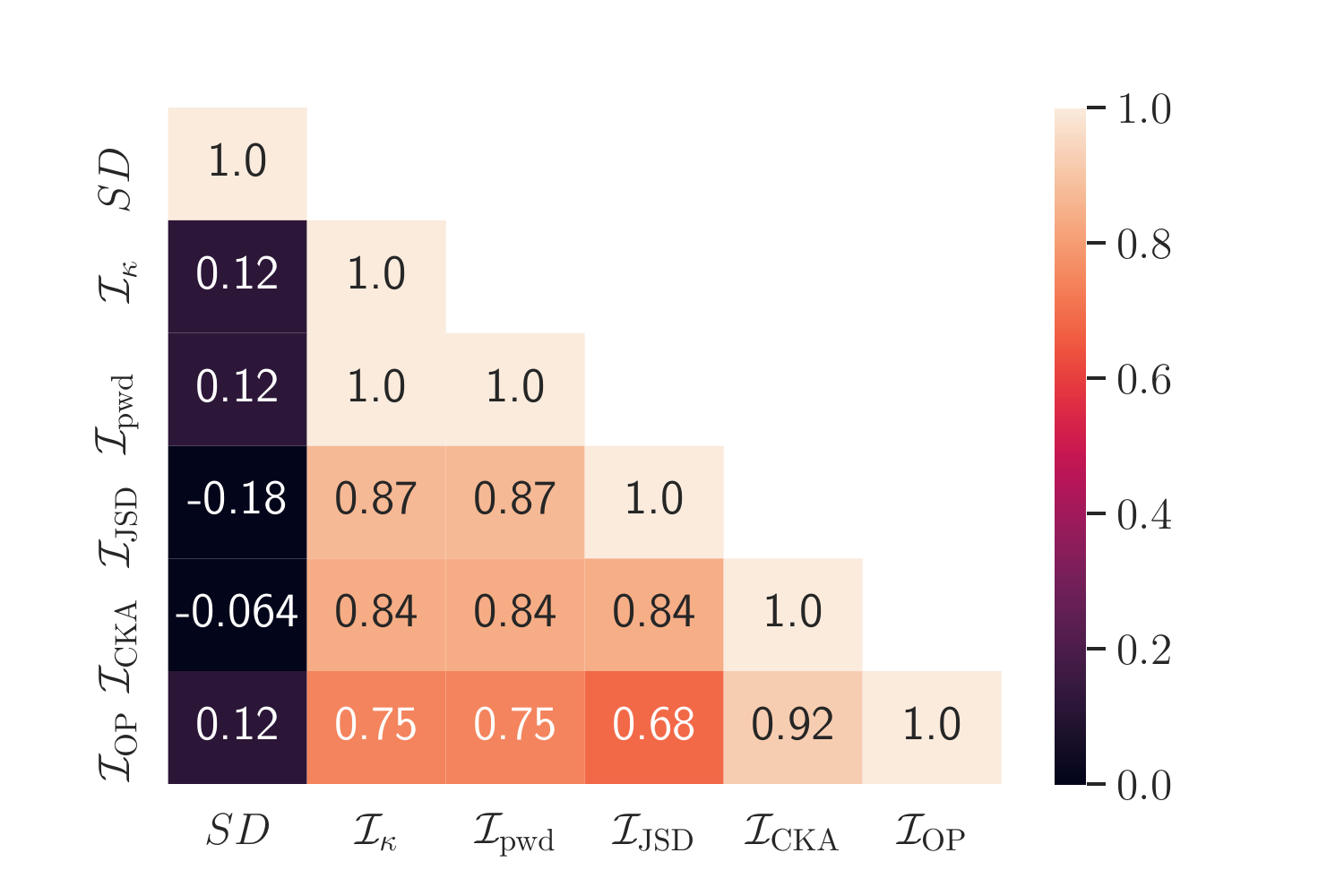}
        \caption{MRPC}
        \label{subfig:bs_mrpc_reinit_bert}
    \end{subfigure}
    \hfill
    \begin{subfigure}[b]{0.32\textwidth}
        \centering
        \includegraphics[width=\textwidth]{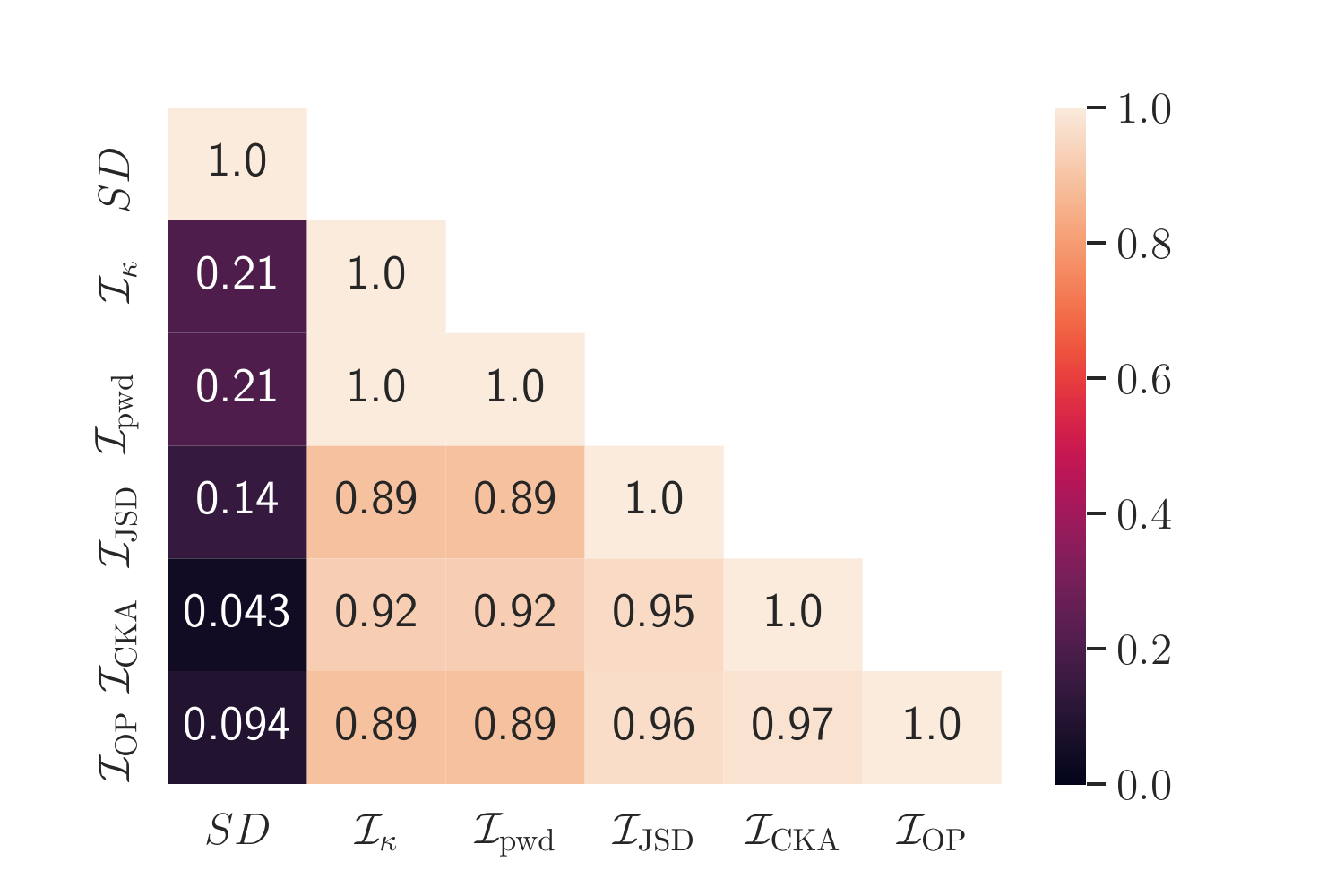}
        \caption{CoLA}
        \label{subfig:bs_cola_reinit_bert}
    \end{subfigure}
    \caption{
        Bootstrapping results of Re-init BERT.
    }
    \label{fig:app_bs_bert_reinit}
\end{figure*}

\begin{figure*}
    \centering
    \begin{subfigure}[b]{0.32\textwidth}
        \centering
        \includegraphics[width=\textwidth]{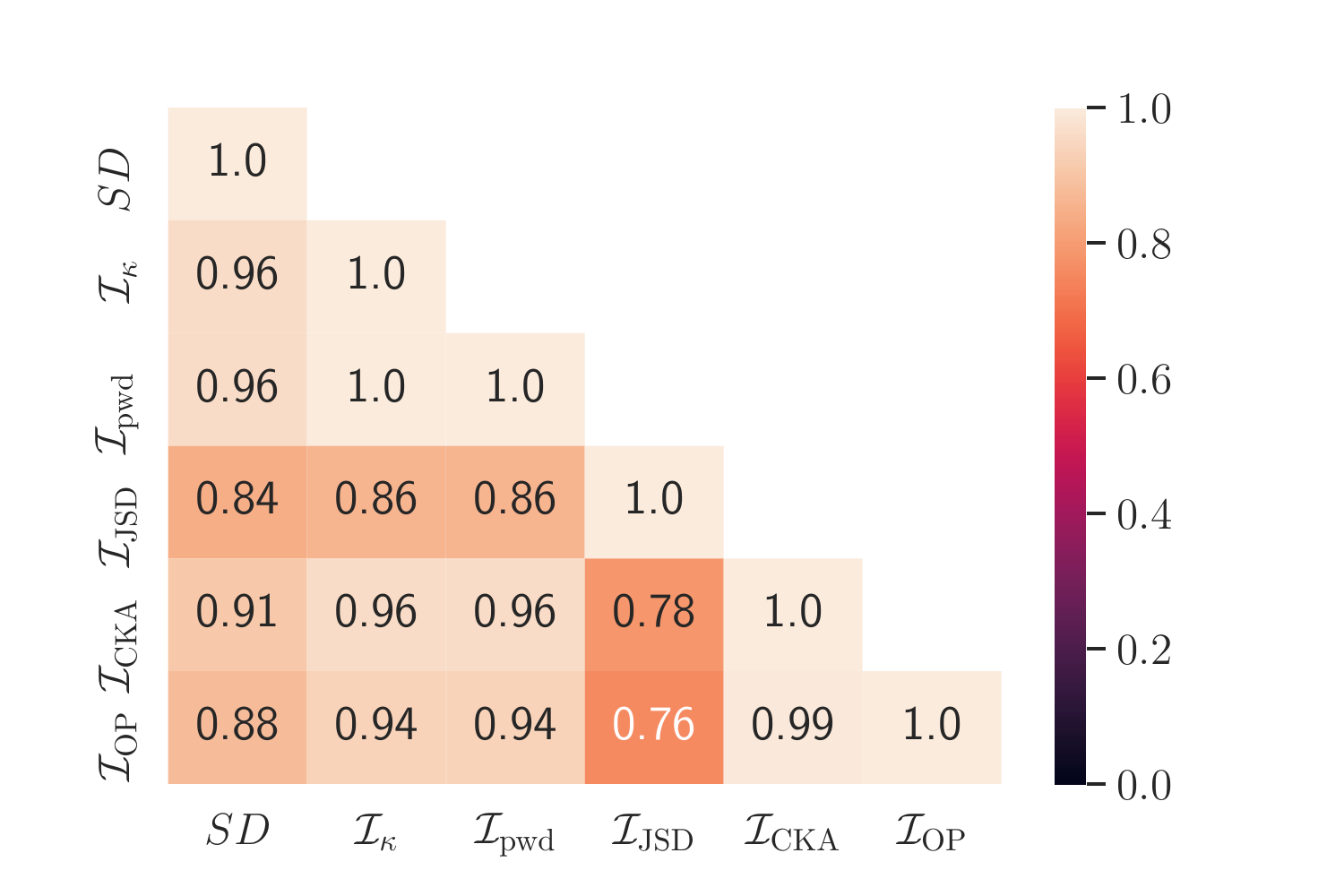}
        \caption{RTE}
        \label{subfig:bs_rte_roberta_original}
    \end{subfigure}
    \hfill
    \begin{subfigure}[b]{0.32\textwidth}
        \centering
        \includegraphics[width=\textwidth]{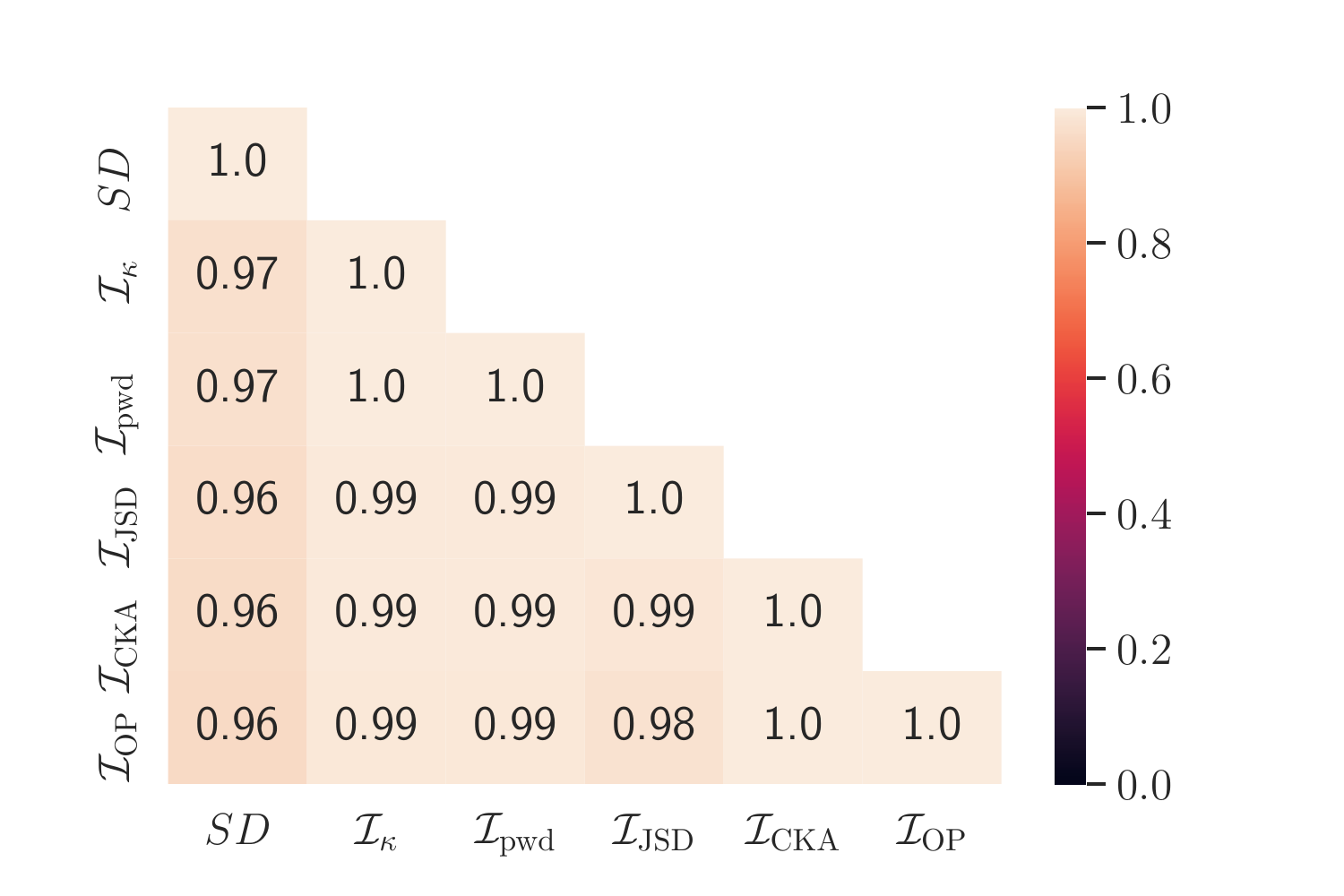}
        \caption{MRPC}
        \label{subfig:bs_mrpc_roberta_original}
    \end{subfigure}
    \hfill
    \begin{subfigure}[b]{0.32\textwidth}
        \centering
        \includegraphics[width=\textwidth]{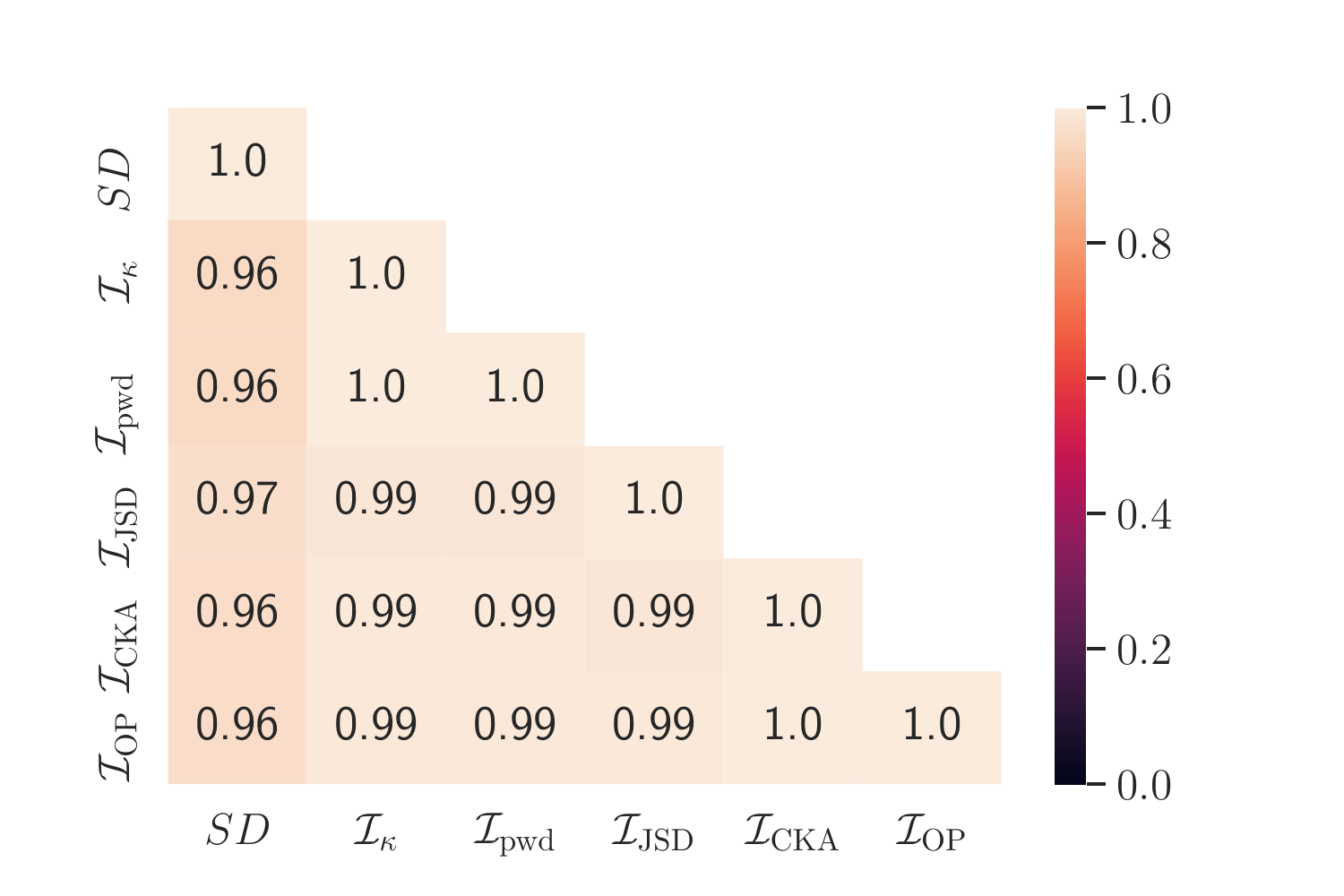}
        \caption{CoLA}
        \label{subfig:bs_cola_roberta_original}
    \end{subfigure}
    \caption{
        Bootstrapping results of \emph{Standard} RoBERTa.
    }
    \label{fig:app_bs_roberta_original}
\end{figure*}

\begin{figure*}
    \centering
    \begin{subfigure}[b]{0.32\textwidth}
        \centering
        \includegraphics[width=\textwidth]{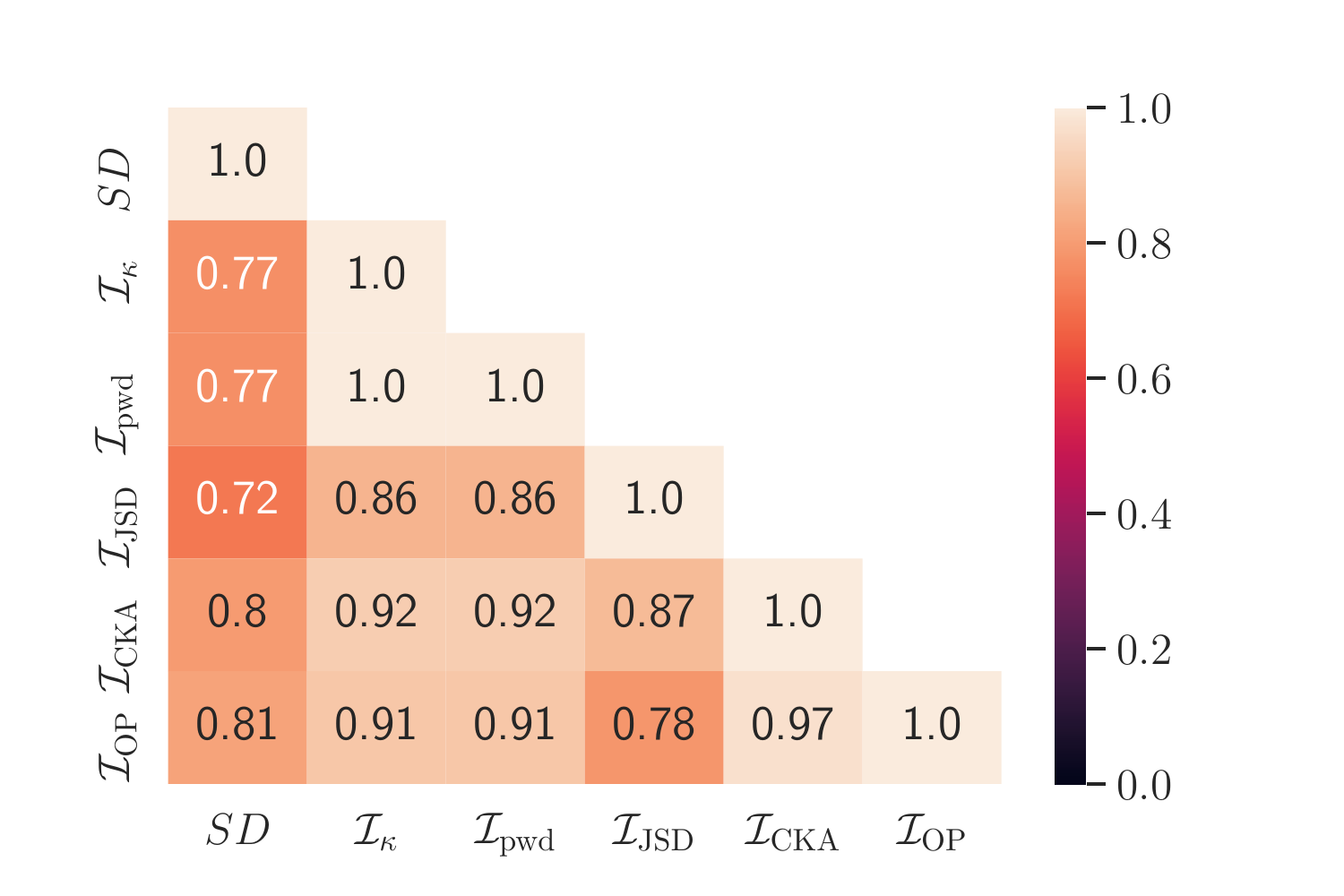}
        \caption{RTE}
        \label{subfig:bs_rte_mixout_roberta}
    \end{subfigure}
    \hfill
    \begin{subfigure}[b]{0.32\textwidth}
        \centering
        \includegraphics[width=\textwidth]{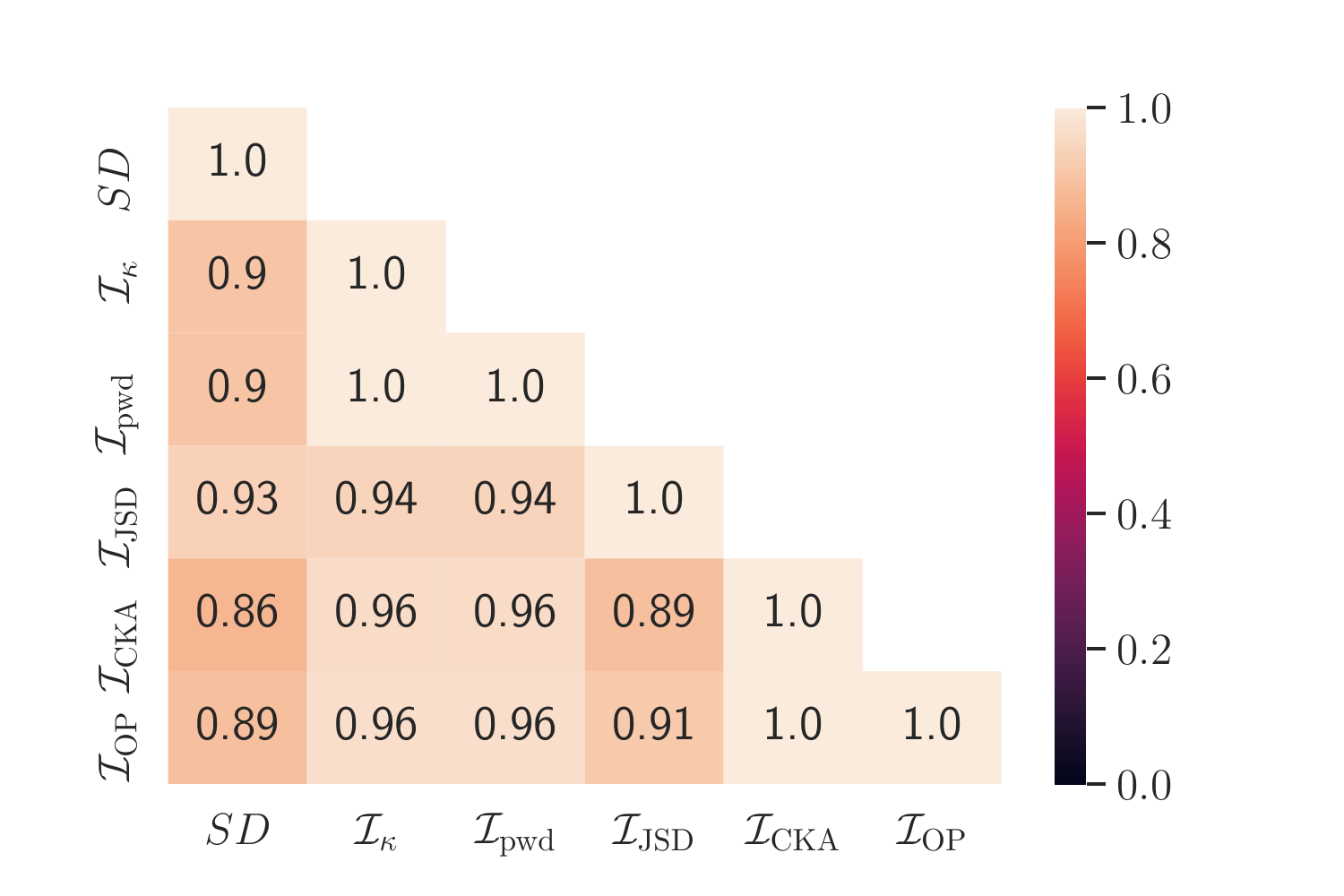}
        \caption{MRPC}
        \label{subfig:bs_mrpc_mixout_roberta}
    \end{subfigure}
    \hfill
    \begin{subfigure}[b]{0.32\textwidth}
        \centering
        \includegraphics[width=\textwidth]{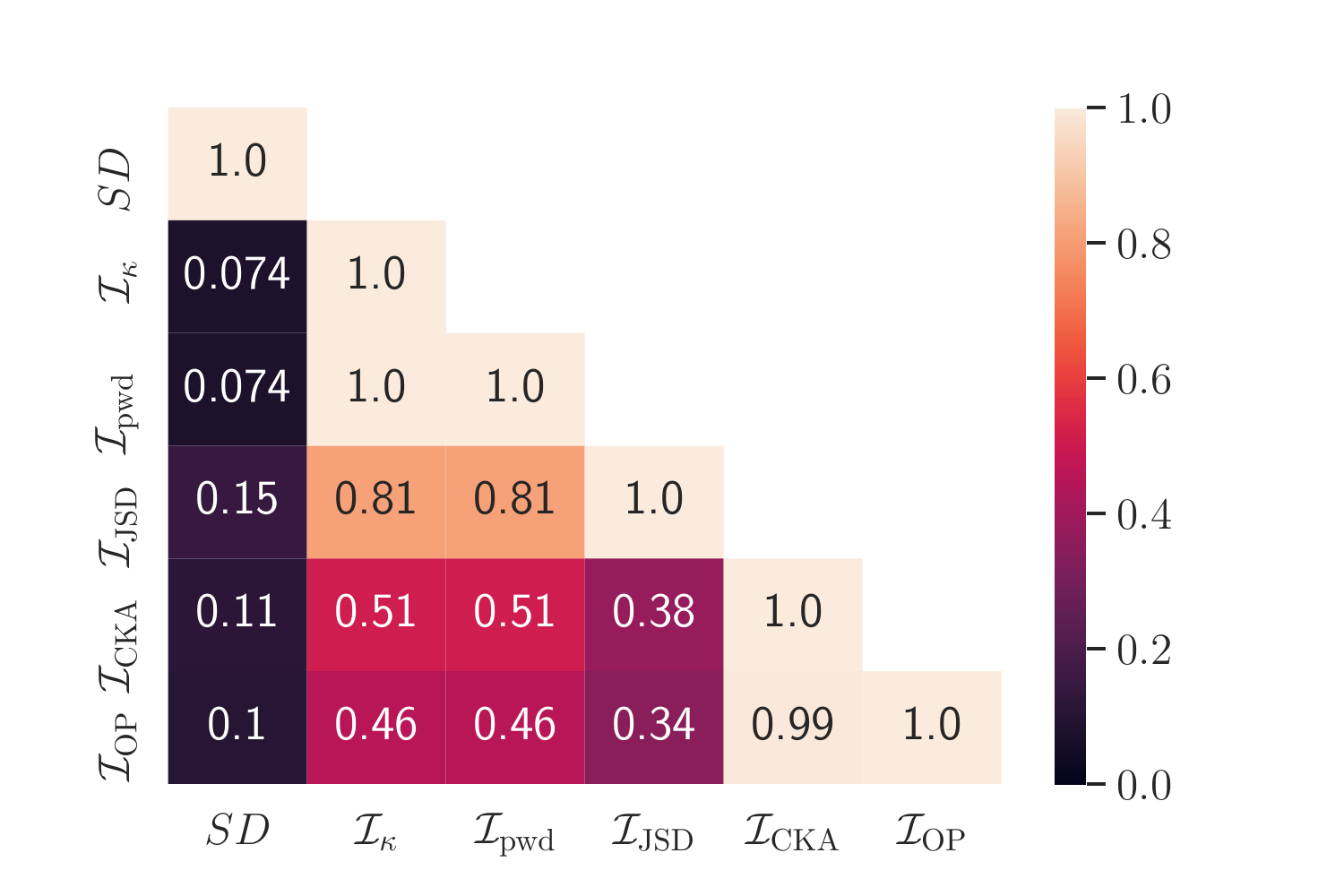}
        \caption{CoLA}
        \label{subfig:bs_cola_mixout_roberta}
    \end{subfigure}
    \caption{
        Bootstrapping results of Mixout RoBERTa.
    }
    \label{fig:app_bs_roberta_mixout}
\end{figure*}

\begin{figure*}
    \centering
    \begin{subfigure}[b]{0.32\textwidth}
        \centering
        \includegraphics[width=\textwidth]{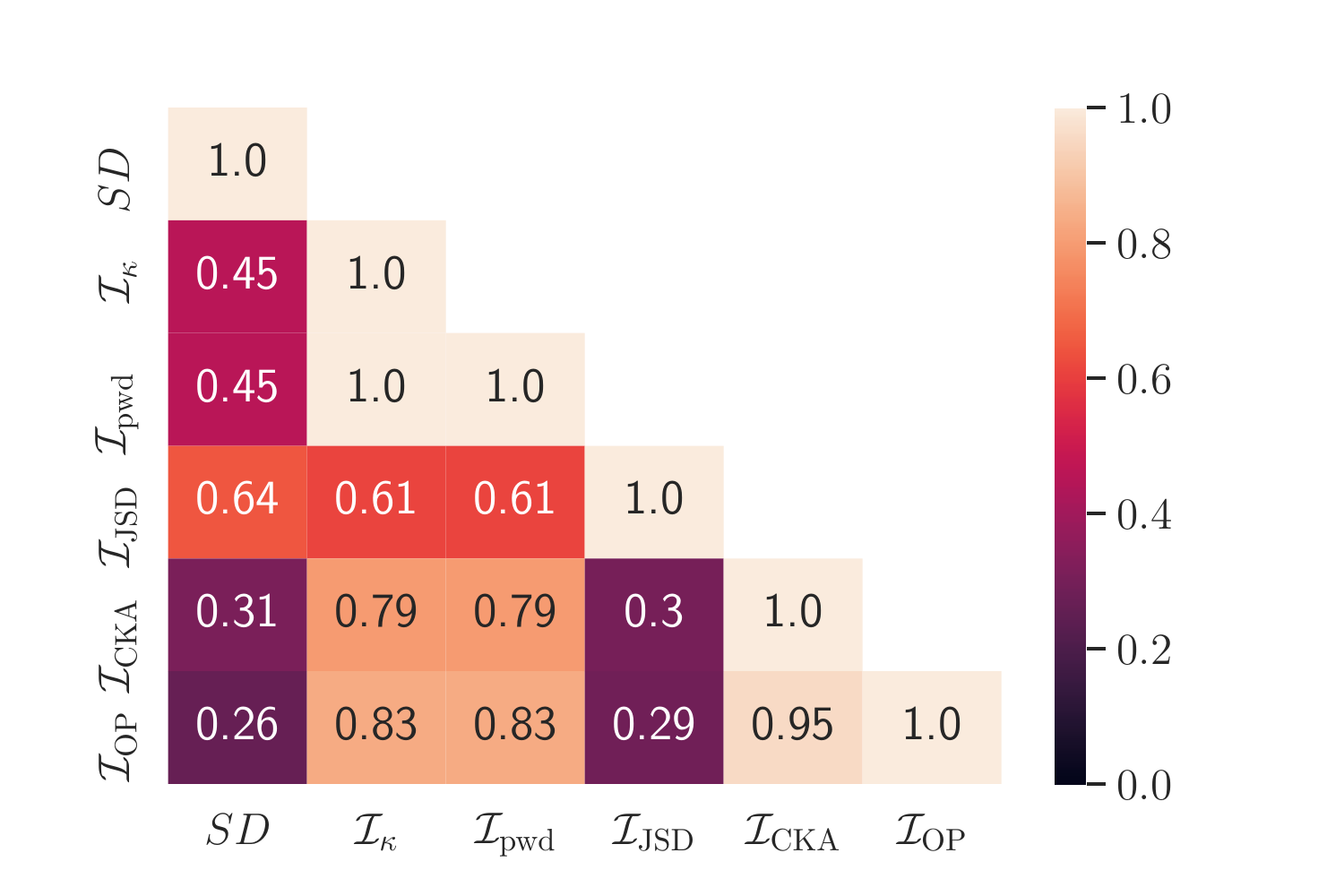}
        \caption{RTE}
        \label{subfig:bs_rte_wd_pre_roberta}
    \end{subfigure}
    \hfill
    \begin{subfigure}[b]{0.32\textwidth}
        \centering
        \includegraphics[width=\textwidth]{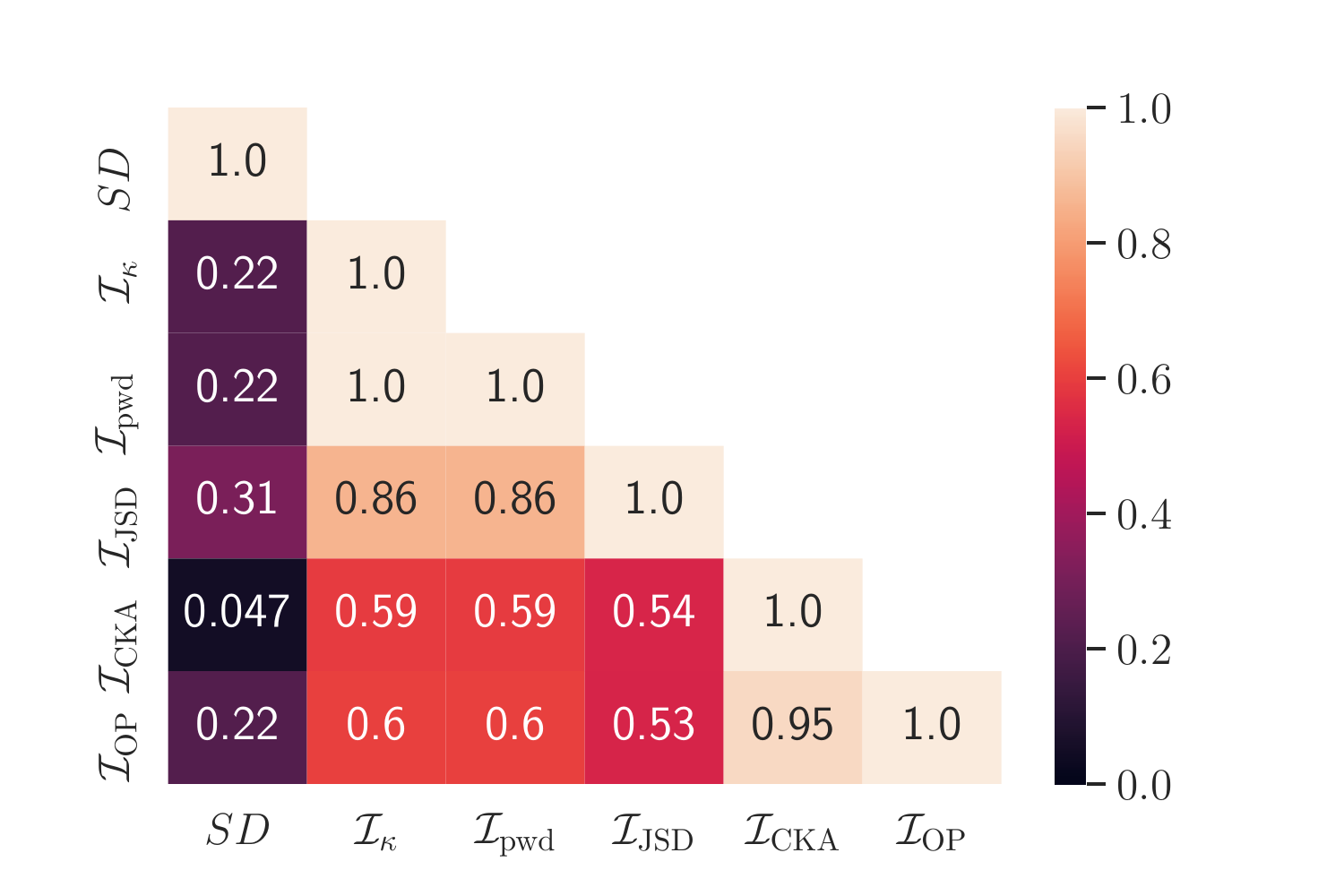}
        \caption{MRPC}
        \label{subfig:bs_mrpc_wd_pre_roberta}
    \end{subfigure}
    \hfill
    \begin{subfigure}[b]{0.32\textwidth}
        \centering
        \includegraphics[width=\textwidth]{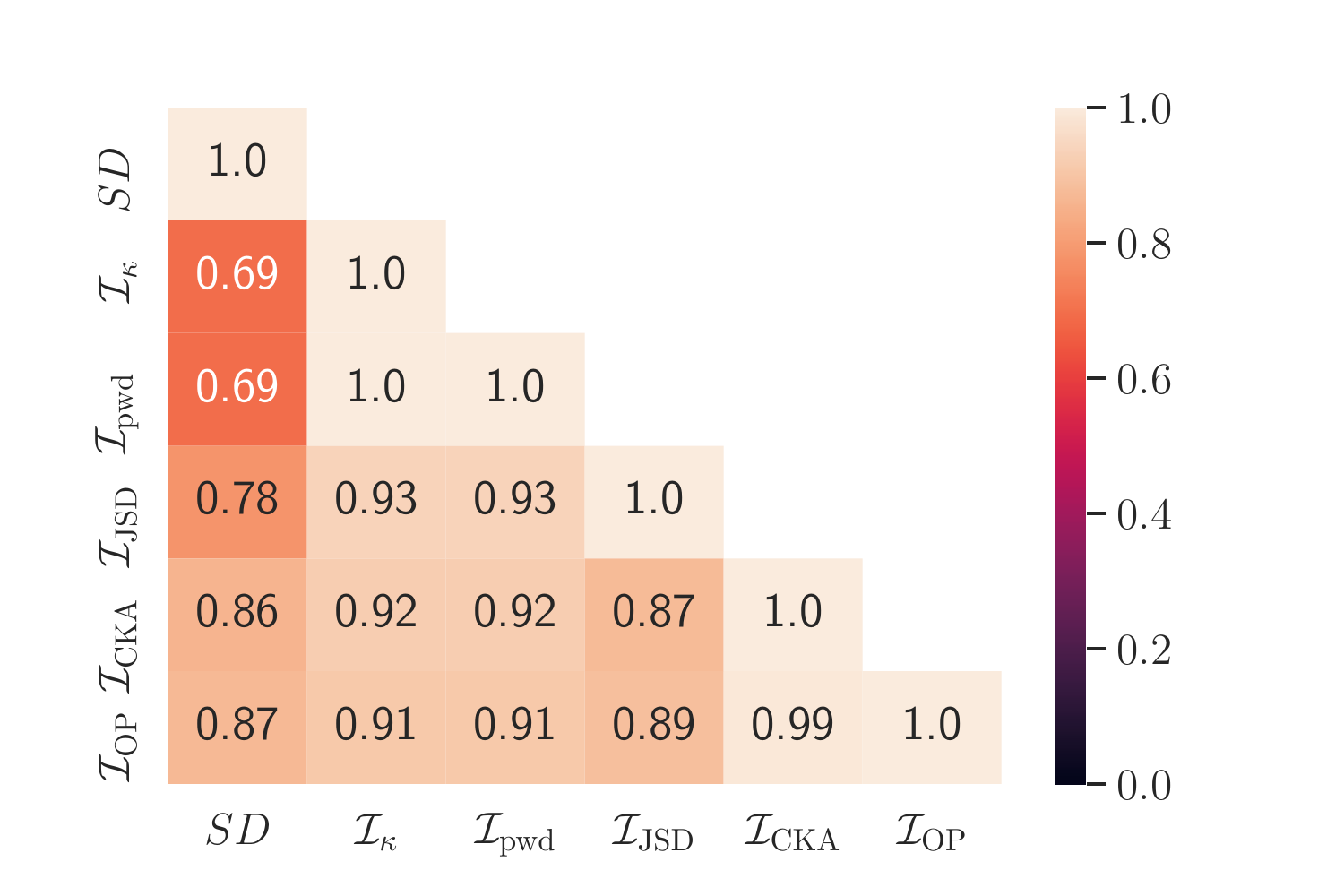}
        \caption{CoLA}
        \label{subfig:bs_cola_wd_pre_roberta}
    \end{subfigure}
    \caption{
        Bootstrapping results of $\mathsf{WD_{pre}}$ RoBERTa.
    }
    \label{fig:app_bs_roberta_wd_pre}
\end{figure*}

\begin{figure*}
    \centering
    \begin{subfigure}[b]{0.32\textwidth}
        \centering
        \includegraphics[width=\textwidth]{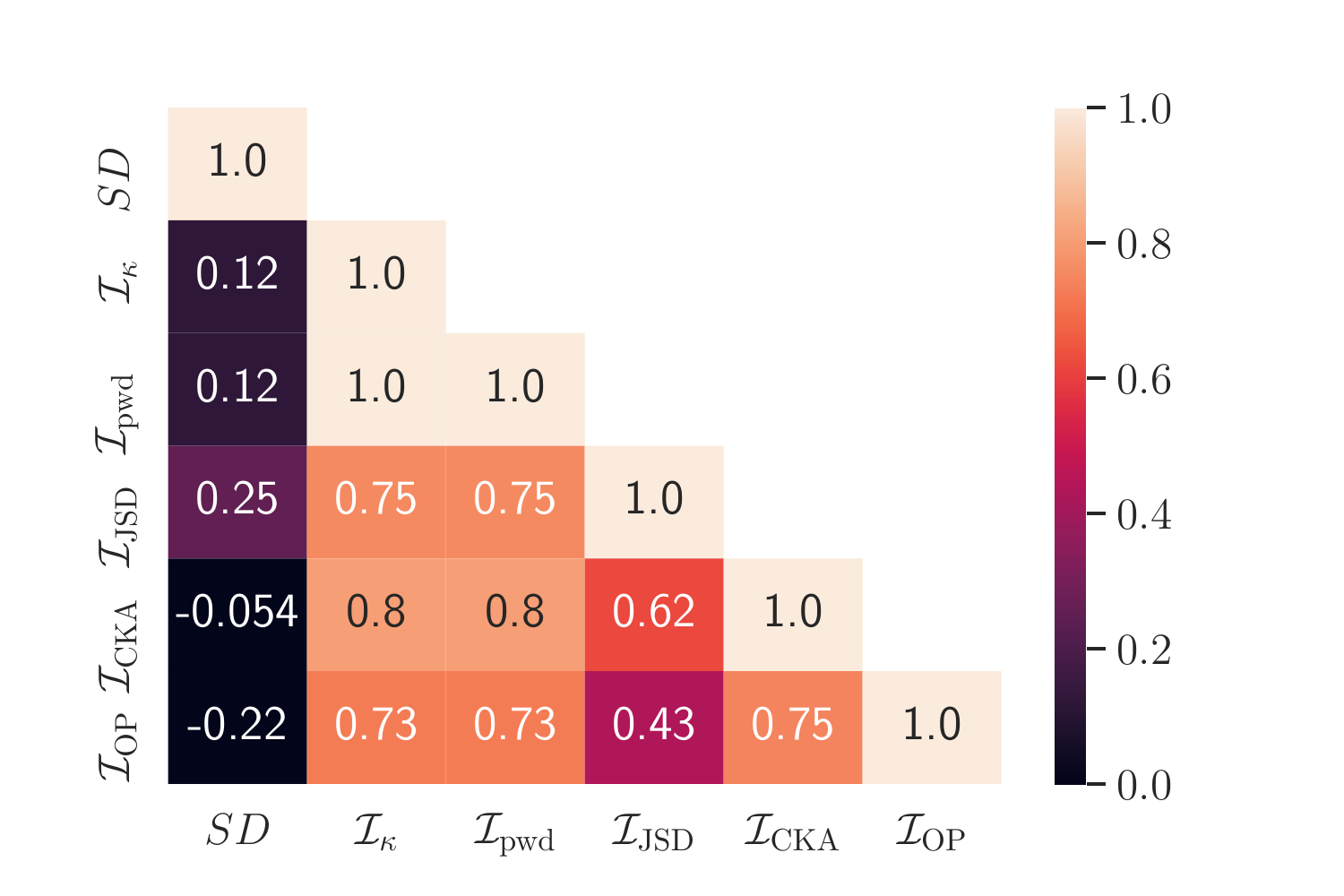}
        \caption{RTE}
        \label{subfig:bs_rte_llrd_roberta}
    \end{subfigure}
    \hfill
    \begin{subfigure}[b]{0.32\textwidth}
        \centering
        \includegraphics[width=\textwidth]{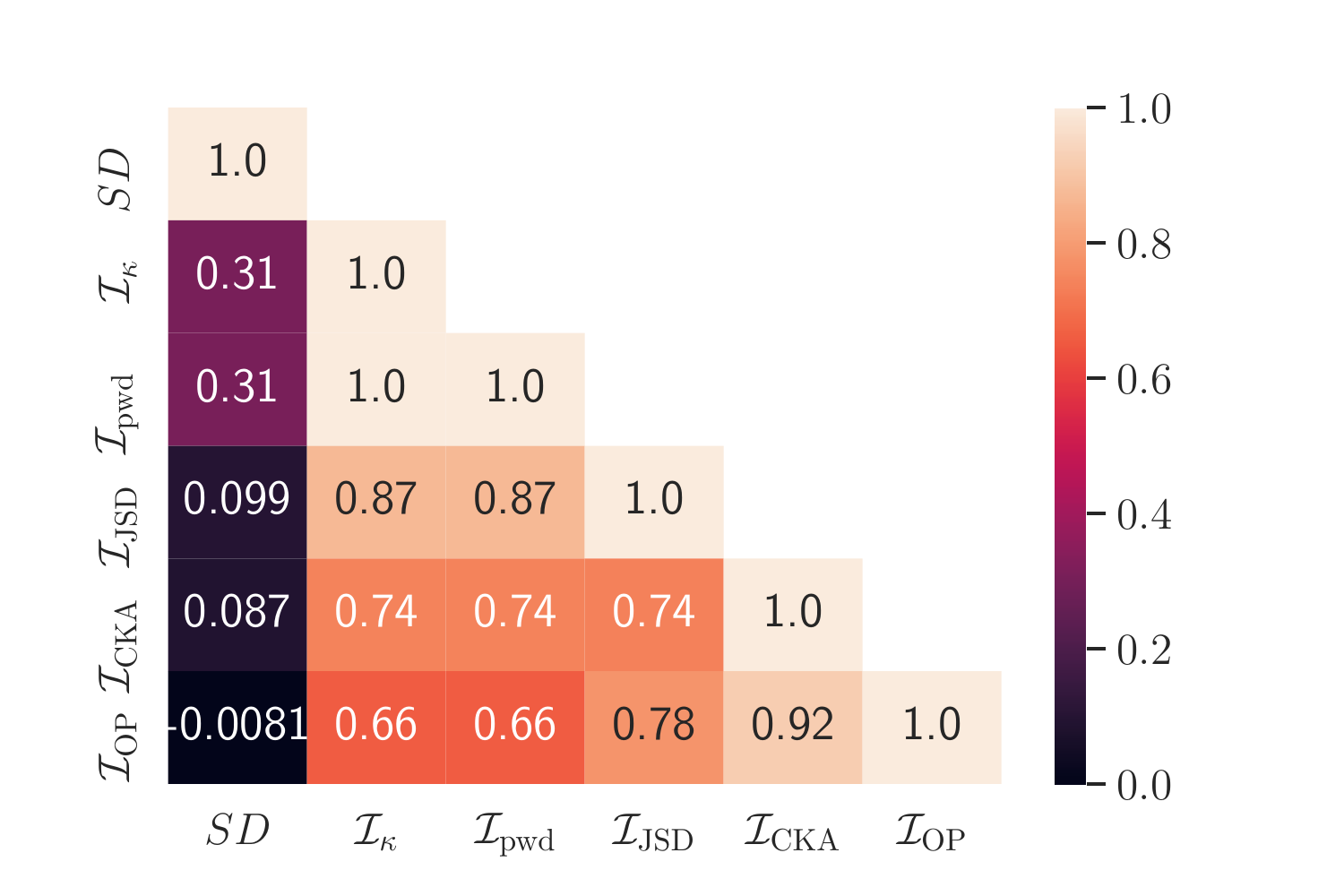}
        \caption{MRPC}
        \label{subfig:bs_mrpc_llrd_roberta}
    \end{subfigure}
    \hfill
    \begin{subfigure}[b]{0.32\textwidth}
        \centering
        \includegraphics[width=\textwidth]{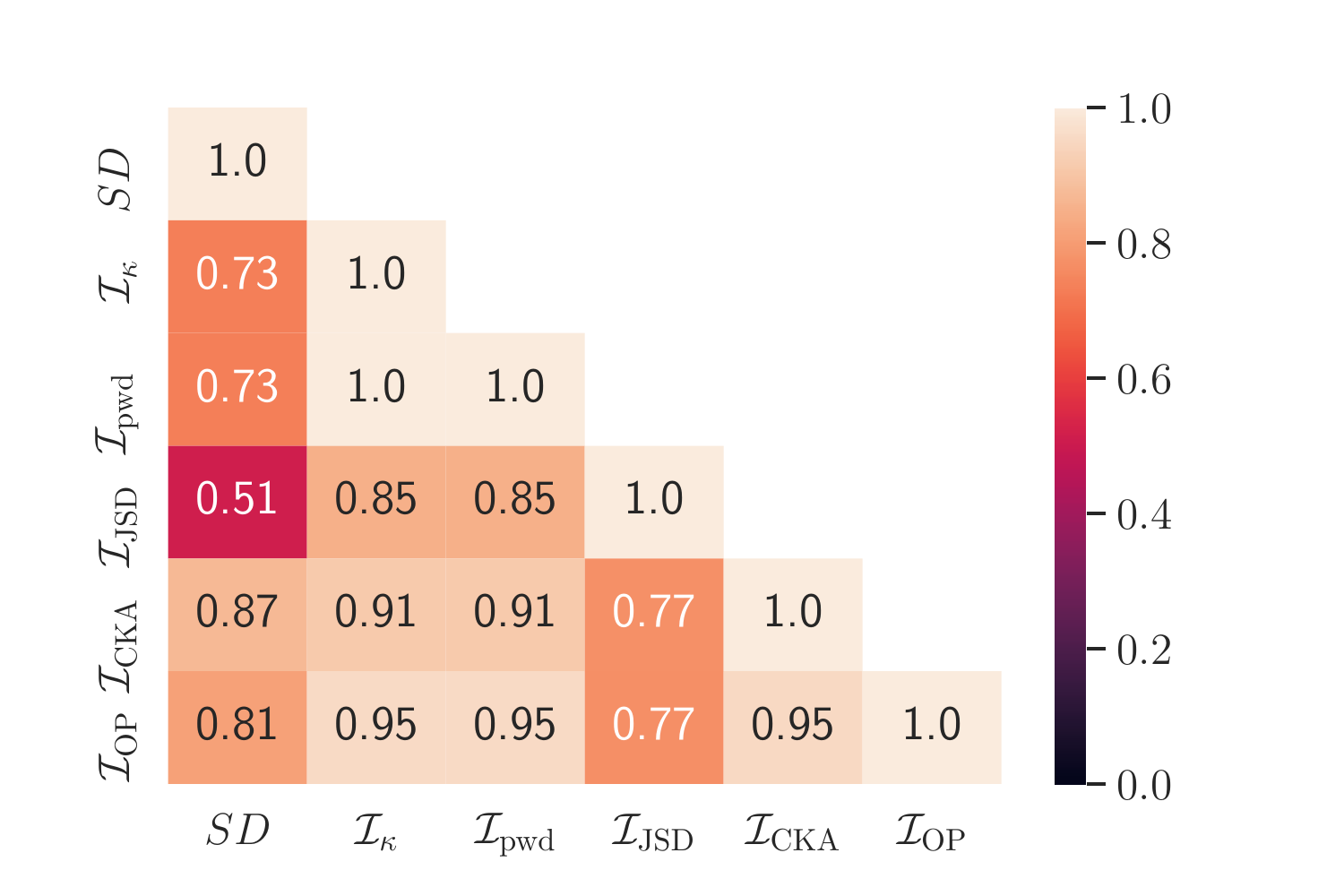}
        \caption{CoLA}
        \label{subfig:bs_cola_llrd_roberta}
    \end{subfigure}
    \caption{
        Bootstrapping results of LLRD RoBERTa.
    }
    \label{fig:app_bs_roberta_llrd}
\end{figure*}

\begin{figure*}
    \centering
    \begin{subfigure}[b]{0.32\textwidth}
        \centering
        \includegraphics[width=\textwidth]{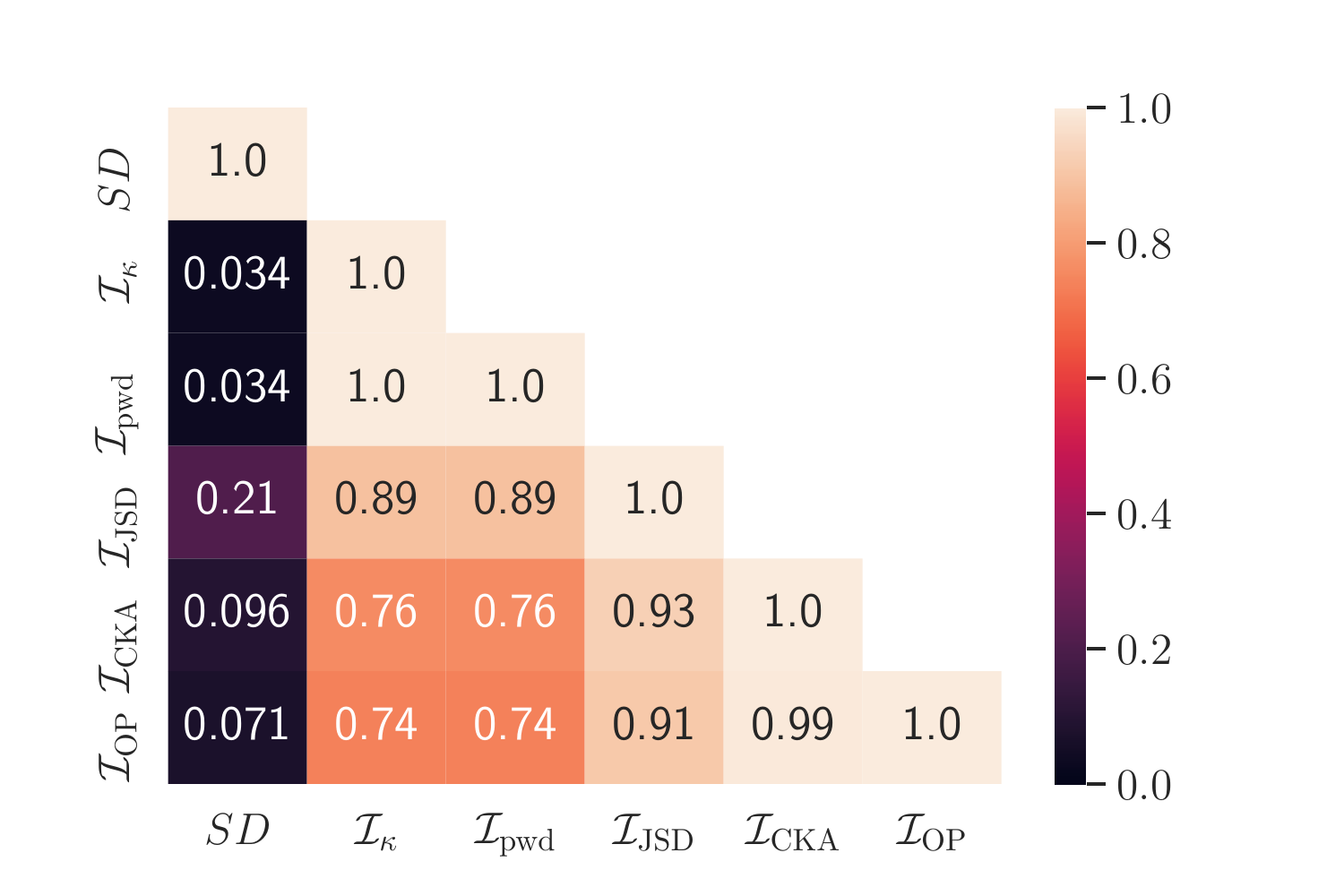}
        \caption{RTE}
        \label{subfig:bs_rte_reinit_roberta}
    \end{subfigure}
    \hfill
    \begin{subfigure}[b]{0.32\textwidth}
        \centering
        \includegraphics[width=\textwidth]{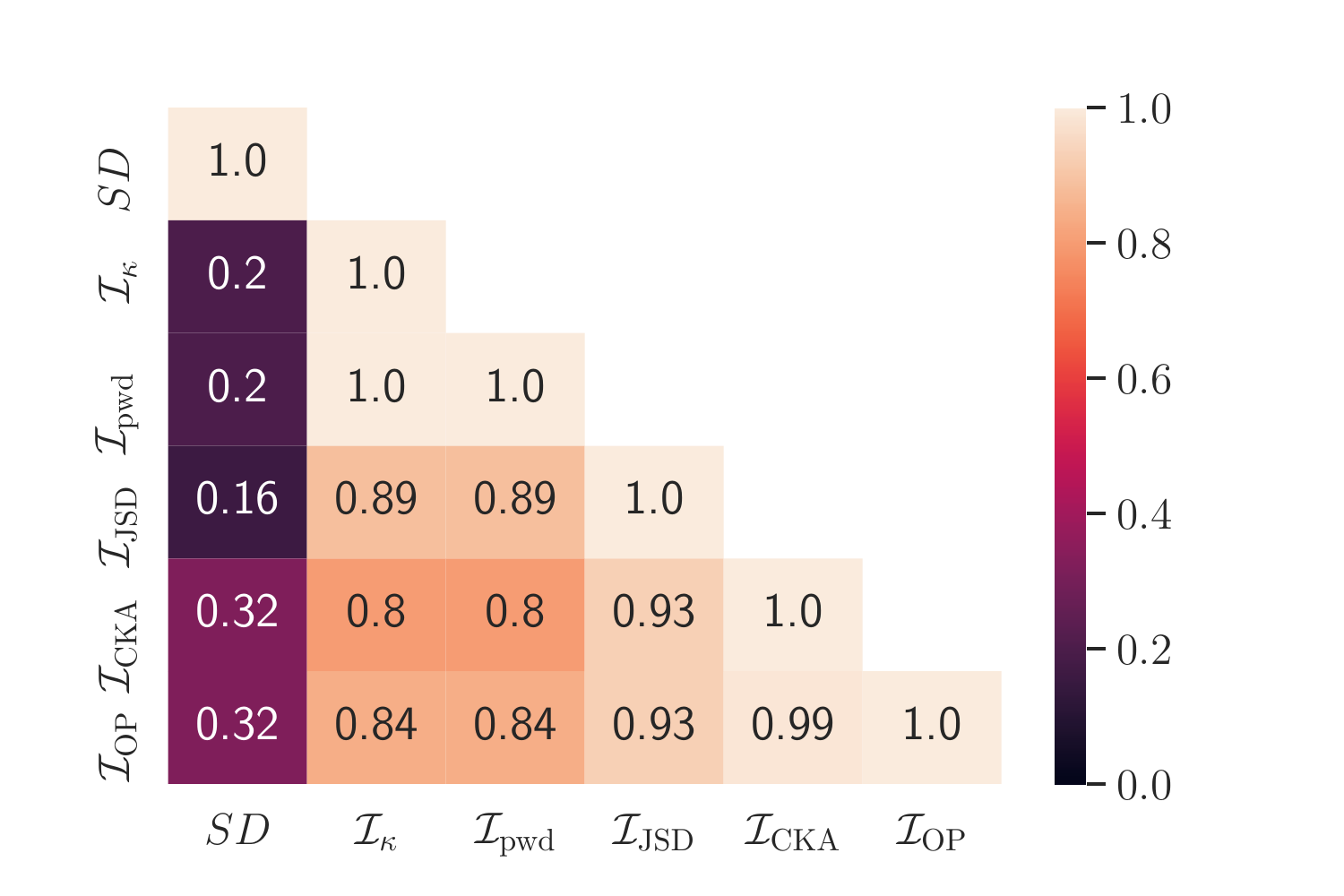}
        \caption{MRPC}
        \label{subfig:bs_mrpc_reinit_roberta}
    \end{subfigure}
    \hfill
    \begin{subfigure}[b]{0.32\textwidth}
        \centering
        \includegraphics[width=\textwidth]{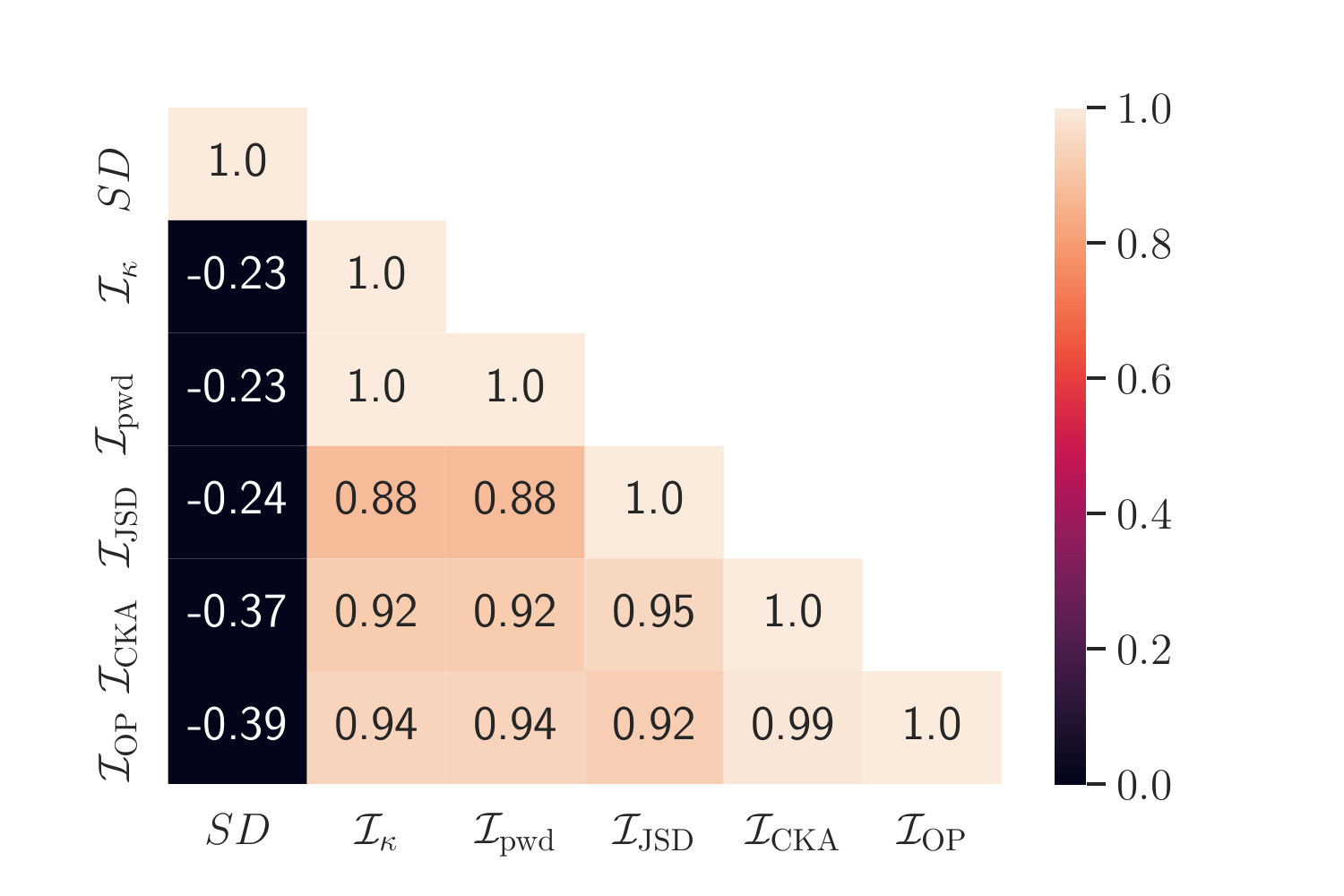}
        \caption{CoLA}
        \label{subfig:bs_cola_reinit_roberta}
    \end{subfigure}
    \caption{
        Bootstrapping results of Re-init RoBERTa.
    }
    \label{fig:app_bs_robert_reinit}
\end{figure*}

\begin{figure*}
    \centering
    \includegraphics[width=0.45\textwidth]{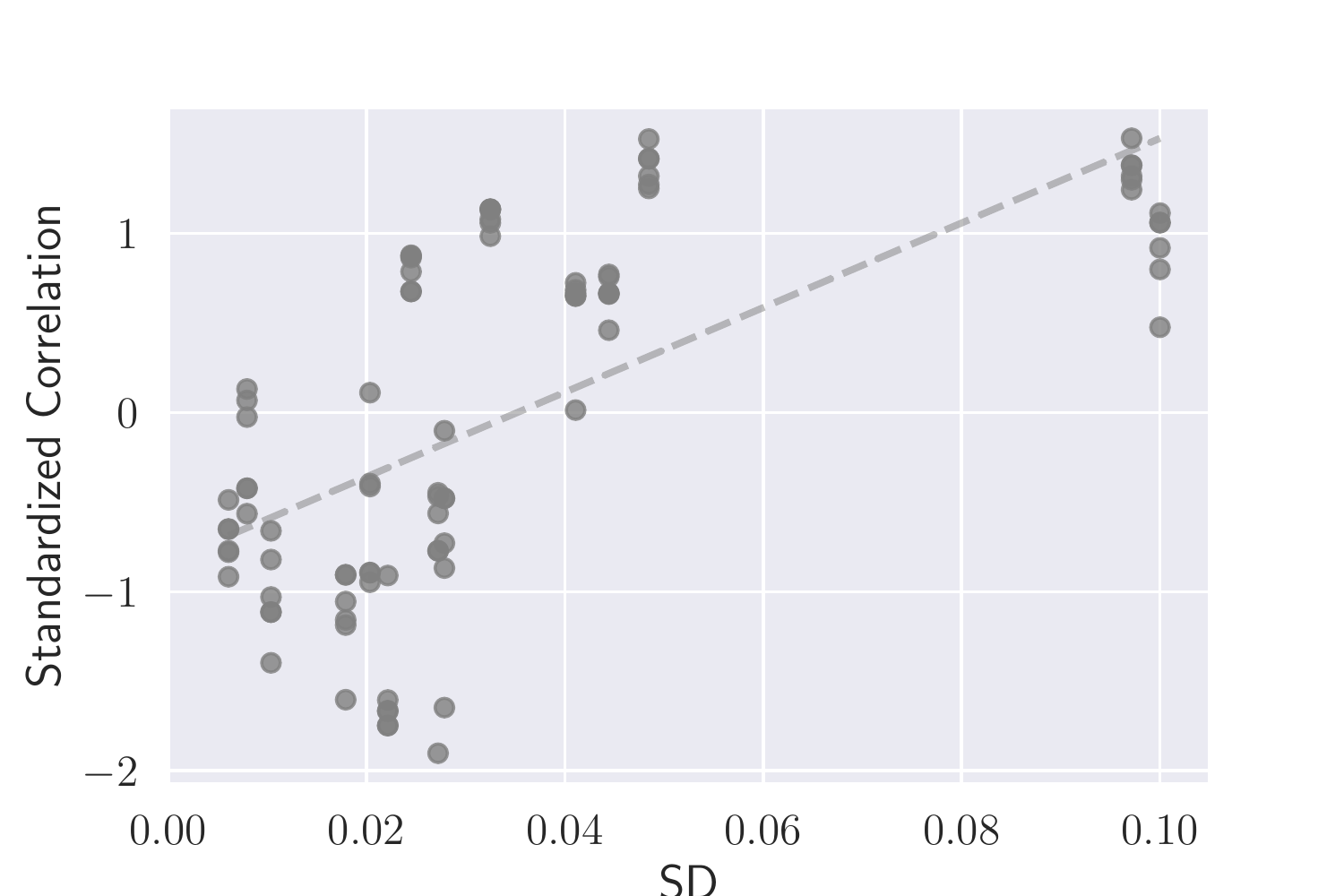}
    \caption{
        Correlation between 
        1) the average standardized correlation between each measure and other measures
        2) the corresponding SD value,
        for RoBERTa on each dataset/IMM combination.
        Pearson's $r=0.653$.
    }
    \label{fig:app_bs_bert_correlation}
\end{figure*}